\documentclass[10pt,journal,compsoc]{IEEEtran}
\usepackage{soul}
\usepackage{url}
\usepackage[hidelinks]{hyperref}
\usepackage[utf8]{inputenc}
\usepackage[small]{caption}
\usepackage{graphicx}
\usepackage{amsmath}
\usepackage{amsthm}
\usepackage{booktabs}
\usepackage{algorithm}
\usepackage[noend]{algpseudocode}

\usepackage{amsfonts}       
\usepackage{nicefrac}       
\usepackage{microtype}      
\usepackage{xcolor}         

\usepackage{multirow}
\usepackage{multicol}

\usepackage{enumerate}
\usepackage{subcaption, caption}
\DeclareCaptionStyle{ruled}{labelfont=normalfont,labelsep=colon,strut=off}

\usepackage[flushleft]{threeparttable}
\usepackage{float}
\usepackage{xspace}
\usepackage{caption}
\usepackage{diagbox}
\theoremstyle{definition}
\newtheorem{defn}{Definition}[]
\newtheorem{theorem}{Theorem}[]

\usepackage{amsmath,amssymb}
\DeclareMathOperator{\E}{\mathbb{E}}

\ifCLASSOPTIONcompsoc
  \usepackage[nocompress]{cite}
\else
  \usepackage{cite}
\fi

\ifCLASSINFOpdf

\else

\fi

\begin{document}
\title{FedIPR: Ownership Verification for Federated Deep Neural Network Models}
\author{Bowen~Li,
        Lixin~Fan,~\IEEEmembership{Member,~IEEE,}
        Hanlin~Gu,
        Jie~Li,~\IEEEmembership{Senior Member,~IEEE,}
        and~Qiang~Yang,~\IEEEmembership{Fellow,~IEEE}
\IEEEcompsocitemizethanks{\IEEEcompsocthanksitem Bowen Li and Jie Li are with the Department of Computer Science and Engineering, Shanghai Jiao Tong University, Shanghai 200240, China. The work is done when Bowen Li is an intern at WeBank.  
E-mail: \{li-bowen, lijiecs\}@sjtu.edu.cn.
\IEEEcompsocthanksitem Lixin Fan and Hanlin Gu are with WeBank AI Lab, WeBank, China. E-mail: lixinfan@webank.com, \{Lixin.Fan01, ghltsl123\}@gmail.com. 
\IEEEcompsocthanksitem  Qiang Yang is with the Department of Computer Science and Engineering, Hong
Kong University of Science and Technology, Hong Kong and  WeBank AI Lab, WeBank, China. E-mail: qyang@cse.ust.hk.}
\thanks {Corresponding author: Lixin Fan.}}


\IEEEtitleabstractindextext{%
\begin{abstract}
    Federated learning models are collaboratively developed upon valuable training data owned by multiple parties. During the development and deployment of federated models, they are exposed to risks including illegal copying, re-distribution, misuse and/or free-riding.  To address these risks, the ownership verification of federated learning models is a prerequisite that protects federated learning model intellectual property rights (IPR) i.e., FedIPR. We propose a novel federated deep neural network (FedDNN) ownership verification scheme that allows private watermarks to be embedded and verified to claim legitimate IPR of FedDNN models. 
    In the proposed scheme, each client independently verifies the existence of the model watermarks and claims respective ownership of the federated model without disclosing neither private training data nor private watermark information. The effectiveness of embedded watermarks is theoretically justified by the rigorous analysis of conditions under which watermarks can be privately embedded and detected by multiple clients. Moreover, 
    extensive experimental results on computer vision and natural language processing tasks demonstrate that varying bit-length watermarks can be embedded and reliably detected without compromising original model performances. 
	Our watermarking scheme is also resilient to various federated training settings and robust against removal attacks.
\end{abstract}

}

\maketitle

\IEEEdisplaynontitleabstractindextext

\IEEEpeerreviewmaketitle

\IEEEraisesectionheading{\section{Introduction}\label{sec:introduction}}


\IEEEPARstart{T}{he} successful applications of 
deep neural network (DNN) to computer vision, natural language processing and data mining tasks come at the cost of the expensive training process: a) the training incurs substantial efforts and costs in terms of expertise, dedicated hardware, and exceedingly long time for the designing and training DNN \textit{models}; 
b) it requires a vast amount of training \textit{data} to boost the model performance, which often increases  monotonically with the volume of training data \cite{deng2009imagenet, wang2021voxpopuli,  chelba2013one, zhu2015aligning}. 
To protect both the valuable training data and the trained DNN models from being illegally copied, re-distributed or misused, therefore, becomes a compelling need that motivates our research work reported in this article.

To protect the Intellectual Property Rights (IPR) of Deep Neural Network, DNN watermarking techniques have been proposed in \cite{fan2021deepip,GANsign-CVPR21,zhang2021deep,lim2022protect, EmbedWMDNN_2017arXiv, DeepMarks_2018arXiv, DeepSigns_2018arXiv,TurnWeakStrength_Adi2018arXiv}  
to embed designated watermarks into DNN models. 
Subsequently, DNN ownership is verified by robustly extracting the embedded watermarks from the model in question.  
Note that both \textit{feature-based watermarks}\cite{EmbedWMDNN_2017arXiv, DeepMarks_2018arXiv, DeepSigns_2018arXiv} and \textit{backdoor-based watermarks}\cite{TurnWeakStrength_Adi2018arXiv} have been proposed to verify ownership of DNN models. 
In order to protect valuable training data in a collaborative learning setting whereas \textit{semi-honest} adversaries may attempt to espy participants’ private information, a secure federated learning (SFL) framework has been proposed\cite{communicationEfficient/mcMahan2017, yang2019federated, OpenFL-19} to 
collaboratively train a federated deep neural network (FedDNN) without giving away to adversaries private training data\cite{DeepLeakage_Han19} 
and data feature distribution\cite{luo2021feature}.
Therefore, each client in federated learning must a) \textit{not} disclose to other parties any information about private training data; 
and b) prove ownership of the trained model without disclosing their private watermarks. 
The first requirement has been fulfilled by protecting the exchanged  local models using techniques such as 
homomorphic encryption (HE) \cite{PPDL_Homomorphic_Encryption/trieu2018}, 
differential privacy (DP)\cite{DLDP_Abadi16} or secret sharing\cite{ARIANN-PPDL-sec20} albeit at cost of degraded model performances\cite{zhang2022no}.  
The second requirement is one of the open problems considered in this work. 

Taking into consideration threat models in both DNN watermarking and secure federated learning, 
we propose a unified framework called FedIPR which consists of two separate processes along with standard SFL learning procedures: 
a) \textit{a watermark embedding process} that allows multiple parties to embed their secret feature-based and backdoor-based watermarks; 
b) \textit{a verification process} that allows each party to independently verify the ownership of FedDNN model. 


Two technical challenges for embedding watermarks into FedDNN model are investigated in this paper: 
\begin{itemize}
  \item \textbf{Challenge A:} \textit{how to ensure that private watermarks embedded by different clients into the same FedDNN model do not discredit each other?} This challenge is unique in a federated learning setting whereas different client's watermarks may potentially conflict with each other (see Fig. \ref{fig:conflict} for an example). As a solution to the challenge, 
  theoretical analysis in Theorem \ref{thm: thm1} elucidates conditions under which multiple 
  feature-based watermarks can be embedded into the same FedDNN model without bringing each other into discredit, and based on the theoretical analysis, a feature-based watermarking method dedicated for horizontal federated learning is proposed. (see Sect. \ref{section: Feature-based embedding} for details). 
  
  \item \textbf{Challenge B:} \textit{how to ensure that embedded watermarks are robust to privacy-preserving learning strategies?} This challenge is due to modifications of model parameters brought by various privacy preserving methods e.g., differential privacy  \cite{DLDP_Abadi16}, defensive aggregation\cite{blanchard2017machine, guerraoui2018hidden, yin2018byzantine} and client selection\cite{communicationEfficient/mcMahan2017}. 
  As a solution, FedIPR adopts robust client-side training to embed both feature-based and backdoor-based watermarks.  
  Our empirical results in Sect. \ref{sect:exper} show that robust feature-based and backdoor-based watermarks are persistent under various federated learning strategies. 
  
\end{itemize}

Moreover, extensive experiments on computer vision and natural language processing tasks demonstrate 
that feature-based watermarks embedded in \textit{normalization scale parameters} (see Sect. \ref{section: Feature-based embedding} for details) are highly reliable, while backdoor-based watermarks can be reliably detected for black-box ownership verification. 
In short, main contributions of our work are threefold: 
\begin{itemize}
    \item We put forth the first general framework called FedIPR for ownership verification of DNN models in a secure federated learning setting. FedIPR is designed in such a way that each client can embed his/her own private feature-based and backdoor-based watermarks and 
    verify watermarks to claim ownership independently. 
    \item We demonstrate successful applications of FedIPR for various DNN model architectures trained in the semi-honest federated learning setting. 
    Theoretical analysis of the \textit{significance} of feature-based watermarks and superior performance with extensive experimental results showcase the efficacy of the proposed FedIPR framework. 
    \item FedIPR also provides an effective method to detect freeriders\cite{fraboni2021free, lin2019free} who do not contribute data or
    computing resources but participate in federated learning to get for free the valuable model. Due to the lack of rightful watermarks embedded in the FedDNN model, freeriders can be discerned from benign participants. 
\end{itemize}

To our best knowledge, the FedIPR framework is the first technical solution that supports the protection of DNN ownerships in a secure federated learning setting such that secret watermarks embedded in FedDNN models do not disclose to semi-honest adversaries. 

The rest of the paper is organized as follows: Section \ref{sect:related} briefly reviews previous work related to secure federated learning and DNN ownership verification. Section \ref{subsect:DNNfeatsig} describes the preliminary background for FedIPR. 
Section \ref{sect:FedIPR-fw} illustrates the proposed FedIPR framework formulation. Section \ref{sect: Embed} delineates the watermark embedding approaches both in white-box and black-box modes, and Section \ref{sect:exper} presents experimental results and showcases the robustness of FedIPR. 
We discuss and conclude the paper in Section  \ref{Discussion}.

\section{Related Work} \label{sect:related}
We briefly review related work in three following aspects and refer readers to survey articles in respective aspects\cite{OpenFL-19,lyu2020threats, boenisch2020survey}.


\subsection{Secure Federated Learning}

Secure Federated learning\cite{yang2019federated,communicationEfficient/mcMahan2017, OpenFL-19} aims to collaboratively train a global machine learning model 
among multiple clients without disclosing private training data to each other \cite{PPDL/shokri2015,DLDP_Abadi16,PPDL_Homomorphic_Encryption/trieu2018,ARIANN-PPDL-sec20}. 
Moreover, privacy-preserving techniques such as homomorphic encryption \cite{PPDL_Homomorphic_Encryption/trieu2018}, 
differential privacy  \cite{DLDP_Abadi16} and secret sharing \cite{ARIANN-PPDL-sec20} were often used to protect exchanged local models\cite{yang2019federated,communicationEfficient/mcMahan2017}. 

\subsection{Threats to Model IPR}
It was shown that FedDNN models of high commercial values were subjected to severe IPR  threats\cite{Gartner, tramer2016stealing, orekondy2019knockoff, fraboni2021free}. Firstly,  unauthorized parties might plagiarize the DNN model with non-technical methods\cite{Gartner}. Secondly, Tramer et al. showcased \textit{model stealing attacks} that aimed to steal deployed victim models  even if attackers have no knowledge of training samples or model parameters\cite{tramer2016stealing}. Thirdly, Fraboni et al. demonstrated that \textit{freeriders} might join in federated learning and plagiarized the valuable models with no real contributions to the improvements of federated models\cite{fraboni2021free}.




\subsection{DNN Watermarking Methods}
As a counter measure against model plagiarisms, private  watermarks are embedded into the DNN model parameters and functionality, which have been strongly combined with the  protected DNN model. Two categories of DNN watermarking methods have been proposed: 

\textit{Backdoor-based} methods proposed to use a particular set of inputs as the triggers and let the model
deliberately output specific incorrect labels\cite{TurnWeakStrength_Adi2018arXiv,ProtectIPDNN_Zhang2018, lukas2020deep}. Backdoor-based methods collected evidence of suspected plagiarism through remote API without accessing internal parameters of models.
We also refer to a recent survey\cite{boenisch2020survey} for more existing watermark embedding schemes. 

\textit{Feature-based} methods proposed to encode designated binary strings as watermarks into layer parameters in DNN models\cite{EmbedWMDNN_2017arXiv,DeepMarks_2018arXiv,DeepSigns_2018arXiv,fan2021deepip,Passaware-NIPS20}. Specifically, Uchida et al. \cite{EmbedWMDNN_2017arXiv} proposed to embed feature-based watermarks into convolution layer weights using a binary cross-entropy loss function. 
Fan et al.\cite{fan2019rethinking} proposed to embed feature-based watermarks into \textit{normalization layer scale parameters} of the convolution block with a hinge-like  regularization term. 
In the verification stage of feature-based watermarks, one must access DNN internal parameters to detect watermarks.



For federated learning model verification scenario, double masking protocols\cite{xu2019verifynet, han2022verifiable} are proposed as FedDNN integrity verification schemes while guaranteeing user’s privacy in the training process.   
However, those model integrity verification methods could not preserve the IPR of FedDNN models. For IPR  protection of FedDNN, Atli et al.\cite{atli2020waffle} adopted \textit{backdoor-based} watermarks to 
enable ownership verification for the central server. 
Nevertheless, they only considered the setting in which the server was responsible for embedding watermarks into the global FedDNN model, and did not allow clients to embed and verify private watermarks. 
Liu et al. \cite{liu2021secure} has adopted client-side  backdoor-based watermarking method under the homomorphic encryption FL framework, while our proposed FedIPR consider both feature-based and backdoor-based watermarking in a general secure federated learning scenario with strategies like differential privacy\cite{DLDP_Abadi16}, homomorphic encryption\cite{PPDL_Homomorphic_Encryption/trieu2018},  defensive aggregation\cite{blanchard2017machine}, etc.


\section{Preliminaries} \label{subsect:DNNfeatsig}
In this section, we first review and formulate key ingredients of the secure horizontal federated learning and existing DNN watermarking methods as follows. We also explain in Tab. \ref{tab:notation} all notations used in this article.

\subsection{Secure Horizontal Federated Learning}\label{Sect. SFL}

A secure horizontal federated learning\cite{yang2019federated} system consists of $K$ clients which build local models\footnote{Other works\cite{communicationEfficient/mcMahan2017} also call them model updates, because the local models are equal to model updates for aggregation.} with their own data and send local models $\{\mathbf{W}_k\}_{k=1}^K$ to an aggregator to obtain a global model. The aggregator conducts the following aggregation process\cite{communicationEfficient/mcMahan2017,yang2019federated,OpenFL-19}:
\begin{equation}
	\mathbf{W} \leftarrow \sum_{k=1}^{K} \frac{n_k}{K} \mathbf{W}_k, 
\end{equation}
where $n_k$ is the weight for each client's local model $\mathbf{W}_k$. 


\noindent\textbf{Remark:} in secure federated learning, local model $\mathbf{W}_k$ might be protected by using Homomorphic Encryption (HE)\cite{PPDL_Homomorphic_Encryption/trieu2018}, Differential Privacy (DP)\cite{DLDP_Abadi16} such that  semi-honest adversaries can not infer private information from $\mathbf{W}_k$. These privacy preserving strategies pose one challenge to be addressed for reliable watermarking (see Sect. \ref{Challenge B}).

\subsection{Freeriders in Federated Learning} \label{freerider}

In federated learning, there might be \textit{freerider} clients\cite{fraboni2021free} 
who do not contribute data or computing resources 
but construct some superficial local models to participate in training only to obtain the global model for free. Specifically, there are several strategies for freeriders to construct local models\cite{fraboni2021free}:

\noindent\textbf{Freeriding with Previous Models (Plain Freerider).}  
Freeriders create a superficial model as follows\cite{fraboni2021free}, 

\begin{equation}
	\mathbf{W}^{free} = Free(\mathbf{W}^t, \mathbf{W}^{t-1}),
\end{equation}
in which $\mathbf{W}^t$, $\mathbf{W}^{t-1}$ denote respectively local models from two previous iterations. Note that the  construction of this superficial model costs nothing for freeriders since they are merely saved copies of model parameters from previous iterations.

\noindent\textbf{Freeriding with Gaussian Noise.} 
Freeriders adopt the previous global model parameters $\mathbf{W}^{t-1}$ and add Gaussian noise to simulate a local model:

\begin{equation}
	\mathbf{W}^{free} = \mathbf{W}^t + \xi_t,\quad \xi_t \sim \mathcal{N} (0, \sigma_t).
\end{equation}

Detection methods are proposed to detect and eliminate superficial local models as such\cite{lin2019free}. However it is required to train a meta freerider detector.

\begin{table}[h]
    \centering  
    \centering  
    \renewcommand\arraystretch{1.2}
	
	\setlength{\tabcolsep}{1mm}
	\centering  
	\resizebox{0.48\textwidth}{!}{

\begin{tabular}{|c|cccc|}
\hline
Notations & \multicolumn{4}{c|}{Descriptions}         \\ \hline
$K$      & \multicolumn{4}{c|}{ Number of clients in Secure Federated Learning} \\\hline
$\mathbb{N}$     & \multicolumn{4}{c|}{ Federated neural network model}\\\hline
$\mathbf{W}$     & \multicolumn{4}{c|}{ Model weights of model $\mathbb{N}$} \\\hline
$\mathbf{W}_k$     & \multicolumn{4}{c|}{ Local model of $k$-th client $\mathbb{N}$} \\\hline
$\mathcal{G()}$      & \multicolumn{4}{c|}{Key \textbf{Generation} Process} \\\hline
$N_{\mathbf{T}}$ & \multicolumn{4}{c|}{Bit-length of backdoor-based watermarks}\\\hline
$\mathbf{T}$ & \multicolumn{4}{c|}{Target \textit{backdoor-based} watermarks} \\ \hline 
  $(\mathbf{X}_{\mathbf{T}}, \mathbf{Y}_{\mathbf{T}})$ & \multicolumn{4}{c|}{Samples and labels of backdoor-based watermarks
$\mathbf{T}$  } \\ \hline

$N$ & \multicolumn{4}{c|}{Bit-length of feature-based watermarks}\\\hline

$\mathbf{B}$ & \multicolumn{4}{c|}{Target \textit{feature-based} watermarks} \\ \hline 
$\tilde{\mathbf{B}}$ & \multicolumn{4}{c|}{\textit{Feature-based} watermarks extracted from the parameters} \\ \hline

 $\theta = \{\mathbf{S}, \mathbf{E}\}$ & \multicolumn{4}{c|}{Secret  parameters for \textit{feature-based} watermarks} \\ \hline 
$\mathbf{S}$ & \multicolumn{4}{c|}{Watermark location parameters} \\ \hline
  $\mathbf{E}$ & \multicolumn{4}{c|}{Watermark embedding matrix}  \\ \hline
  $\mathcal{E()}$      & \multicolumn{4}{c|}{Watermark \textbf{Embedding} Process} \\\hline
  $\mathbf{W}_k^t$      & \multicolumn{4}{c|}{Local model of $k$-th client at communication round $t$} \\\hline
  $\mathbf{W}^t$      & \multicolumn{4}{c|}{Global model at communication round $t$} \\\hline
  $L_{D}$ & \multicolumn{4}{c|}{The loss function for the main learning task} \\\hline 
  $L_{\mathbf{T}}$ & \multicolumn{4}{c|}{Backdoor-based watermark embedding regularization term} \\ \hline
  
 $L_{\mathbf{B}, \theta}$ &\multicolumn{4}{c|}{Feature-based watermark embedding regularization term} \\ \hline
   $\mathcal{A()}$      & \multicolumn{4}{c|}{\textbf{Aggregation} Process in Secure Federated Learning} \\\hline
  $\mathcal{V()}$      & \multicolumn{4}{c|}{Watermark \textbf{Verification } Process} \\\hline
  $\mathcal{V}_W()$ & \multicolumn{4}{c|}{White-box verification} \\\hline 
  $\mathcal{V}_B()$ & \multicolumn{4}{c|}{Black-box verification} \\\hline 
  $\eta_F$ & \multicolumn{4}{c|}{Detection rate of feature-based watermarks} \\\hline
  $\eta_T$ & \multicolumn{4}{c|}{Detection rate of backdoor-based  watermarks} \\\hline
  
\end{tabular}}
\caption{Notations used in this article.}  
\label{tab:notation}  
\end{table}

\subsection{DNN Watermarking Methods}
\begin{figure}[t]
\vspace{-10pt}
	\centering
	\includegraphics[width=3.5in]{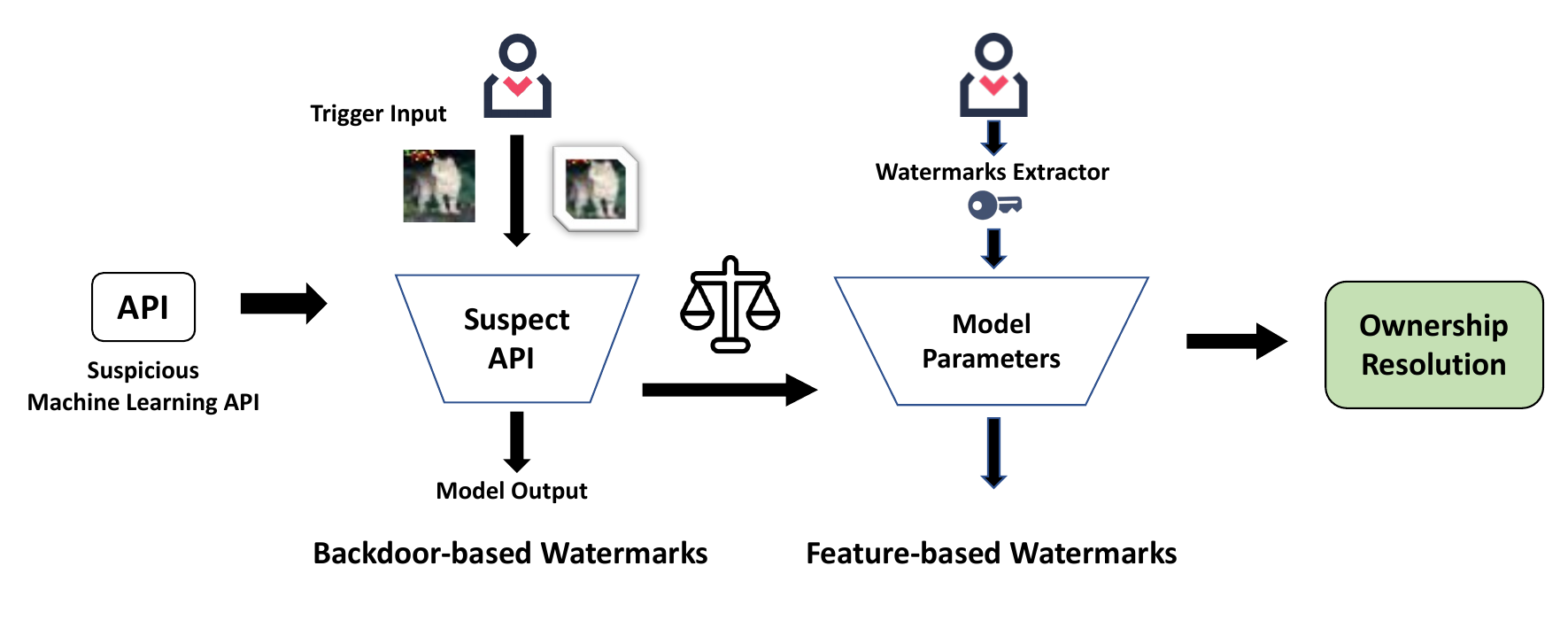}
   
	\caption{Ownership verification processes composed of backdoor-based watermarks and feature-based watermarks}
	\label{fig:verification}
	\vspace{-10pt}
\end{figure}

There are broadly two categories of DNN watermarking methods: 

\noindent\textbf{Backdoor-based Watermarks\cite{TurnWeakStrength_Adi2018arXiv, lukas2020deep}.}
Backdoor-based watermarks $\mathbf{T} = \{ (\mathbf{X}_{\mathbf{T}}^1, \mathbf{Y}_{\mathbf{T}}^1), \cdots, (\mathbf{X}_{\mathbf{T}}^{N_{\mathbf{T}}}, \mathbf{Y}_{\mathbf{T}}^{N_{\mathbf{T}}})  \}$ are embedded into the model function $\mathbb{N}$ during the training time by incorporating a loss function of backdoor samples.  

In the verification step (as shown in the second procedure of Fig. \ref{fig:verification}), backdoor samples are used as the trigger input to the model $\mathbb{N}$. 
The ownership is successfully verified if the detection error of designated backdoor labels is less than a threshold $\epsilon_B$: 
\begin{equation}\label{eq:verfiyDNN_black}
	\mathcal{V}_B\big( \mathbb{N}, \mathbf{T} \big) =  \left\{ \begin{array}{cc}
	\text{TRUE}, & \text{if } \mathop{{}\mathbb{E}}_{\mathbf{T}_n}( \mathbb{I}(\mathbf{Y}_{\mathbf{T}} \neq \mathbb{N}(\mathbf{X}_{\mathbf{T}}) ) ) \leq \epsilon_B,  \\
	\text{FALSE}, & \text{otherwise},  \\
	\end{array} \right.
\end{equation}  
in which $\mathcal{V}_B()$ is the ownership verification process that only accesses model API in black-box mode.

\noindent\textbf{Remark:} $(\mathbf{B}, \mathbf{\theta}, \mathbf{T})$ are private watermarking information that should be kept secret without disclosing to other parties. 

\noindent\textbf{Feature-based Watermarks\cite{EmbedWMDNN_2017arXiv,DeepMarks_2018arXiv,DeepSigns_2018arXiv,fan2019rethinking}.} In the watermark embedding step, $N$-bits target binary watermarks $\mathbf{B}\in \{0,1\}^N$  are embedded during the learning of model parameters $\mathbf{W}$, 
by adding regularization terms to the original learning task. 

During the verification step (as shown in the third procedure of Fig. \ref{fig:verification}), feature-based watermarks  $\mathbf{\tilde{B}}$ extracted with extractor  $\mathbf{\theta}$ from DNN parameters is then matched with the designated watermarks $\mathbf{B}$, to judge if Hamming distance 
$\text{H}(\mathbf{B},\mathbf{\tilde{B}})$ is less than a preset threshold  $\epsilon_W$: 
\begin{equation}\label{eq:verfiyDNN}
	\mathcal{V}_W\big( \mathbf{W}, (\mathbf{B}, \mathbf{\theta}) \big) =  \left\{ \begin{array}{cc}
	\text{TRUE}, & \text{if }\text{H}(\mathbf{B}, \mathbf{\tilde{B}}) \leq \epsilon_W,  \\
	\text{FALSE}, & \text{otherwise},  \\
	\end{array} \right.
\end{equation}  
in which $\mathcal{V}_W()$ is the ownership verification process that has to access model parameters in a white-box mode.

\begin{figure*}[t]
     \vspace{-5pt}
    \centering
    \includegraphics[width=6.4in]{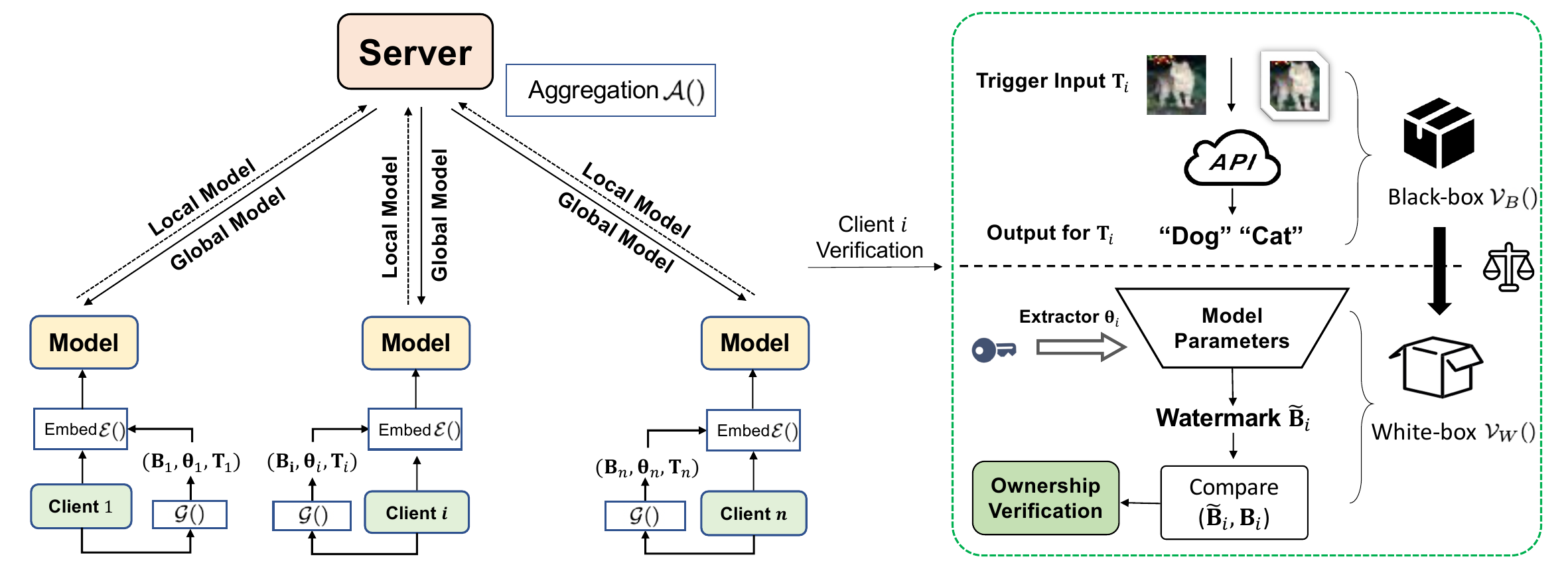}
    \caption{\small An illustration of federated DNN (FedDNN) watermark Embedding and Verification scheme. Private watermarks are generated and embedded into the local models, which are then aggregated using the \texttt{FedAvg} algo. (the left panel). In case the federated model is plagiarized, each client may invoke verification processes to extract watermarks from the plagiarized model in both black-box and white-box manner to claim his/her ownership of the federated model (the right panel).}\label{fig: framework}
    \label{fig:2}
        \vspace{-5mm}
\end{figure*}
\section{Federated DNN Ownership Verification} \label{sect:FedIPR-fw}

We propose a novel watermark embedding and ownership verification scheme called FedIPR for the secure horizontal federated learning scenario. 
FedIPR is designed in the way such that each client can a) protect his/her private data; and b) embed and verify his/her own private watermarks without disclosing information about private watermarks.

\subsection{FedIPR: FedDNN Ownership Verification with Watermarks} \label{subsect:FedDNN-form}



Following the framework of SFL in Sect. \ref{Sect. SFL}, 
we give below a formal definition of the FedIPR ownership verification scheme, which is pictorially illustrated in Fig. \ref{fig:2}.  


	\begin{defn}
	\label{def:ovs}
	A Federated Deep Neural Network (FedDNN) model ownership verification scheme (FedIPR) for a given network $\mathbb{N}[]$ is defined as a tuple $\mathcal{V} = (\mathcal{G}, \mathcal{E}, \mathcal{A}, \mathcal{V}_W, \mathcal{V}_B)$ of processes, consisting of: 
	\begin{enumerate}[I)]
		
		\item For client $k \in \{1, \cdots K\}$, a client-side \textit{key generation process} $\mathcal{G}() \rightarrow (\mathbf{B}_k, \theta_k, \mathbf{T}_k)$  generates 
		target watermarks  $\mathbf{B}_k$, watermark extraction parameters $\theta_k = \{\mathbf{S}_k, \mathbf{E}_k\}$ and a \textit{trigger set} (backdoor-based watermarks) $\mathbf{T}_k = \{ (\mathbf{X}_{\mathbf{T}_k}^1, \mathbf{Y}_{\mathbf{T}_k}^1), \cdots, (\mathbf{X}_{\mathbf{T}_k}^{N_{\mathbf{T}}}, \mathbf{Y}_{\mathbf{T}_k}^{N_{\mathbf{T}}})  \}$;

		\noindent\textbf{Remark:} 
		the $(\mathbf{B_k}, \mathbf{\theta_k}, \mathbf{T_k})$ are private watermarking parameters that should be kept secret without disclosing to other clients. In the extraction parameters $\theta_k = \{\mathbf{S}_k, \mathbf{E}_k\}$, $\mathbf{S}_k$ denotes the location of watermarks $\mathbf{B}_k$, and $\mathbf{E}_k$ denotes the secret embedding matrix for watermarks $\mathbf{B}_k$.
		
		\item A client-side FedDNN \textit{embedding process} $\mathcal{E}()$  minimizes the combined loss $L_k$ of the main task , and two regularization terms $L_{\mathbf{T}_k}$ and $L_{\mathbf{B}_k, \theta_k}$ to embed trigger samples $\mathbf{T}_k$ 
		and feature-based watermarks $\mathbf{B}_k$ respectively\footnote{A client $k$ may opt-out and not embed watermarks or trigger samples by setting $\alpha_k=0.0$ or $\beta_k=0.0$. 
		Following \cite{CanBDFL-arx19}, we adopt a \textit{random sampling} strategy in experiments to assign non-zero values to $\alpha_k, \beta_k$ to simulate the situation that clients make decisions on their own.}, once receives the global model $\mathbf{W}^t$ at communication round $t$, 
		\begin{equation}\label{eq:FedDNN-min-loss}
        \begin{aligned}
			L_k := \underbrace{L_{D_k}( \mathbf{W}^t)}_{\text{main task}} & + \alpha_k \underbrace{ L_{\mathbf{T}_k}(\mathbf{W}^t)}_{\text{backdoor-based}} + \beta_k \underbrace{ L_{\mathbf{B}_k, \theta_k}\big(\mathbf{W}^t\big)}_{\text{feature-based}}, \\
            & k \in \{1, \cdots K\},  
		\end{aligned}
        \end{equation} 
        where $D_k$ denotes the training data of client $k$,  $\alpha_k$\footnote{If backdoor samples are filled in batches in the implementation side of backdoor-based watermarking task, the parameter $\alpha_k$ is equal to 1} denotes the parameters to 
		control the backdoor-based watermarking loss $L_{\mathbf{T}_k}$, and $\beta_k$ denotes the factor for feature-based watermarking regularization term $L_{\mathbf{B}_k,\theta_k}$.
        
        \noindent\textbf{Remark:}  Note that a $\text{ClientUpdate}(L_k, \mathbf{W}^{t}) =: argmin L_k$  sub-routine seeks the optimal parameters and sends local model to the aggregator (see below for a server-side aggregation process). 
		
		\item A server-side FedDNN \textit{aggregation process} $\mathcal{A}()$  collects local models from $m$ randomly selected clients and performs model aggregation using 
		the \texttt{FedAvg} algorithm \cite{communicationEfficient/mcMahan2017} i.e.,
		\begin{equation}\label{eq:FedDNN-AVG}
        \begin{aligned}
			&\mathbf{W}^{t+1} \leftarrow \sum_{k=1}^{K} \frac{n_k}{n} \mathbf{W}^{t+1}_k,
		\end{aligned}
        \end{equation} 
        where $\mathbf{W}^{t+1}_k \leftarrow \text{ClientUpdate}(L_k,\mathbf{W}^{t})$ is the local model of client $k$ at round $t$, and $\frac{n_k}{n}$ denoted the aggregation weight for \texttt{Fedavg} algorithm.
        
        \noindent \textbf{Remark:}
        in SFL, strategies like Differential Privacy\cite{wei2020federated}, defensive aggregation mechanism\cite{blanchard2017machine,yin2018byzantine, guerraoui2018hidden} and 
client selection\cite{communicationEfficient/mcMahan2017} are widely used for privacy, security and efficiency.

After the global model $\mathbf{W}$ is trained with convergence, each client can conduct ownership verification as follows. 

		\item A client-side \textit{black-box verification process} $\mathcal{V}_B()$ checks whether the detection error of designated labels $ \mathbf{Y}_{\mathbf{T}_k}$ generated by trigger samples $ \mathbf{X}_{\mathbf{T}_k}$  is smaller than $\epsilon_B$, 
			\begin{equation}\label{eq:FedDNN-VB}
            \begin{small}
		\mathcal{V}_B\big( \mathbb{N}, \mathbf{T}_k \big) =  \left\{ \begin{array}{cc}
			\text{TRUE}, & \text{if }\mathop{{}\mathbb{E}}_{\mathbf{T}_k}( \mathbb{I}(\mathbf{Y}_{\mathbf{T}_k}  \neq \mathbb{N}( \mathbf{X}_{\mathbf{T}_k} ) ) ) \leq \epsilon_B,  \\
			\text{FALSE}, & \text{otherwise},  \\
		    \end{array} \right.
            \end{small}
		   \end{equation}  
		in which $\mathbb{I}()$ is the indicator function and $\mathop{{}\mathbb{E}}$ is the expectation over trigger set $\mathbf{T}_k$.        

		\item A client-side \textit{white-box verification process} $\mathcal{V}_W()$ extracts feature-based watermarks $\tilde{\mathbf{B}}_k =  \textit{sgn}\big(\mathbf{W}, \theta_k \big)$ with sign function $\textit{sgn}()$ from 
		the global model parameters $\mathbf{W}$, and verifies the ownership as follows,   
			\begin{equation}\label{eq:FedDNN-VW}
            \begin{small}
			\mathcal{V}_W\big( \mathbf{W}, (\mathbf{B}_k, \theta_k) \big) =  \left\{ \begin{array}{cc}
			\text{TRUE}, & \text{if }\text{H}(\mathbf{B}_k, \mathbf{\tilde{B}}_k) \leq \epsilon_W,  \\
			\text{FALSE}, & \text{otherwise},   \\
			\end{array} \right.
            \end{small}
			\end{equation}  
			in which $\text{H}(\mathbf{B}_k, \mathbf{\tilde{B}}_k)$ is the Hamming distance between $\mathbf{\tilde{B}}_k)$ and the target watermarks  $\mathbf{B}_k$, and $\epsilon_W$ is a preset threshold. 



	\end{enumerate}
	
\end{defn}

\noindent\textbf{Watermark Detection Rate.} For client ownership verification, the \textit{watermark detection rate} can be defined as:  

\begin{itemize}
    \item For $N$ bit-length feature-based watermarks $\mathbf{B}$, detection rate $\eta_F$ is calculated as 
\begin{equation}
    \eta_F := 1 - \frac{1}{N} H(\mathbf{B},\tilde{\mathbf{B}}),
\end{equation}
where $H(\mathbf{B},\tilde{\mathbf{B}})$ measures Hamming distance between extracted binary watermark string $\tilde{\mathbf{B}}$ and the target watermarks $\mathbf{B}$; 
    \item For backdoor-based watermarks $\mathbf{T}$, the detection rate is 
\begin{equation}
    \eta_T := \mathbb{E}_{\mathbf{T}}( \mathbb{I}(\mathbf{Y}_{\mathbf{T}} = \mathbb{N}(\mathbf{X}_{\mathbf{T}}) ) ),
\end{equation}
which is calculated as the ratio of backdoor samples that are classified as designated labels w.r.t. the  total number $N_{\mathbf{T}}$ of trigger set. 
\end{itemize}


Given the definition of FedIPR, let us proceed to illustrate two technical challenges to be addressed by FedIPR.
\subsection{Challenge A: Conflicting goals of more than one watermarks in FedDNN}\label{Challenge A}

\vspace{-5pt}


For all clients, the \textit{watermark capacity} measures the overall bit-length of watermarks that can be significantly verified. The first challenge is to determine the maximal capability of multiple watermarks that can be embedded by $K$ clients in SFL without discrediting each other.

Specifically for \textit{feature-based} watermarks, it remains an open question whether there is a common solution for different clients to embed their private designated watermarks. 
To illustrate the potential conflict between multiple  watermarks, let us investigate the following examples: 


\noindent\textbf{Example 1:} two clients need to embed different watermarks  $\mathbf{B}_1=010$ and $\mathbf{B}_2= 101$, respectively, into the same parameters $\mathbf{W}=(w_1, w_2,w_3,w_4,w_5)$ with the same embedding matrix 
\begin{equation}
\mathbf{E} = \left(
\begin{array}{ccccc}
    e_{11} & e_{12} & e_{13} & e_{14} & e_{15}\\
    e_{21} & e_{22} & e_{23} & e_{24} & e_{25} \\
    e_{11} & e_{12} & e_{33} & e_{34} & e_{35} \\
\end{array}
\right)^T,
\end{equation}
the watermarks extracted from parameters $\mathbf{W}=(w_1, w_2,w_3,w_4,w_5)$ are $\tilde{\mathbf{B}} = \mathbf{W}\mathbf{E}$, and regularization terms are used to  constrain the parameters to satisfy:
\begin{equation}
        \left\{ \begin{array}{cc}
		\text{(010):} & \sum^{5}_{i=1}w_ie_{1i}<0, \sum^{5}_{i=1}w_ie_{2i}>0, \sum^{5}_{i=1}w_ie_{3i}<0 , \\
		\text{(101):} &\sum^{5}_{i=1}w_ie_{1i}>0, \sum^{5}_{i=1}w_ie_{2i}<0, \sum^{5}_{i=1}w_ie_{3i}>0 ,  \\
		\end{array} \right.
\end{equation}
it is obvious that two different watermarks again impose conflicting constraints that cannot be simultaneously satisfied by the same  global model parameters.

\noindent\textbf{General Case:} for the feature-based watermarks $\{(\mathbf{B}_k, \mathbf{\theta}_k)\}_{k=1}^K$ embedded into the same model parameters $\mathbf{W}$ by $K$ different clients, each client $k$ embeds $N$ bit-length of feature-based watermarks $\mathbf{B}_k = (t_{k1}, t_{k2}, \cdots, t_{kN}) \in \{+1, -1\}^N$, the extracted watermarks $\tilde{\mathbf{B}}_k = \mathbf{W}\mathbf{E}_k$ should be consistent with targeted watermarks $\mathbf{B}_k$, i.e.,:
\begin{equation}
     \forall j \in \{1, 2, \dots, N\}\quad and \quad k \in K,  t_{kj} (\mathbf{W}\mathbf{E}_k)_j > 0.  
\end{equation}

\begin{figure}[htbp]
\vspace{-7pt}
	\centering
	\includegraphics[width=3.5in]{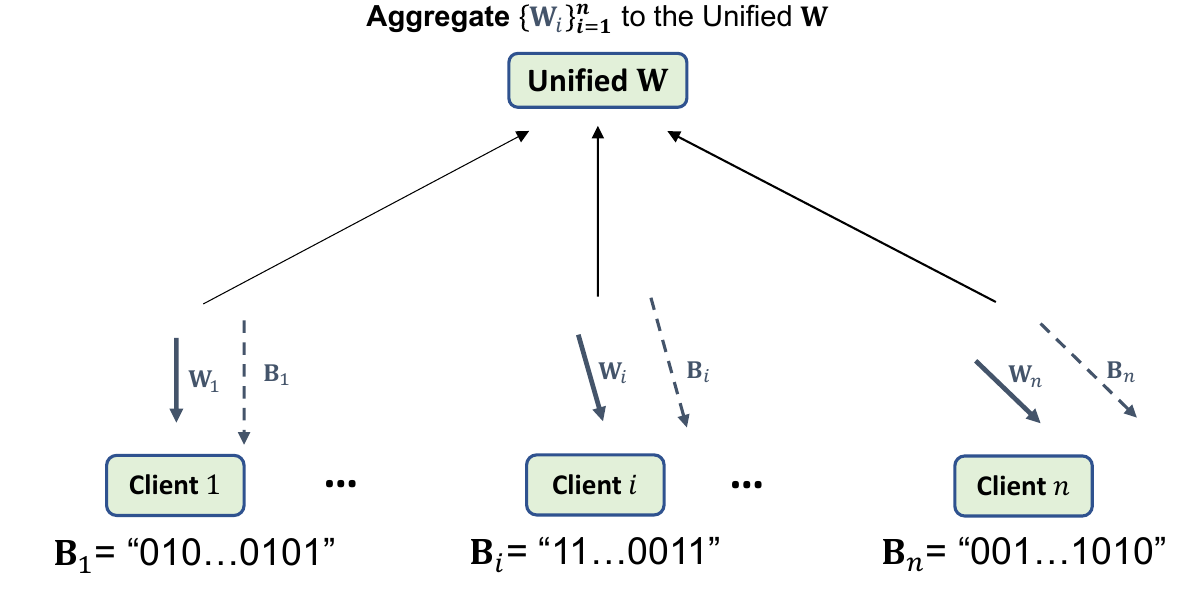}
	\caption{Different clients in federated learning adopt different regularization terms to embed feature-based watermarks}
	\label{fig:conflict}
\vspace{-5pt}
\end{figure}

As Fig.  \ref{fig:conflict} shows, each client $k$ tries to guide the target parameter $\mathbf{W}_k$ to a special direction conditioned on $\mathbf{B}_k$, but each $\mathbf{W}_k$ is aggregated into an unified $\mathbf{W}$ according to Eq. \eqref{eq:FedDNN-AVG} in SFL, these constraints may conflict with each other.  


Theorem \ref{thm: thm1} elucidates the condition under which a feasible solution exists for $K$ different watermarks to be embedded without conflicts and provides the lower bound of detection rate $\eta_F$.


\begin{theorem} \label{thm: thm1}
For $K$ different watermarks ($N$ bit-length each) to embed in $M$ channels of the global model parameters $\mathbf{W}$, 
take their detection rate to be measured as $\eta_F = 1 - \frac{1}{N} H(\mathbf{B},\tilde{\mathbf{B}})$, where $H(\mathbf{B},\tilde{\mathbf{B}})$ is the hamming distance between extracted watermarks $\tilde{\mathbf{B}}$ and the target watermarks $\mathbf{B}$. The watermark detection rate $\eta_F$ satisfies: 
\begin{item}
\item \textbf{Case 1:} If $KN\leq M$ \footnote{Another condition for this theorem is the embedding matrix $\textbf{E}$ clients decide needs to be column (line) non-singular matrix (see the detail in Appendix A)}, then there exists $\mathbf{W}$ such that $\eta_F =1$.

\item \textbf{Case 2:} If $KN>M$, then, there exists $\mathbf{W}$ such that
\begin{equation}
 \eta_F \geq \frac{KN+M}{2KN}.
\end{equation}\label{eq: case2}
\end{item}
The proof is deferred in Appendix A. 
\end{theorem}

\noindent\textbf{Remark:} results in \textbf{case 1} ($KN<M$) demonstrate the existence of a solution for watermark embedding when the total bit-length of all clients' watermarks $KN$ is smaller than the total number $M$\footnote{For example, $M=896$ channels across the last 3 layers for AlexNet,  $M=2048$ channels across the last 4 layers for ResNet18 and $M=2304$ channels across the last 3 layers for DistilBERT (illustrated in Appendix C).} of network channels that can be used to embed watermarks;   
results in \textbf{case 2} ($KN>M$) provides a lower bound of detection rate of embedded  watermarks. 
For example, if $K=10$ clients decide to each embed $N=100$ bit-length watermarks into $M=600$ channels of model parameters $\mathbf{W}$, the lower bound of detection rate is $\eta_F = 0.8$.

Taking both cases into account, we give below the optimal bit-length assigned to multiple watermarks under different situations (detailed analysis is deferred to Appendix B).

\noindent\textbf{Optimal bit-length and maximal bit-length $N$\footnote{However, other factors may influence $\eta_F$. For example, as experiment results in Fig. \ref{fig:removal}, \ref{fig:DP} demonstrated, random noise added to the federated learning process or removal attacks launched by plagiarizers, all conspire to degrade $\eta_F$ to various extents.}.} 
 We treat feature-based ownership verification as a hypothesis testing, where $\mathcal{H}_0$ is "the model is not plagiarized" versus  $\mathcal{H}_1$ "model is plagiarized", and we get p-value as the statistical significance of watermarks, the upper bound is given:

\begin{equation}
	p\text{-}value \leq \sum_{i=\eta_FN}^{N}\binom{N}{i} (1/C)^i(1-1/C)^{N-i}, 
\end{equation}
in which $C=2$.



With analysis of p-value (see detailed analysis in Appendix B), the optimal bit-length for the smallest p-value is $N_{opt} = M/K$, which is determined by the "smallest p-value" i.e., the most significant watermark verification. 

\begin{figure}[H]
	\centering
	\vspace{-5pt}
	\includegraphics[width=2.2in]{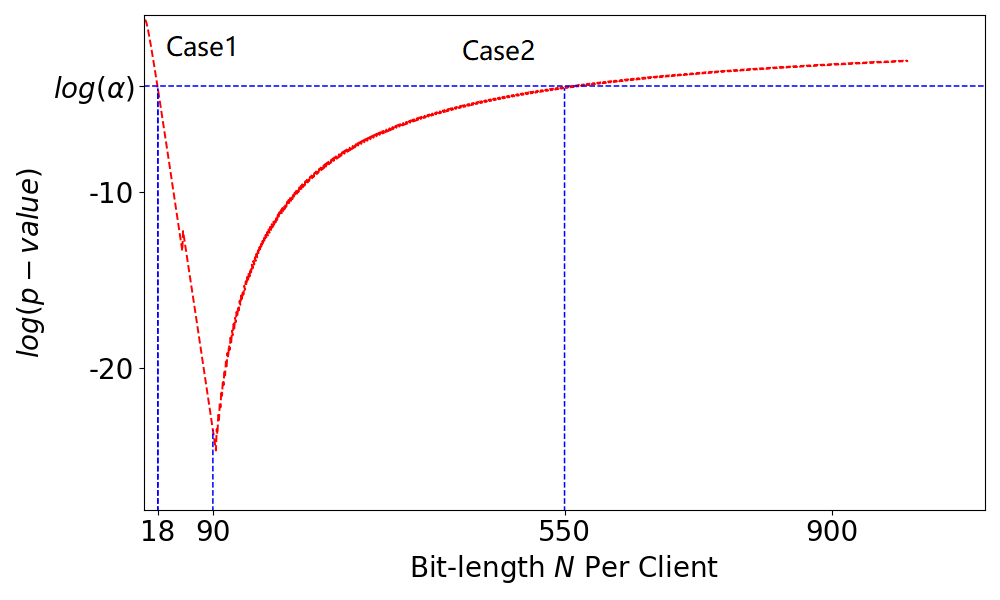}
	\caption{\textit{The optimal bit-length} and \textit{the acceptable range of watermark bit-length} that provide strong confidence of ownership verification. As shown in figure, $K=10, M=896$, and $\eta_F=0.98$ (we take by default) in case 1, the optimal bit-length for p-value is $N_{opt}=M/K=90$, for an acceptable level $\alpha=0.0001$, the acceptable range of watermark bit-length is from $N=18$ to $N=550$. }
	\label{fig:p-value}
	\vspace{-10pt}
\end{figure}

In case that a p-value is only required to be less than a given statistical significance level $\alpha$ e.g., 0.0001, we can determine the range of bit-length $N$ ($N_{max}$ and $N_{min}$) that can guarentee that p-value is lower than the given level $\alpha$.
Take an example as in Fig. \ref{fig:p-value}, if $K=10$ clients decide to embed watermarks into $M=896$ channels of model parameters, $N_{opt} = 90$, when $\alpha = 0.0001$, the corresponding $N_{max}$ is 550 and $N_{min}$ is 18. 

In short, analysis of Theorem \ref{thm: thm1} specifies \textit{the optimal bit-length} and \textit{the acceptable range of watermark bit-length} that allows reliable detection rates with sufficient significance to support ownership verification.  Moreover, it is shown that empirical results in Sect. \ref{sect:exper} are in accordance with the analysis elucidated in this section.




\subsection{Challenge B: Robustness of Watermarks in FedDNN}\label{Challenge B}

The \textit{robustness} of watermarks indicates whether the detection rate is persistent against various \textit{training strategies} and \textit{attacks} that attempt to remove the watermarks. 

We investigate the robustness of both the feature-based and backdoor-based watermarks in the FedDNN model. Particularly, we measure the detection rate and statistical significance of watermarks with/without training strategies and attacks to report the robustness. The measurement settings including impacting factors (training strategies and attacks), targeted watermarks  (feature-based watermarks in \textit{Normalization parameters} $\mathbf{W}_{\gamma}$ and convolution parameters $\mathbf{W}_C$ and backdoor-based watermarks), and metrics are summarized in Tab. \ref{tab: Robustness}:

\begin{table}[htbp]
  \renewcommand{\arraystretch}{1.2}

      \normalsize
    
     \centering
        
     \resizebox{0.49\textwidth}{!}{
      \begin{tabular}[l]{ccccc}
        \toprule
        \multicolumn{1}{c}{\multirow{2}{*}{ Impacting factor }} & \multicolumn{2}{c}{Watermarks}                   & \multicolumn{2}{c}{Metrics}                     \\ \cline{2-5}
\multicolumn{1}{c}{}                       & \multicolumn{1}{c}{feature-based} & \multicolumn{1}{c}{backdoor-based } & \multicolumn{1}{c}{Detection Rate }   & \multicolumn{1}{c}{ p-value} \\ \midrule
        
        Differential Privacy  & $\mathbf{W}_{\gamma}$ & \checkmark& \checkmark & \checkmark  \\
        Client Selection& $\mathbf{W}_{\gamma}$& \checkmark& \checkmark & \checkmark \\
        Defensive Aggregation  & $\mathbf{W}_{\gamma}$& \checkmark& \checkmark & \checkmark   \\\midrule
        Pruning Attack  & $\mathbf{W}_{\gamma}$, $\mathbf{W}_{C}$& \checkmark & \checkmark &  \\
        Fine-tuning Attack  & $\mathbf{W}_{\gamma}$, $\mathbf{W}_{C}$& \checkmark & \checkmark &  \\ 
       \bottomrule
      \end{tabular}
      } \caption{Reported investigation setting of robustness. }\label{tab: Robustness}
      \vspace{-7pt}
\end{table}


\noindent\textbf{Training Strategies:} In SFL, strategies like differential privacy\cite{wei2020federated}, defensive aggregation mechanism\cite{blanchard2017machine,yin2018byzantine, guerraoui2018hidden} and 
client selection\cite{communicationEfficient/mcMahan2017} are widely used for privacy, security and efficiency. 
Those training strategies modify the training processes of SFL, which may affect the detection rate i.e., the significance of watermarks:

\begin{itemize}
	\item To protect data privacy, \textit{differential privacy} mechanism\cite{DLDP_Abadi16} in SFL add noise to the local model of each client. 
	\item For defending the model poisoning attack\cite{blanchard2017machine, bagdasaryan2018backdoor}, \textit{defensive aggregation}\cite{blanchard2017machine,yin2018byzantine, guerraoui2018hidden} in SFL perform detect and filter some local models from each client.
	\item For communication efficiency, in each communication round, the server adopts certain \textit{client selection} strategies\cite{communicationEfficient/mcMahan2017} to pick up a random subset of clients and have their local models aggregated. Other clients do not need to upload local models until they are selected in future communication rounds. 
\end{itemize} 

\noindent\textbf{Removal Attack.} The attacker that steals the model may try to remove the watermarks while inheriting most model performance. Following previous DNN watermarking methods\cite{fan2021deepip, zhang2021deep,TurnWeakStrength_Adi2018arXiv}, we investigate watermark robustness under \textit{fine-tuning} and \textit{pruning} attacks (see Algo. \ref{alg:Removal attack} for pseudocodes).  

\begin{algorithm}[htbp]
	\caption{Removal attack}
	\label{alg:Removal attack}
	\textbf{Input:} Model $\mathbb{N}$, pruning rate $p$, additional training data $D_{add}$.
	\begin{algorithmic}[1]
		\Procedure{Pruning}{}
		\State Pruning the model $\mathbb{N}$ with $p$ pruning rate.
	    \EndProcedure
        \Procedure{Finetuning}{}
        \For{epochs in 50}
        \State Train the model $\mathbb{N}$ only in main classification task with additional training data $D_{add}$.
        \EndFor
        \EndProcedure

	\end{algorithmic}

\end{algorithm}

We adopt client-side watermark embedding method to investigate the  watermark robustness. Extensive experimental results in Sect. \ref{Robustness: Strategies} show that both backdoor-based and feature-based watermarks can be reliably detected with high significance (p-value less than 2.89$e^{-15}$ is guaranteed).

\section{Implementation of FedDNN Ownership Verification}\label{sect: Embed}

Sect. \ref{sect:FedIPR-fw} proposes the FedIPR framework, which allows clients to independently embed secret watermarks into the model and verify whether the designated watermarks exist in the model in question. We illustrate below a specific implementation of FedIPR framework with algorithm pseudocodes given in Algo. \ref{alg:generation},  \ref{alg:training embedding} and  \ref{alg:wmdet}.

\subsection{Watermark Generation} 
For a client $k$ in the federated learning system, the watermarks adopted to mark trained FedDNN model include backdoor-based watermarks $\mathbf{T}_k$ 
and feature-based watermarks $(\mathbf{B}_k$, $\theta_k)$. Those watermarks are initialized and kept in secret as shown in Algo. \ref{alg:generation}. 

\begin{algorithm}[htbp]
	\caption{Generation $\mathcal{G}()$ of Watermarks}\label{alg:generation}
	\begin{algorithmic}[1]
		\Procedure{watermark Generation}{}
		\For{client $k$ in $K$ clients}
		\State Initialize $(\mathbf{B}_k$, $\theta_k) = \{\mathbf{S}_k, \mathbf{E}_k\}$.
		\State Encode $\mathbf{B}_k$ into binary string.
		\State Initialize $\mathbf{T}_k = \{ (\mathbf{X}_{\mathbf{T}_k}^1, \mathbf{Y}_{\mathbf{T}_k}^1), \cdots, (\mathbf{X}_{\mathbf{T}_k}^{N_{\mathbf{T}}}, \mathbf{Y}_{\mathbf{T}_k}^{N_{\mathbf{T}}})  \}$.  
		\EndFor
		\State \Return $\{(\mathbf{B}_k, \theta_k, \mathbf{T}_k)\}_{k=1}^{k=K}$
		
		\EndProcedure
	\end{algorithmic}
	
\end{algorithm}
\vspace{-5pt}

As illustrated in Algo. \ref{alg:training embedding}, the client-side FedDNN \textit{embedding process} $\mathcal{E}()$ minimizes the weighted combined loss of main task and two regularization terms to embed watermarks $\mathbf{T}_k$ and  $\mathbf{B}_k$ respectively. 


\subsection{Backdoor-based Watermark Embedding and Verification}\label{section: Backdoor-based embedding}
To embed backdoor-based watermarks $\mathbf{T}_k = \{ (\mathbf{X}_{\mathbf{T}_k}^1, \mathbf{Y}_{\mathbf{T}_k}^1), \cdots, (\mathbf{X}_{\mathbf{T}_k}^{N_{\mathbf{T}}}, \mathbf{Y}_{\mathbf{T}_k}^{N_{\mathbf{T}}})  \}$, the model owner trains the model with an additional backdoor training task 
where the loss function of backdoor training $L_{\mathbf{T}} (\mathbf{W}^t)$ is defined with cross entropy (CE) loss: 
\begin{equation}
	L_{\mathbf{T}}(\mathbf{W}^t) = CE(\mathbf{Y}_{\mathbf{T}_k}, \mathbb{N}(\mathbf{X}_{\mathbf{T}_k})).
\end{equation}

\noindent\textbf{Adversarial Samples as Triggers.}
In our FedIPR scheme, we adopt adversarial samples as the triggers. Basically, adversarial samples 
$(\mathbf{X}_{\mathbf{T}},\mathbf{Y}_{\mathbf{T}})$ are generated from original data $(\mathbf{X},\mathbf{Y})$ with Projected Gradient Descent (PGD)\cite{madry2017towards}. 
Our backdoor-based watermark scheme adopts those backdoor samples as the trigger input during both training time and inference time. 


Each backdoor-based watermarks $\mathbf{X}_{\mathbf{T}}$ is verified, provided that for the  input $\mathbf{X}_{\mathbf{T}}$, the model outputs the designated label, i.e., $\mathbb{N}(\mathbf{X}_{\mathbf{T}}) = \mathbf{Y}_{\mathbf{T}}$.

\subsection{Feature-based Watermark Embedding and Verification} \label{section: Feature-based embedding}
In FedIPR, each client $k$ chooses its own $(\mathbf{B}_k, \mathbf{\theta}_k)$ as the feature-based watermarks, which are embedded with a regularization term $L_{\mathbf{B}_k, \theta_k}$ along with main task loss.


\noindent\textbf{Approach of FedIPR.} 
FedIPR proposes that each client embeds its own watermarks $\mathbf{B}_k$ with secret parameters $\theta_k = (\mathbf{S}_k, \mathbf{E}_k)$:
\begin{equation}\label{eq:unified-R}
	L_{\mathbf{B}_k, \theta_k}\big(\mathbf{W}^t\big) = L_{\mathbf{B}_k} (\mathbf{S}_k, \mathbf{W}^t, \mathbf{E}_k), 
\end{equation}
whereas the secret watermarking parameters $\theta_k = (\mathbf{S}_k, \mathbf{E}_k)$ are only known for client $k$, and FedIPR proposes to embed the watermarks into the \textit{normalization layer scale parameters} of the convolution block, i.e.,  $\mathbf{S}_k(\mathbf{W}) = \mathbf{W}_{\gamma}=\{\gamma_1, \cdots, \gamma_C\}$, where $C$ is the number of normalization channels in $\mathbf{W}_{\gamma}$.

We adopt a secret embedding matrix $\mathbf{E}_k \in \theta_k = (\mathbf{S}_k, \mathbf{E}_k)$ to embed and extract watermarks, the distance between targeted watermarks and extracted watermarks is implemented as following regularization term:

\begin{equation}
\begin{aligned}
	L_{\mathbf{B}_k, \theta_k}\big(\mathbf{W}^t\big) = & L_{\mathbf{B}_k}\big(\mathbf{W}^t_{\gamma}\mathbf{E}_k, \mathbf{B}_k \big) \\ = &\text{HL}\Big( \mathbf{B}_k, \mathbf{\tilde{B}}_k \Big) =  \sum^{N}_{j=1} \max( \mu - b_j t_j, 0 ),  
\end{aligned}
\end{equation}
where we note $\mathbf{\tilde{B}}_k = \mathbf{W}^t_{\gamma}\mathbf{E}_k$ as extracted watermarks and we implement the regularization term as \textit{hinge-like} loss $\text{HL}()$ on the target watermarks $\mathbf{B}_k= (t_1,\cdots,t_N) \in \{0,1\}^N$ and extracted watermarks $\mathbf{\tilde{B}}_k = (b_1, \cdots, b_N) \in \{0,1\}^N$, and $\mu$ is the parameter of hinge loss. 

\begin{algorithm}[htbp]
	\caption{Embedding Process $\mathcal{E}()$ of Watermarks}\label{alg:training embedding}
	\begin{algorithmic}[1]
		\State Each client k with its own watermark tuple $(\mathbf{B}_k, \theta_k, \mathbf{T}_k)$ 
		\For{ communication round $t$ }
		\State The server distributes the global model parameters $\mathbf{W}^t$ to each clients and randomly selects $cK$ out of $K$ clients.
		\State \textbf{Local Training:}
		\For{$k$ in selected $cK$ of $K$ clients } 
			\State Sample mini-batch of $m$ training samples 
			$\mathbf{X}\{\mathbf{X}^{(1)}$, $\cdots$, $\mathbf{X}^{(m)}$\} and targets $\mathbf{Y}\{\mathbf{Y}^{(1)}$, $\cdots$, $\mathbf{Y}^{(m)}$\}.
			\If{Enable backdoor-based watermarks}
			    \State Sample $t$ samples $\{\mathbf{X}_{\mathbf{T}_k}^{(1)}$, $\cdots$, $\mathbf{X}_{\mathbf{T}_k}^{(t)}$\} ,  \{$\mathbf{Y}_{\mathbf{T}_k}^{(1)}$, $\cdots$, $\mathbf{Y}_{\mathbf{T}_k}^{(t)}\}$ from trigger set $(\mathbf{X}_{\mathbf{T}_{k}}, \mathbf{Y}_{\mathbf{T}_{k}})$
				\State Concatenate $\mathbf{X}$ with $\{\mathbf{X}_{\mathbf{T}_k}^{(1)}$, $\cdots$, $\mathbf{X}_{\mathbf{T}_k}^{(t)}$\} , $\mathbf{Y}$ with \{$\mathbf{Y}_{\mathbf{T}_k}^{(1)}$, $\cdots$, $\mathbf{Y}_{\mathbf{T}_k}^{(t)}$\}.
			\EndIf			
			\State Compute cross-entropy loss $L_c$ using $\mathbf{X}$ and $\mathbf{Y}$\Comment{Batch poisoning approach is adopted, thus $\alpha_l=1$, $L_c = L_{D_K}+ L_{\mathbf{T}_k}$}. 
			\For{layer $l$ in targeted layers set $\mathsf{L}$}
				\State Compute feature-based regularization term $L_{\mathbf{B}_k,\theta_k}^l$ using $\theta_k$ and $\mathbf{W}^l$
			\EndFor
			\State $L_{\mathbf{B}_k,\theta_k} \gets \sum_{l \in \mathsf{L}} L_{\mathbf{B}_k,\theta_k}^l$
			\State $L_k$ = $L_c$ + $\beta_kL_{\mathbf{B}_k,\theta_k}$
			\State Backpropagate using $L_k$ and update $\mathbf{W}^t_k$
		\EndFor 
		\State \textbf{Server Update:}
        \State Aggregate local models $\{\mathbf{W}^t_k\}_{k=1}^{K}$ with $\texttt{FedAvg}$ algorithm
		\EndFor
	\end{algorithmic}
\end{algorithm}

In SFL, we implement feature-based watermark embedding for two different network architectures including convolution neural network (CNN) 
and transformer-based neural network.

\subsubsection{Feature-based Watermarks in CNN}
\begin{figure}[htbp]
	\centering
	\includegraphics[width=3.4in]{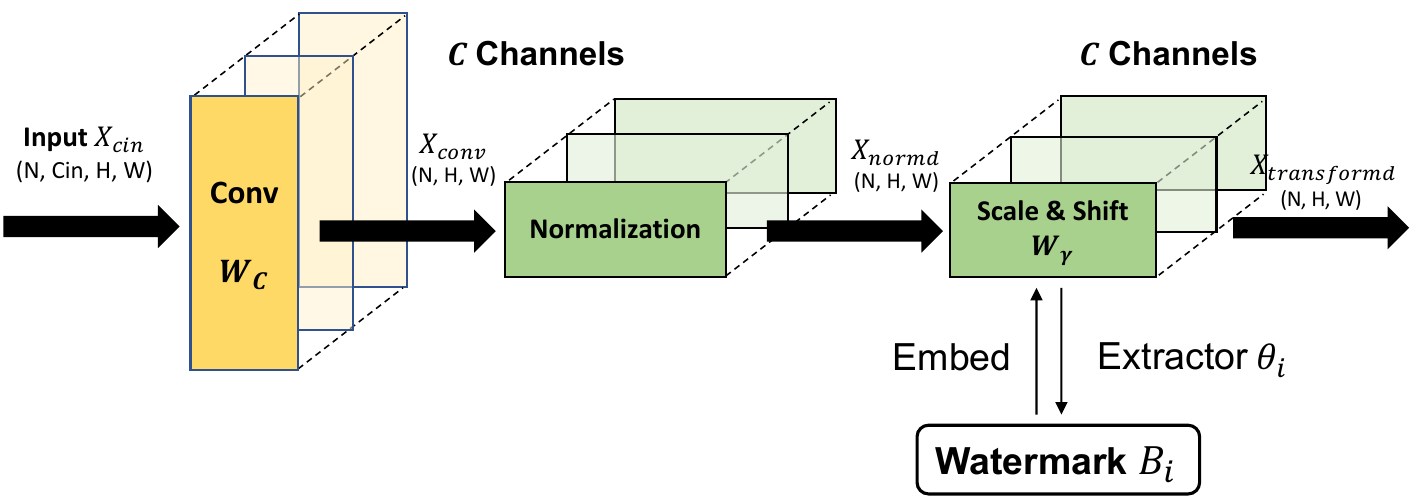}
	\caption{Layer structure of a convolution layer: normalization layer weights $\mathbf{W}_{\gamma}$ (in green) are used to embed watermarks, and the watermarks extracted in a white-box manner with a secret embedding  matrix. }
	\label{fig:white-box}
	\vspace{-10pt}
\end{figure}

As illustrated in Fig. \ref{fig:white-box}, for a convolution neural network $\mathbb{N}(\mathbf{W})$, we may choose the convolution kernel weights $\textbf{W}_C$ or 
the normalization layer weights $\textbf{W}_{\gamma}$ to embed feature-based watermarks. 

Fan et al.\cite{fan2019rethinking, fan2021deepip} has reported that normalization layer weights $\mathbf{W_\gamma} = (\gamma_1, \cdots, \gamma_{C}) \in \{-1, +1\}^C$ are suitable model parameters to embed robust binary watermark strings as follows: 
\begin{equation}
	O(x^p_i) = \gamma_i * x^p_i + \beta_i, 
\end{equation}
in which $x^p_i$ is the model parameters in channel $i$ and $\beta_i$ is the offset parameter of normalization (see \cite{fan2019rethinking, fan2021deepip} and Appendix C for details). 

In FedIPR, we choose $\mathbf{W_\gamma}$ to embed feature-based watermarks for  robust performance, which is reported in Sect. \ref{sect:exper}. For ablation study, we  compare the robustness of watermarks in the normalization layers and convolution layers in Fig. \ref{fig:removal}, the results show that watermarks in the normalization layers are more persistent against removal attacks. 


\subsubsection{Feature-based Watermarks in Transformer-based Networks}
\begin{figure}[htbp]
\vspace{-5pt}
	\centering
	\includegraphics[width=3in]{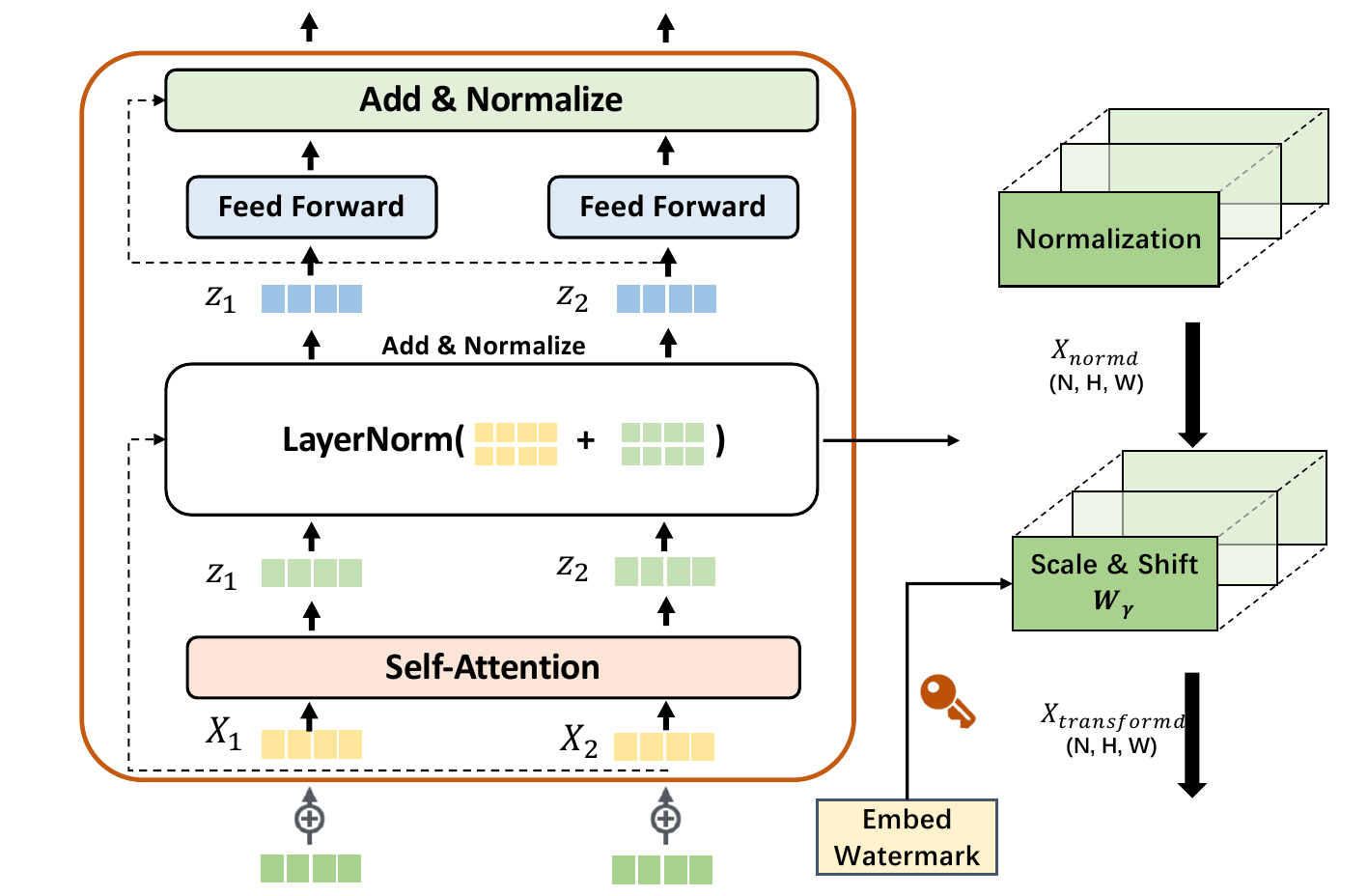}
	\caption{Layer structure of an encoder block: normalization layer weights $\mathbf{W_\gamma}$ (in green) are used to embed feature-based watermarks which are extracted in white-box manner. }
	\label{fig:transformer_white_box}
\vspace{-15pt}
\end{figure}

As illustrated in Fig. \ref{fig:transformer_white_box}, feature-based watermarks can also be applied to transformer-based network. 
A transformer encoder block\cite{vaswani2017attention} is organized with a \textit{Layer-Normalization layer} (the mean output value is normalized in the channel direction, which has an obvious effect for 
accelerating the convergence performance). The same to normalization in CNN, 
the process is controlled by parameters $\mathbf{W}_{\gamma}$ in the \textit{Layer-Normalization layer}.

In FedIPR, we choose $\mathbf{W_\gamma}$ to embed the feature-based watermarks, 
such that the watermarks can be persistent in the model architecture.

\subsection{Ownership Verification with Watermarks}

Once the model is plagiarized by unauthorized party, the model owner can call experiments of ownership verification. As Algo. \ref{alg:wmdet} and Fig. \ref{fig:verification} illustrates, for the backdoor-based watermark verification, the model owner query the API with designed triggers, and for feature-based verification, it detected the feature-based watermarks embedded in the  normalization layer according to the sign of according channel parameters. Combined with both backdoor-based and feature-based watermarks, the statistical significance of ownership verification is obtained. 

\begin{algorithm}[htbp]
	\caption{Ownership Verification $\mathcal{V}_B()$ and $\mathcal{V}_W()$}
	\label{alg:wmdet}
	\textbf{Input:} API $\mathbb{N}()$ offered by adversaries, Triggers  $(\mathbf{X}_{\mathbf{T}},\mathbf{Y}_{\mathbf{T}})$ provided by owner;
	Model weights $\mathbf{W}$ of model, secret parameters $\mathbf{\theta}= (\mathbf{S}, \mathbf{E})$ and target watermarks $\mathbf{B}$ provided by user.
	\begin{algorithmic}[1]
		\Procedure{watermark detection}{}
		\State Input the backdoor $\mathbf{X}_{\mathbf{T}}$ into model $\mathbb{N}$ to derive the classification label $\mathbb{N}(\mathbf{X}_{\mathbf{T}})$ 
		\State Match $\mathbb{N}(\mathbf{X}_{\mathbf{T}})$ with target backdoor label $\mathbf{Y}_{\mathbf{T}}$ 
		\State Compute the backdoor detection rate $\eta_T = \mathcal{V}_B(\mathbf{X}_{\mathbf{T}}$, $\mathbf{Y}_{\mathbf{T}}$, $\mathbb{N})$
		\State $\tilde{\mathbf{B}} \gets sgn(\mathbf{S}(\mathbf{W})\mathbf{E})$
		\State Match decoded $\tilde{\mathbf{B}}$ with target watermark $\mathbf{B}$
		\State Compute the watermark detection rate $\eta_F = \mathcal{V}_W(\mathbf{W}, \mathbf{B}, \mathbf{\theta})$
		\State Compute p-value corresponding to the detection rate $\eta_T$ and $\eta_F$
		\EndProcedure
	\end{algorithmic}
	\textbf{Output:} p-value corresponding to the detection rate $\eta_T$ and $\eta_F$.
\end{algorithm}
\vspace{-10pt}

\section{Experimental Results}\label{sect:exper}

This section illustrates the empirical study of  the proposed FedIPR in terms of \textit{fidelity}, \textit{significance} and \textit{robustness} of watermarks.
Superior detection performances of both backdoor-based watermarks
and feature-based watermarks  in the presence of \textbf{Challenge A and B}
demonstrate that FedIPR provides a reliable and robust scheme for FedDNN ownership verification.

\begin{table*}[htbp]
  \renewcommand{\arraystretch}{1}

     \centering
        
     \resizebox{0.96\textwidth}{!}{
      \begin{tabular}[l]{cccccccc}
        \toprule
        \multirow{2}{*}{ Architecture }& \multirow{2}{*}{ Datasets }& \multicolumn{2}{c}{Watermarks}                   & \multicolumn{3}{c}{Metrics}           &   \multirow{2}{*}{ Freeriders }       \\ \cmidrule(r){3-4} \cmidrule(r){5-7}
 &  & Feature-based & Backdoor-based  & \multicolumn{1}{c}{Fidelity}   & Significance & Robustness & \\ \midrule
        
        AlexNet & CIFAR10 & Fig. \ref{fig:reliability}, \ref{fig:thm} & Tab. \ref{tab:reliablity blackbox} & Fig. \ref{fig:fidelity} &  Tab. \ref{tab: p_value_feature}, \ref{tab: p_value_backdoor}   & Fig. \ref{fig:DP}, \ref{fig:Fraction}, \ref{fig:removal}. Tab. \ref{tab: p-value_DP}, \ref{tab: p-value_Selection}, \ref{tab: Bulyan_blackbox} & Fig. \ref{fig:freerider}\\
        ResNet18 & CIFAR100 & Fig. \ref{fig:reliability} & Tab. \ref{tab:reliablity blackbox} & Fig. \ref{fig:fidelity} &  Tab. \ref{tab: p_value_feature}, \ref{tab: p_value_backdoor} & Fig. \ref{fig:DP}, \ref{fig:Fraction}, \ref{fig:removal}. Tab. \ref{tab: p-value_DP}, \ref{tab: p-value_Selection} & Fig. \ref{fig:freerider} \\
        DistlBERT  & SST2, QNLI & Fig. \ref{fig:reliability} &  & Fig. \ref{fig:fidelity} & Tab. \ref{tab: p_value_feature} &   &
        \\ \bottomrule
      \end{tabular}
      } \caption{Reported experiment results under different settings for proposed  feature-based watermarks and backdoor-based watermarks. }\label{tab: Setting}
\end{table*}
\vspace{-10pt}

\subsection{Experiment Settings}\label{appendix: EXP Setting}
This subsection illustrates the settings of the empirical study of our FedIPR framework, which is summarized in Tab. \ref{tab: Setting}.

\noindent\textbf{DNN Model Architectures.}  The deep neural network architectures we investigated include the well-known AlexNet, ResNet-18, and DistilBERT\cite{sanh2019DistilBERT}. 
For convolution neural networks, feature-based binary watermarks are embedded into 
normalization scale weights $\mathbf{W}_{\gamma}$ of multiple 
convolution layers in AlexNet and ResNet-18; for tranformer-based neural networks, feature-based watermarks are embedded into \textit{Layer-Normalization} scale weights $\mathbf{W}_{\gamma}$ of multiple encoders in DistilBERT. The detailed model architectures are shown in Appendix  C. 
\\\textbf{Datasets.} For image classification tasks, FedIPR is evaluated on CIFAR10 and CIFAR100 datasets, and for natural language processing tasks, FedIPR is evaluated on GLUE benchmark including SST2 and QNLI datasets.
\\\textbf{Federated Learning Settings.} We simulate a horizontal federated learning setting in which clients upload local models in each communication round, and the server adopts \texttt{Fedavg}\cite{communicationEfficient/mcMahan2017} algorithm to aggregate the local models. 
Detailed experimental  hyper-parameters to conduct FedIPR are listed in Appendix C. Our source codes for implementation are available at \href{https://github.com/purp1eHaze/FedIPR}{https://github.com/purp1eHaze/FedIPR}.

\subsection{Evaluation Metrics} \label{subsect:exp-metrics}
Following previous DNN watermarking methods\cite{fan2021deepip, zhang2021deep,TurnWeakStrength_Adi2018arXiv}, to measure the \textit{fidelity}, \textit{watermark significance} and \textit{robustness} of the proposed FedIPR framework, 
we apply a set of metrics as below: 
\\\textbf{Fidelity.} We use classification accuracy on the main task $Acc_{main}$ as the metrics for \textit{fidelity}. It is expected classification accuracy should not be degraded by watermarks embedded in FedDNN (see Sect. \ref{subsect: fidelity} for experimental results).
\\\textbf{Watermark Significance.} The \textit{watermark significance} measures the statistical significance that the watermarks can provide to rightfully support the ownership verification. We treat the watermark detection as a hypothesis testing process (illustrated in Algo. \ref{alg:wmdet}), the watermark significance is calculated in two phases:
\begin{itemize}
    \item \textbf{Watermark Detection Rate.} In the first phase, the watermark detection could be formulated as a classification problem which returns the watermark detection rate $\eta_{T}$ and $\eta_{F}$ (defined in Sect. \ref{subsect:FedDNN-form}). 
    \item \textbf{Statistical Significance (p-value).} In the second phase, we further adopt the p-value of hypothesis testing to quantify the statistical significance of watermarks. 
\end{itemize}

\noindent\textbf{Robustness.} We measure the detection rate and statistical significance of watermarks with/without training strategies and attacks to report the robustness. 


\subsection{Fidelity}\label{subsect: fidelity}

We compare the main task  performance $Acc_{main}$ of FedIPR against \texttt{FedAvg} to report the fidelity of the proposed FedIPR. 
In four different training tasks, varying number (from 10 to 100) of clients may decide to embed different bit-length of backdoor-based watermarks (20 to 100 per client) and  feature-based watermarks (50 to 500 bits per client). 

\begin{figure}[h]
    \vspace{-10pt}
	\centering
	\begin{subfigure}{0.24\textwidth}
		\centering
		\includegraphics[keepaspectratio=true, width=120pt]{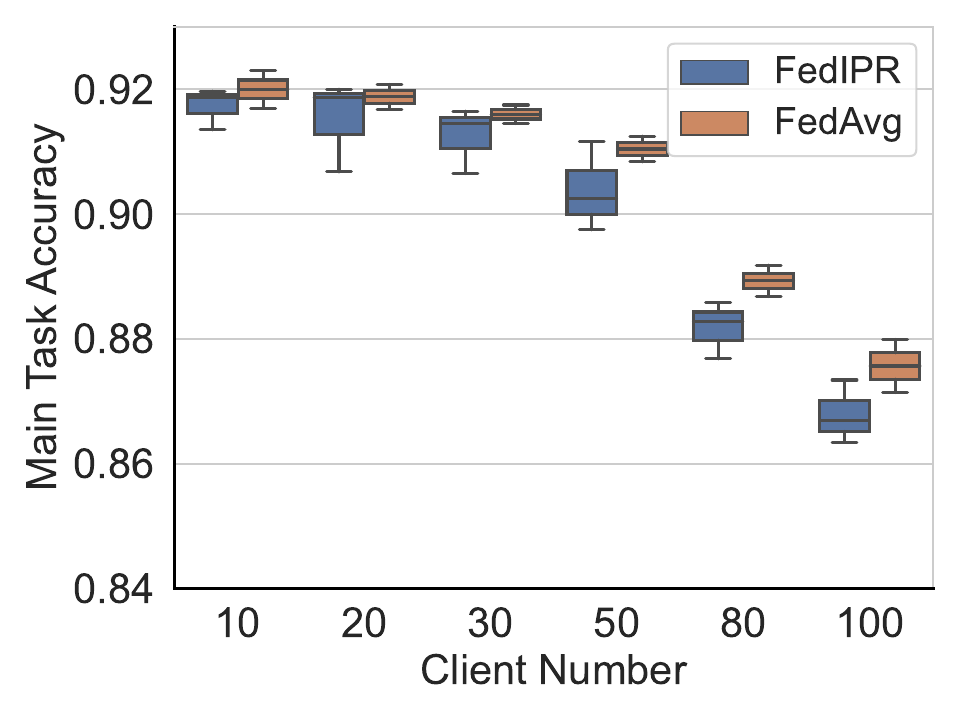}
		\subcaption{\small AlexNet with CIFAR10}
	  \end{subfigure}
	 \begin{subfigure}{0.24\textwidth}
		\centering
		\includegraphics[keepaspectratio=true, width=120pt]{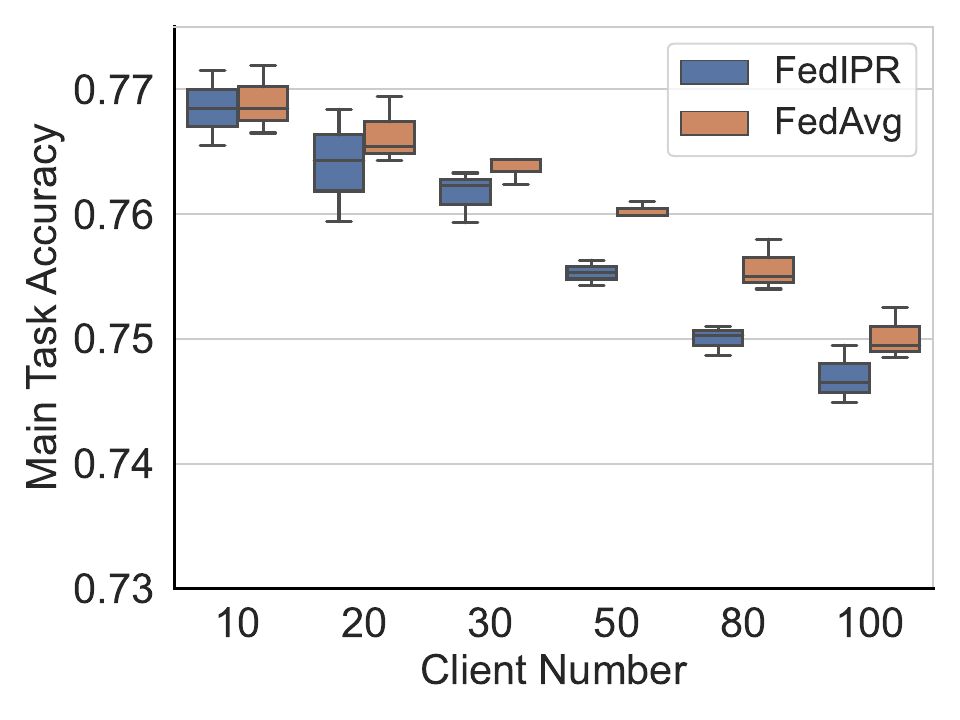}
		\subcaption{\small ResNet18 with CIFAR100}
	  \end{subfigure}
	\begin{subfigure}{0.24\textwidth}
		\centering
		\includegraphics[keepaspectratio=true, width=120pt]{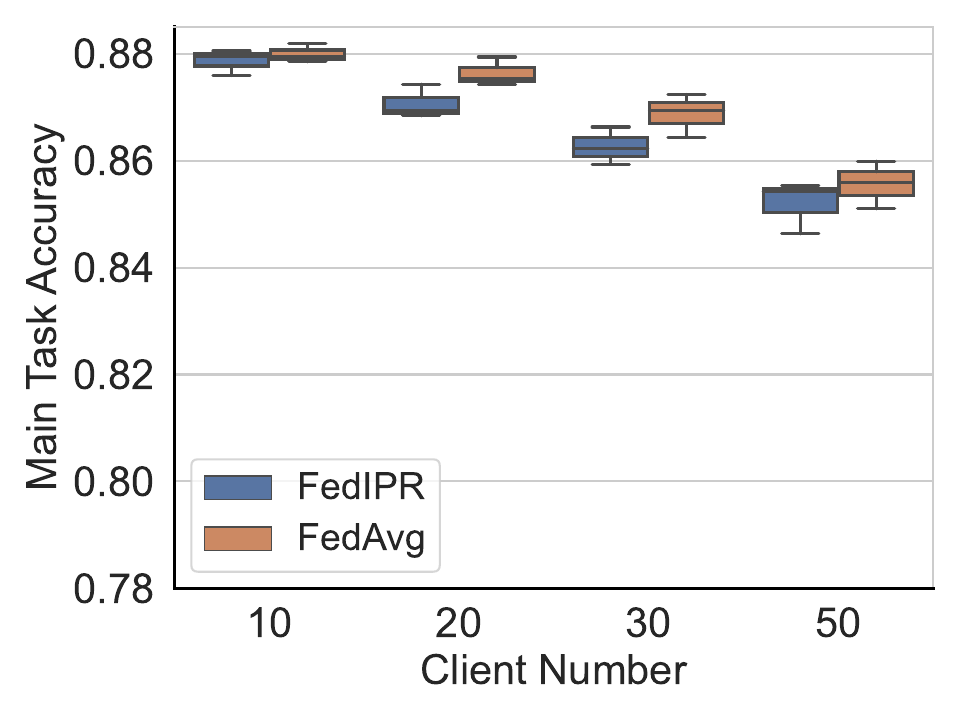}
		\subcaption{\small DistilBERT with SST2}
	  \end{subfigure}
	\begin{subfigure}{0.24\textwidth}
		\centering
		\includegraphics[keepaspectratio=true, width=120pt]{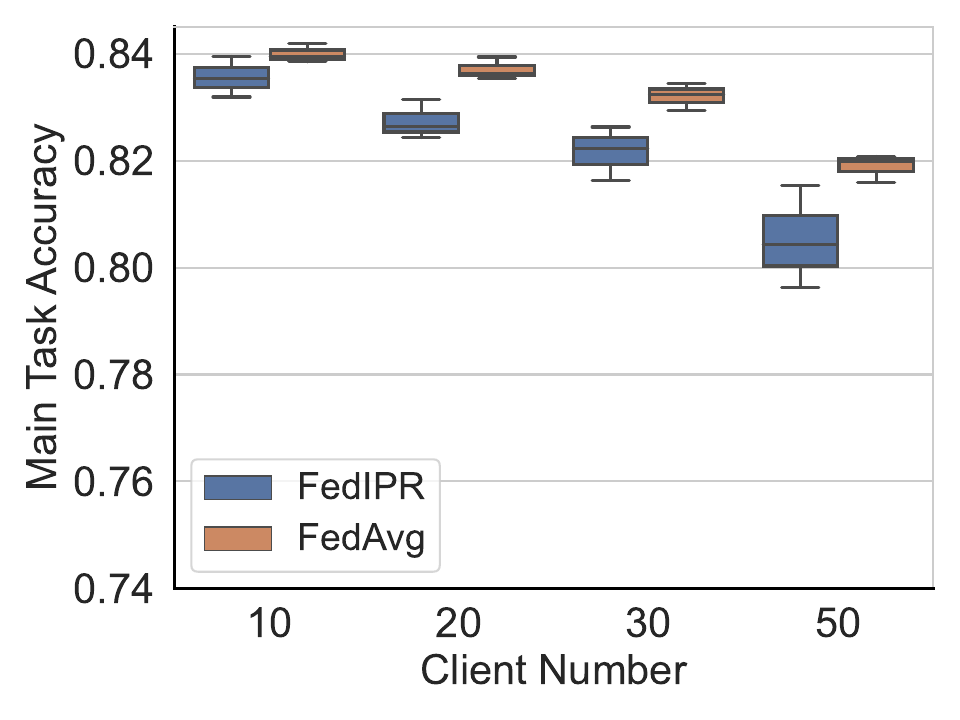}
		\subcaption{\small DistilBERT with  QNLI}
	  \end{subfigure}
	
	\caption{ 
	Figure (a)-(d), respectively, illustrate the main task accuracy $Acc_{main}$ in image and text classification tasks with varying number $K$ of total clients (from 10 to 100), 
	the results are based on cases of varying settings of feature-based and backdoor-based  watermarks, the main task accuracy $Acc_{main}$ of FedIPR has  slight dropped (not more than 2\%) compared to \texttt{FedAvg} scheme.} 
	\label{fig:fidelity}
	\vspace{-5pt}
	\end{figure}

Fig. \ref{fig:fidelity} (a)-(d) present the worst drop of classification accuracy $Acc_{main}$.  
It is observed that under various watermarking settings, slight model performance drop (not more than 2\% as compared with that of \texttt{Fedavg}) is observed for 
four seperated tasks. 

Tab. \ref{tab_fidelity} reports the main task accuracy with backdoor-based watermarking and feature-based watermarking respectively, the results show that the model performance drop (not more than 2 percent) is mainly caused by feature-based watermarking, a possible reason is that the regularization of feature-based watermarks may lead model parameters $\mathbf{W}$ to converge in a subspace of total space.

\begin{table}[htbp]
\renewcommand{\arraystretch}{1.2}
\centering
\resizebox{0.49\textwidth}{!}{
\begin{tabular}{cccccc}
\toprule
\multirow{2}{*}{Dataset} &  \multicolumn{2}{c}{Backdoor-based } &\multicolumn{2}{c}{Feature-based } & \multirow{2}{*}{Bassline}  \\

\cmidrule(r){2-3}  \cmidrule(r){4-5} 
                              & $N_{\mathbf{T}}=$ 50& $N_{\mathbf{T}}=$100& $N=$50 &$N=$ 100 &\\ 
\midrule
CIFAR10                & 91.69\% $\pm$ 0.15\%  & 91.53\% $\pm$ 0.18\% & 90.89\% $\pm$ 0.23\%   & 90.62\% $\pm$ 0.29\% & 91.72\% $\pm$ 0.12\% \\
CIFAR100                & 76.47\% $\pm$ 0.26\%  & 76.32\% $\pm$ 0.13\%  & 75.12\% $\pm$ 0.27\%  & 74.29\% $\pm$ 0.33\%  & 76.52\% $\pm$ 0.23\% \\ 
\bottomrule 
\end{tabular}}
\caption{In the FedIPR setting with 20 clients, table shows the main task accuracy $Acc_{main}$ with different watermarking methods, the CIFAR10 and CIFAR100 datasets are correspondly trained with AlexNet and ResNet. 
}\label{tab_fidelity}
\end{table}


\subsection{Watermark Significance} \label{subsect: reliability}

	\begin{figure}[htbp]
	\centering
	\begin{subfigure}{0.24\textwidth}
		\centering
		\includegraphics[keepaspectratio=true, width=130pt]{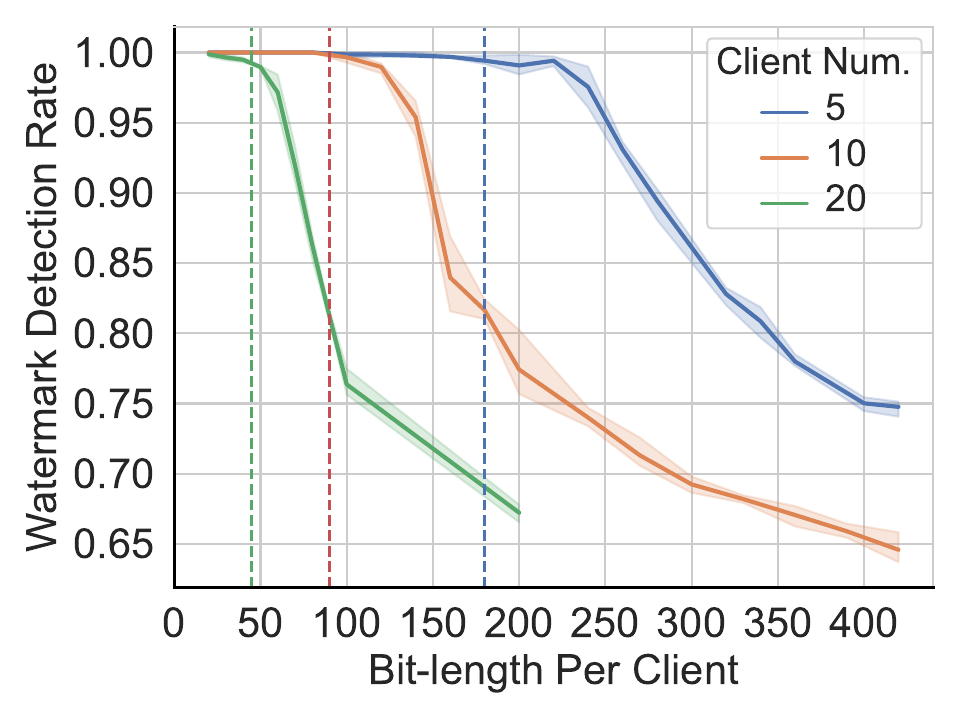}
		\subcaption{\small AlexNet with CIFAR10}
	  \end{subfigure}
	 \begin{subfigure}{0.24\textwidth}
		\centering
		\includegraphics[keepaspectratio=true, width=130pt]{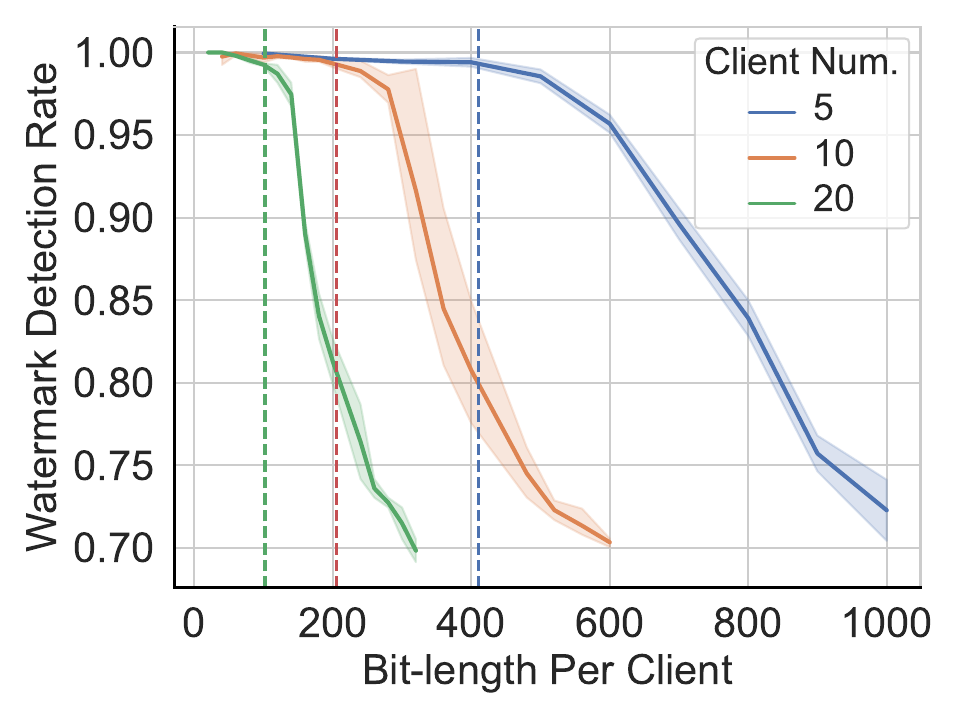}
		\subcaption{\small ResNet18 with CIFAR100}
	  \end{subfigure}
	\begin{subfigure}{0.24\textwidth}
		\centering
		\includegraphics[keepaspectratio=true, width=130pt]{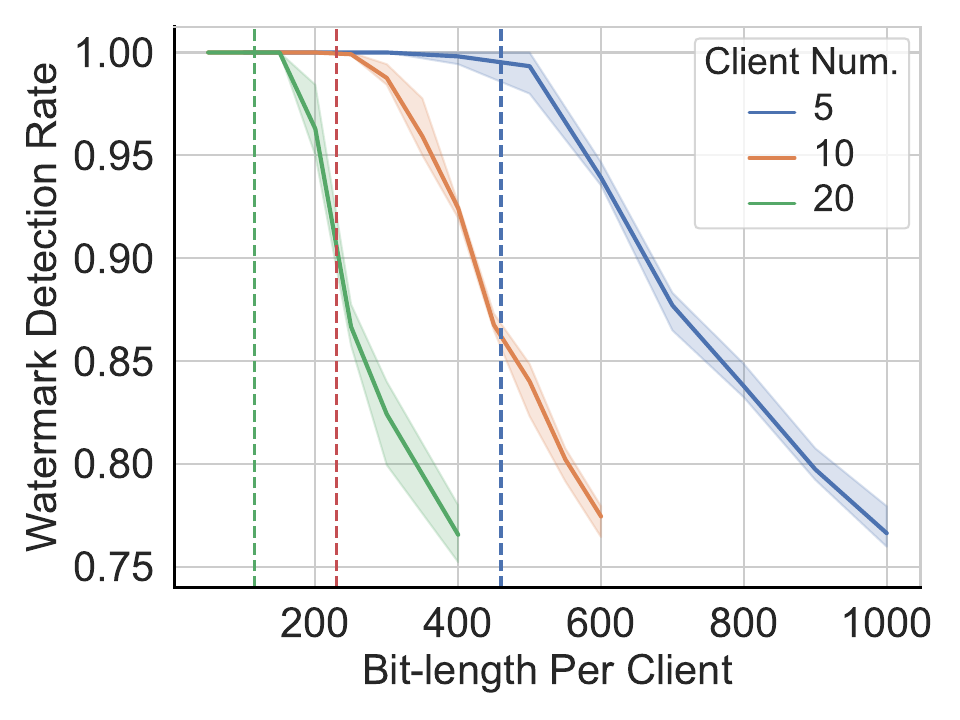}
		\subcaption{\small DistilBERT with SST2}
	  \end{subfigure}
	\begin{subfigure}{0.24\textwidth}
		\centering
		\includegraphics[keepaspectratio=true, width=130pt]{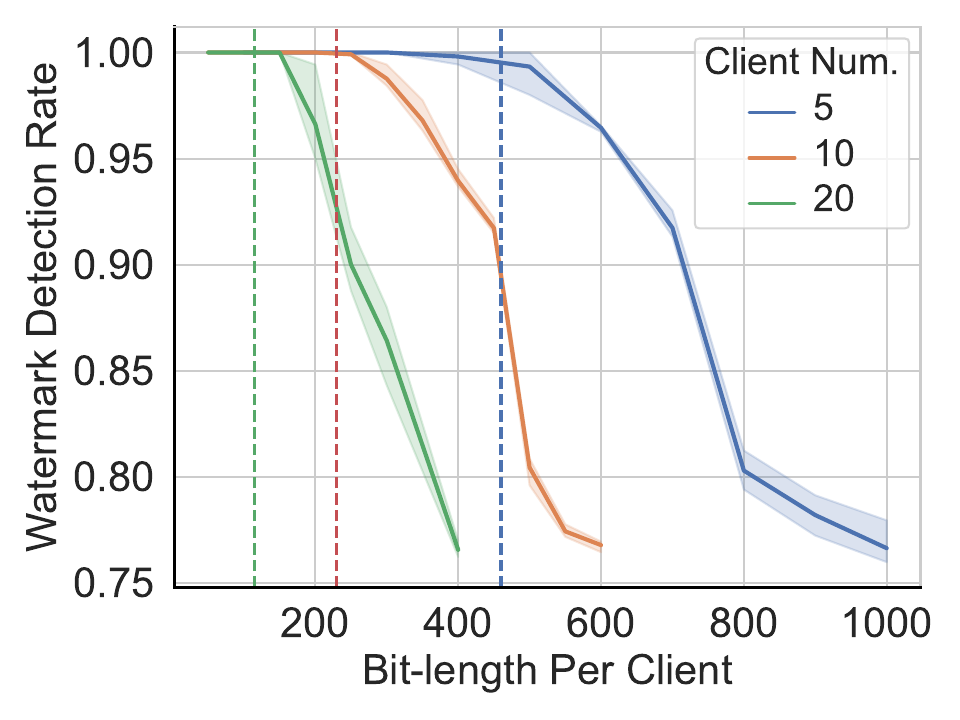}
		\subcaption{\small DistilBERT with QNLI}
	  \end{subfigure}
	
	\caption{ 
	Figure (a)-(d), respectively, illustrate the feature-based watermark detection rate $\eta_F$ in image and text classification tasks with varying bit-length per client, in SFL with $K$ = 5, 10, 20 clients, the dot  vertical line indicates $M/K$, which is the theoretical bound given by Theorem \ref{thm: thm1}. }
	\label{fig:reliability}
	\vspace{-5pt}
	\end{figure}
We present the watermark detection rate and statistical significance to report the watermark significance of the proposed FedIPR framework. 

\begin{figure}[h] 
	        	
	\centering
	 \begin{subfigure}{0.24\textwidth}
		\centering
		\includegraphics[keepaspectratio=true, width=130pt]{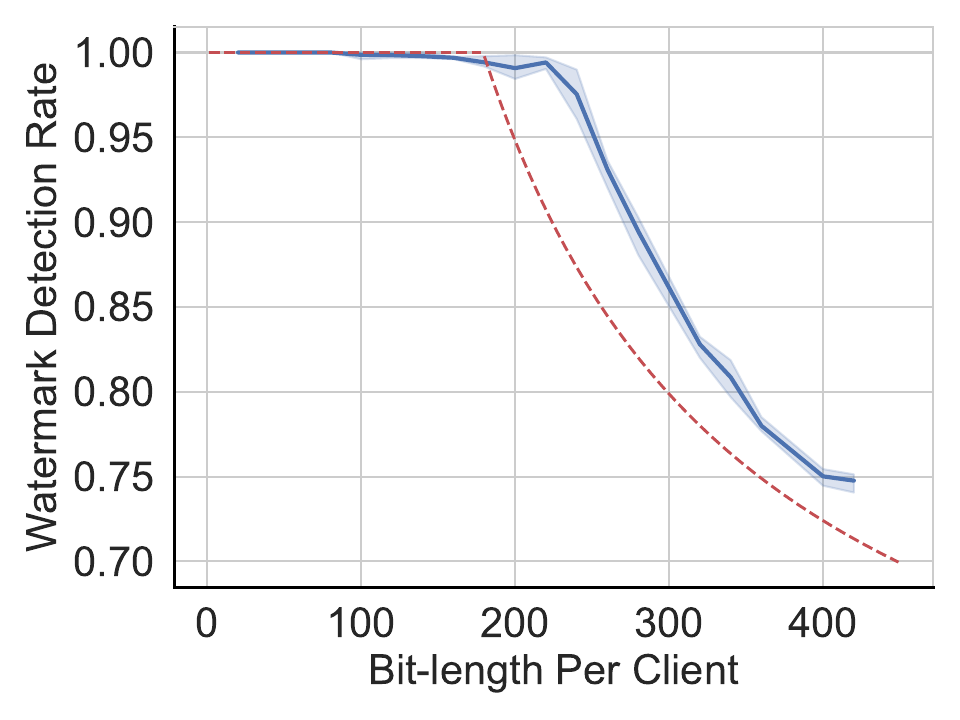}
		\subcaption{\small AlexNet on CIFAR10}
	  \end{subfigure}
	 \begin{subfigure}{0.24\textwidth}
		\centering
		\includegraphics[keepaspectratio=true, width=130pt]{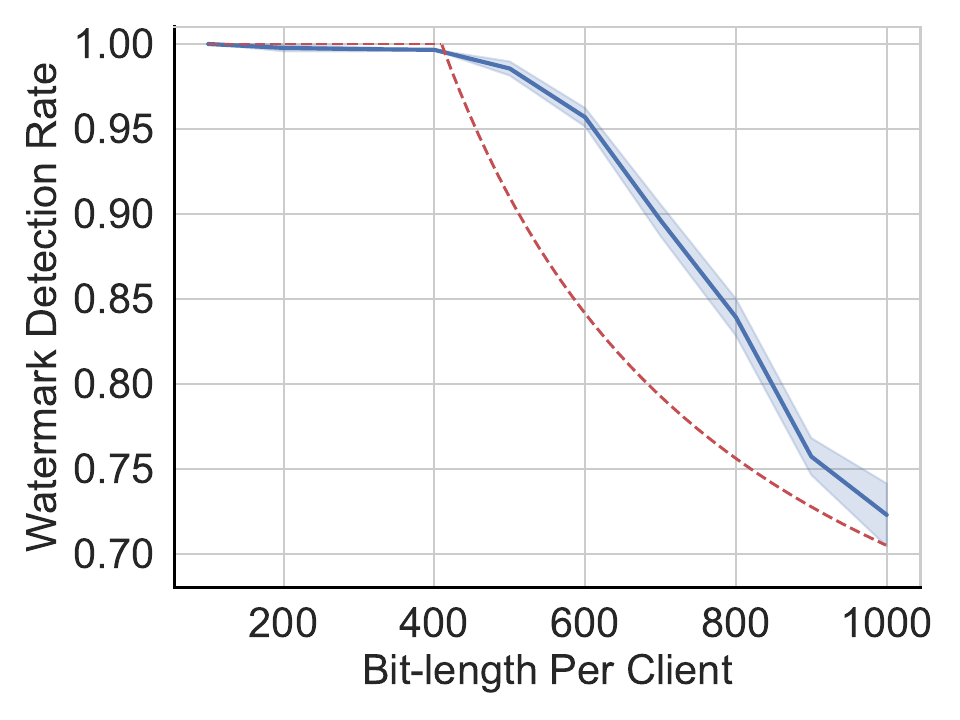}
		\subcaption{\small ResNet on CIFAR100}
	  \end{subfigure}
	  \begin{subfigure}{0.24\textwidth}
		\centering
		\includegraphics[keepaspectratio=true, width=130pt]{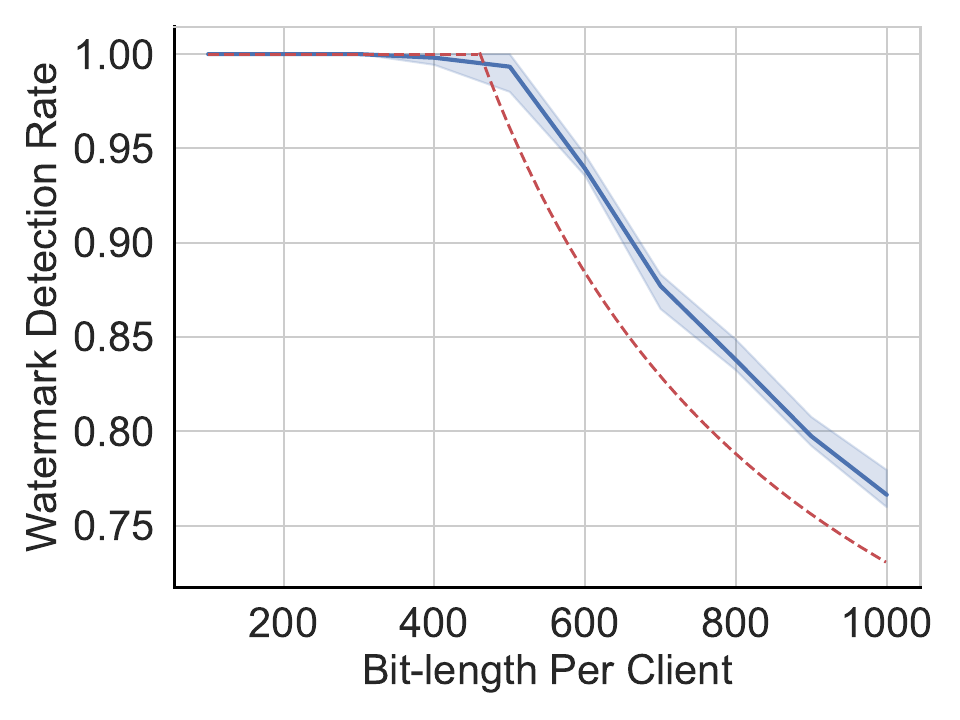}
		\subcaption{\small DistilBERT on SST2}
	  \end{subfigure}
	 \begin{subfigure}{0.24\textwidth}
		\centering
		\includegraphics[keepaspectratio=true, width=130pt]{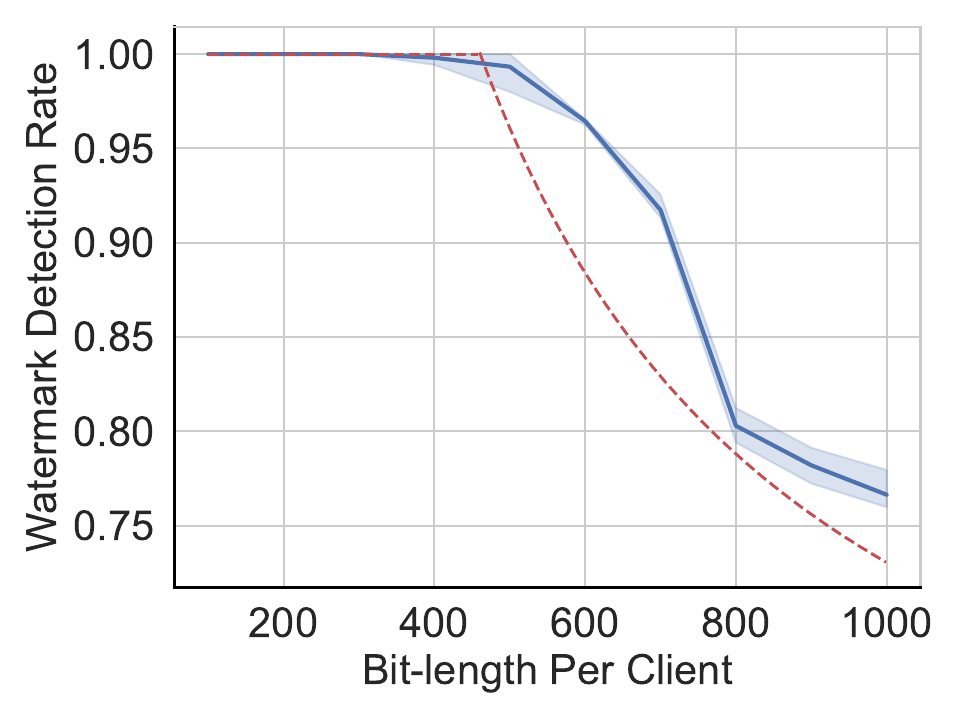}
		\subcaption{\small DistilBERT on QNLI}
	  \end{subfigure}	 

	\centering
	\caption{ Figure provides the lower bound (red dot line) of feature-based watermark detection rate $\eta_F$ given by Theorem \ref{thm: thm1}, and the  empirical results (blue line) are demonstrated to be above the theoretical bound (Case 2) in a SFL setting of $K=5$ clients.}

	\label{fig:thm}
\end{figure}

\noindent\textbf{Feature-based Watermarks}. Fig. \ref{fig:reliability} (a)-(d) illustrate feature-based watermark detection rates $\eta_F$ of varying bit-length of feature-based watermarks, respectively  on four different datasets. 
For convenience, we take that each client in SFL embeds the same length $N$ of watermarks, the significance of feature-based watermarks is as below: 
\begin{itemize}
\item \textbf{Case 1:} As shown in Fig. \ref{fig:reliability}, the detection rate $\eta_F$ remains constant (100\%) within the vertical line (i.e., $M/K$), 
where the total bit-length $KN$ assigned by multiple  ($K$ = 5, 10 or 20) clients does not exceed the capacity of network parameters, which is decided by the channel number $M$ of parameter $\mathbf{S}(\mathbf{W}) = \mathbf{W}_{\gamma}$, 
respectively, e.g., $M=896$ channels across the last 3 layers for AlexNet and $M=2048$ channels across the last 4 layers for ResNet18 (illustrated in Appendix C). 
Therefore, when the total bit-length $KN$ assigned by clients does not exceed the channel number $M$, almost 
all bits of feature-based watermarks can be reliably detected, which is in accordance to the Case 1 of Theorem \ref{thm: thm1}\footnote{If $KN\leq M$, then there exists $\mathbf{W}$ such that    
$\eta_F =1$, the results shown that $\eta_F$ sometimes goes slightly below the lower bound (100\%) provided by Case 1 of Theorem \ref{thm: thm1}. We believe it is because the training is a multi-task optimization process as defined in Eq. \eqref{eq:FedDNN-min-loss}, watermarking optimization $L_{\mathbf{B},\theta}$ is affected by the main training task optimization $L_D$, so the solution of watermarking is compromised to maintain the main task performance.
}.
\item \textbf{Case 2:} When total length of watermarks $KN$ exceeds the channel number $M$ ($KN>M$), 
Fig. \ref{fig:thm} presents 
the detection rate $\eta_F$ drops to about 80\% due to the conflicts 
of overlapping watermark assignments, yet the measured $\eta_F$ is greater than the lower bound given by Case 2 of Theorem \ref{thm: thm1} (denoted by the red dot line).

\end{itemize}

As illustrated in \textbf{Case 2},  feature-based watermarks embedded by $K=5$ clients in SFL may conflict with each other. We give some examples of statistical significance by p-value in Tab. \ref{tab: p_value_feature},  even in the worst case of experiments on four different tasks, the p-value of watermarks is guaranteed below 1.17e-17, which provides a strong evidence to support claim of ownership.

\begin{table}[htbp]
 \renewcommand{\arraystretch}{1.2}
\resizebox{0.48\textwidth}{!}{
\begin{tabular}{ccccc}
\toprule
Task                      & CIFAR10     & CIFAR100 & SST2  & QNLI     \\ \midrule
$N$ Per Client & 400         & 400         & 1000         & 1000         \\ 
Detection Rate $\eta_F$         & 75\%    & 71\%    & 76\%    & 75\%    \\ 
p-value               & 1.29e-24 & 1.14e-17 & 8.61e-64 & 6.73e-59 \\ \bottomrule
\end{tabular}}
\caption{\label{tab: p_value_feature} In the worst case of detection rate, table shows the statistical significance of feature-based watermarks.}

\end{table}

According to Case 1 of Theorem 1, when $KN<M$, feature-based watermarks can be effectively embedded. In order to meet the confidence requirement, each client needs a bit-length $N$ larger than 40, when the number of clients scales to $10^4$, a large number of channels $M$ is required, i.e., a large model is required. In industrial practice, as the number of clients involved increases, the data and the model become larger\cite{lai2021oort}. It is a challenging task to implement FedIPR at scale, given the huge computation costs, we will solve this challenge in our future work.


\noindent\textbf{Backdoor-based Watermarks.}
\begin{table*}[t]
\centering
\setlength{\tabcolsep}{1.0mm}
\renewcommand\arraystretch{1.2}
\begin{tabular}{cccccccc}
\toprule
\multirow{2}{*}{Model/Dataset}    & \multirow{2}{*}{ Client Num.} & \multicolumn{6}{c}{Trigger sample number $N_{\mathbf{T}}$ per client}                                                                                                                                                                                                \\ \cline{3-8} 
                          &                                                  & \multicolumn{1}{c}{50}                   & \multicolumn{1}{c}{100}                   & \multicolumn{1}{c}{150}                  & \multicolumn{1}{c}{200}                  & \multicolumn{1}{c}{250}                   & 300                  \\ \midrule
\multirow{3}{*}{AlexNet/CIFAR10} & 20            & 99.34\% $\pm$ 0.31\% & 99.30\% $\pm$ 0.60\%  & 99.35\% $\pm$ 0.31\% & 99.03\% $\pm$ 0.57\%  & 99.17\% $\pm$ 0.47\%  & 98.85\% $\pm$ 0.69\%  \\ \cline{2-8} 
     & 10    & 99.59\% $\pm$ 0.23\%  & 98.92 \% $\pm$ 0.20\% & 98.45\% $\pm$ 0.67\% & 98.24\% $\pm$ 0.57\%   & 98.43\% $\pm$ 0.15 \% & 97.56\% $\pm$ 1.07\% \\ \cline{2-8} 
     & 5   & 99.29\% $\pm$ 0.38\%  & 99.03\% $\pm$ 0.44\% & 98.15\% $\pm$ 0.74\%   & 98.71\% $\pm$ 0.43\%  & 98.28\% $\pm$ 0.30\%  & 98.39\% $\pm$ 0.64\% \\ \hline
\multirow{3}{*}{ResNet18/CIFAR100}  & 20  & 99.64\% $\pm$ 0.31\% & 99.60\% $\pm$  0.20\%  & 99.35\% $\pm$ 0.31\% & 99.59\% $\pm$ 0.46\% & 99.93\% $\pm$ 0.05\%  & 99.92\% $\pm$ 0.07\%  \\ \cline{2-8} 
    & {10   }  & 99.86\% $\pm$ 0.05\%  & 99.58\% $\pm$ 0.41\%  & 98.56\% $\pm$ 0.57\% & 99.84\% $\pm$ 0.04\%   & 99.83\% $\pm$ 0.15 \% & 99.88\% $\pm$ 0.03\% \\ \cline{2-8} 
    & {5 } & 98.89\% $\pm$ 0.80\%  & 98.54\% $\pm$ 1.3\%    & 99.07\% $\pm$ 2.34\%   & 98.94\% $\pm$ 0.73\%   & 99.45\% $\pm$ 0.06\%   & 98.44\% $\pm$ 0.25\%  \\ \bottomrule
\end{tabular}
\caption{\label{tab:reliablity blackbox} Table presents the superior backdoor-based watermark detection rate (above 95\%). Respectively, table  illustrates $\eta_T$ of varying bit-length $N_{\mathbf{T}}$ of watermarks, where the datasets investigated include CIFAR10 and CIFAR100 datasets and the client number is 5, 10, 20.}

\end{table*}
Tab. \ref{tab:reliablity blackbox} illustrates the detection rate $\eta_T$
\footnote{The trigger samples are regarded as correctly detected when the designated targeted adversarial labels are returned.} and statistical significance of backdoor-based watermarks, 
where  different number of clients ($K$ = 5, 10 or 20) embed backdoor-based watermarks (triggers) generated by Projected Gradient Descent (PGD) method \cite{nguyen2015deep}. 
The results show that the watermark detection rate $\eta_T$ almost keeps constant even the trigger number per client increases as much as $N_{\mathbf{T}}=$ 300. 
Moreover, detection rate $\eta_T$ of watermarks embedded in the more complex ResNet18 is 
as stable as those watermarks embedded in AlexNet. Also, it is noticed that the detection rate is not influenced by the varying number $N_{\mathbf{T}}$ of backdoor samples. 
We ascribe the stable detection rate $\eta_T$ to the generalization capability of over-parameterized networks as demonstrated in \cite{ConvergOverPara_Zhu18,RethinkGene_Zhang17}. 

While with a large set of backdoor-based watermarks are embedded in FedDNN model, we give some examples of the statistical significance by p-value in Tab. \ref{tab: p_value_backdoor}. Even if the detection rate is lower than 100\%, the p-value of watermarks is guaranteed below 4.02e-142, which provides a strong evidence to support claim of ownership. 

\begin{table}[H]
 \renewcommand{\arraystretch}{1.2}
\resizebox{0.48\textwidth}{!}{
\begin{tabular}{ccccc}
\toprule
Task                      & CIFAR10    & CIFAR10  & CIFAR100  & CIFAR100 \\ \midrule
Client Number $K$         & 10         & 5         & 10      & 5    \\ 
$N_{\mathbf{T}}$ Per Client          & 300         & 150         & 150     & 100    \\ 
Detection Rate $\eta_T$         & 97\%    & 98\%    & 98\%    & 98\%    \\ 
p-value               & 1.86e-275 & 4.02e-142 & 5.35e-289 & 4.85e-193 \\ \bottomrule
\end{tabular}}
\caption{\label{tab: p_value_backdoor} In the worst cases of detection rate, table shows  statistical significance of backdoor-based watermarks.}
\vspace{-5pt}
\end{table}

		



\subsection{Robustness under Federated Learning Strategies}\label{Robustness: Strategies}

As illustrated in technical \textbf{Challenge B} of Sect. \ref{Challenge B}, strategies like differential privacy\cite{wei2020federated}, client selection\cite{communicationEfficient/mcMahan2017} and defensive aggregation mechanism\cite{blanchard2017machine, guerraoui2018hidden, yin2018byzantine} 
are widely used for privacy, security and efficiency in secure  federated learning. 
Those strategies intrinsically bring performance decades on the main classification task. 
Respectively, we evaluate the detection rate $\eta_F$ and $\eta_T$ of watermarks under \textbf{Challenge B} to report the robustness of FedIPR. 
\subsubsection{Robustness Against Differential Privacy}

\begin{figure}[htbp]
 
  \centering
  \begin{subfigure}{0.24\textwidth}
    \centering
    \includegraphics[keepaspectratio=true, width=125pt]{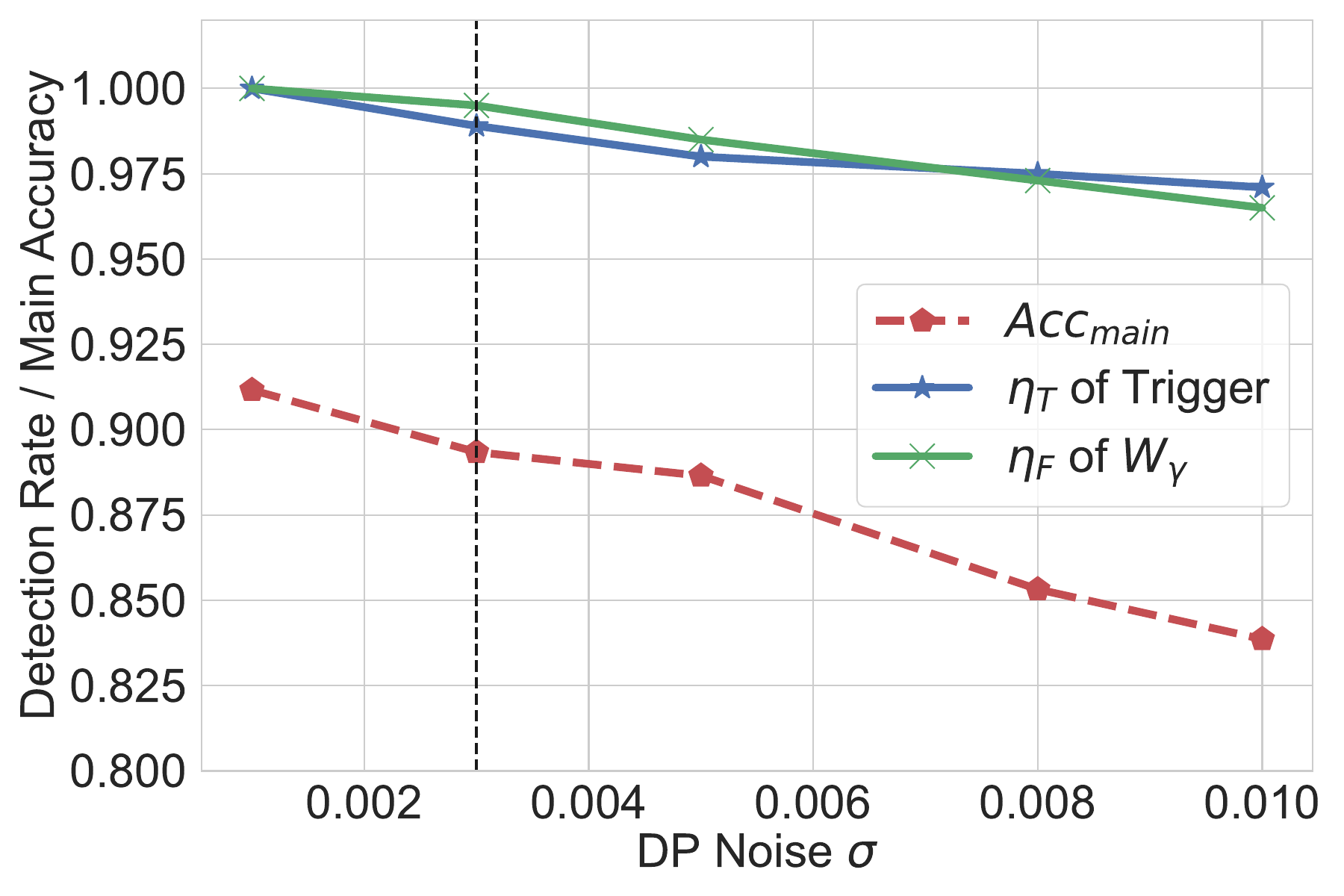}
    \subcaption{\small DP Noise $\sigma$ with AlexNet}
  \end{subfigure}
  \begin{subfigure}{0.24\textwidth}
  \centering
    \includegraphics[keepaspectratio=true, width=125pt]{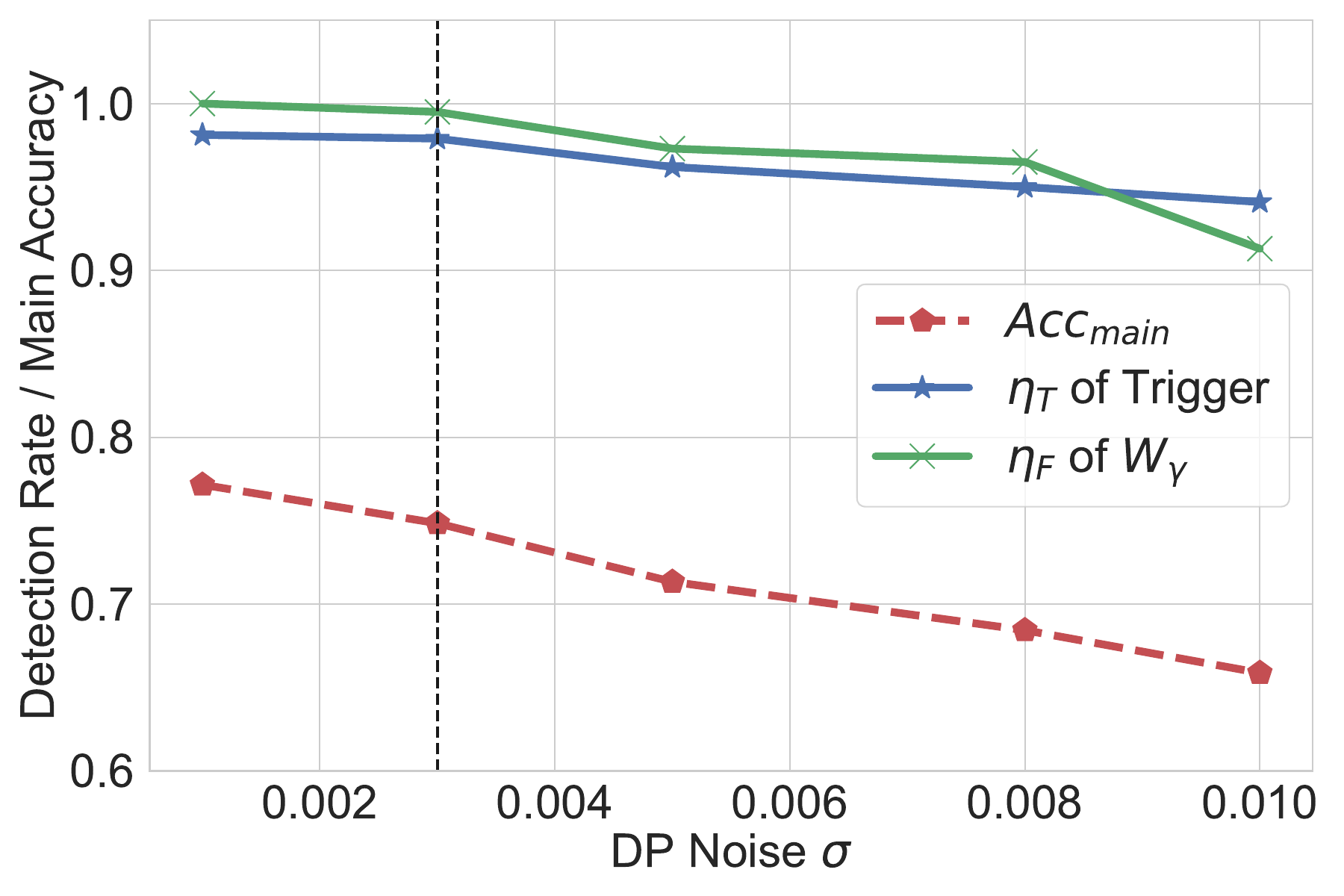}
    \subcaption{\small DP Noise $\sigma$ with ResNet}
  \end{subfigure}

  \caption{\small This figure describes performance of FedIPR under differential privacy strategies using random noise to protect exchanged model information. 
  In a federated learning setting of 10 clients, 
  respectively, figure (a)-(b) illustrate feature-based detection rate $\eta_F$ and backdoor-based detection rate $\eta_T$  under varying differential private noise $\sigma$, where the dot lines
illustrate the main task accuracy $Acc_{main}$.} \label{fig:DP}

\end{figure}

We adopt the Gaussian noise-based method to provide differential privacy guarantee for
federated learning. Specifically, we vary the standard deviation $\sigma$  of Gaussian noise on the local models before clients send local models 
to the server. As Fig. \ref{fig:DP} (a)-(b) show, the main task performance $Acc_{main}$ decreases severely as the $\sigma$ of noise increases, 
and the feature-based detection rate $\eta_F$ and backdoor-based detection rate $\eta_T$ drop a little while the $Acc_{main}$ is within usable range (more than 85\%). 
In a concrete way, when $\sigma$ equals 0.003, classification accuracy $Acc_{main}$, detection rate $\eta_F$ and $\eta_T$ keep a high performance, 
which demonstrates the robustness of watermarks under differential privacy strategy.\vspace{1mm}

We provide some examples of statistical significance by p-value in Tab. \ref{tab: p-value_DP}, even in the worst case of detection rate, the p-value of watermarks is guaranteed below 2.89e-15, which provides a strong evidence to support claim of ownership.

\begin{table}[htbp]

 \renewcommand{\arraystretch}{1.2}
\resizebox{0.48\textwidth}{!}{
\begin{tabular}{ccccc}
\toprule
Task                      & CIFAR10 & CIFAR10 & CIFAR100 & CIFAR100 \\ \midrule
Watermark Type            & Feature & Backdoor & Feature & Backdoor \\ 
$N/N_{\mathbf{T}}$ Per Client            & 80         & 80         & 80        & 80         \\ 
Detection Rate          & 96.25\%    & 97.50\%    & 91.25\%    & 93.75\%    \\ 
p-value               & 7.06e-20 & 2.56e-75 &  2.89e-15 & 2.28e-143 \\ \bottomrule
\end{tabular}}
\caption{\label{tab: p-value_DP} In the worst case of detection rate,  table shows the  statistical significance of watermarks under differential privacy strategy using random noise.}

\end{table}

\subsubsection{Robustness Against Client Selection}
\begin{figure}[htbp]
  
  \centering
  \begin{subfigure}{0.24\textwidth}
    \centering
    \includegraphics[keepaspectratio=true, width=120pt]{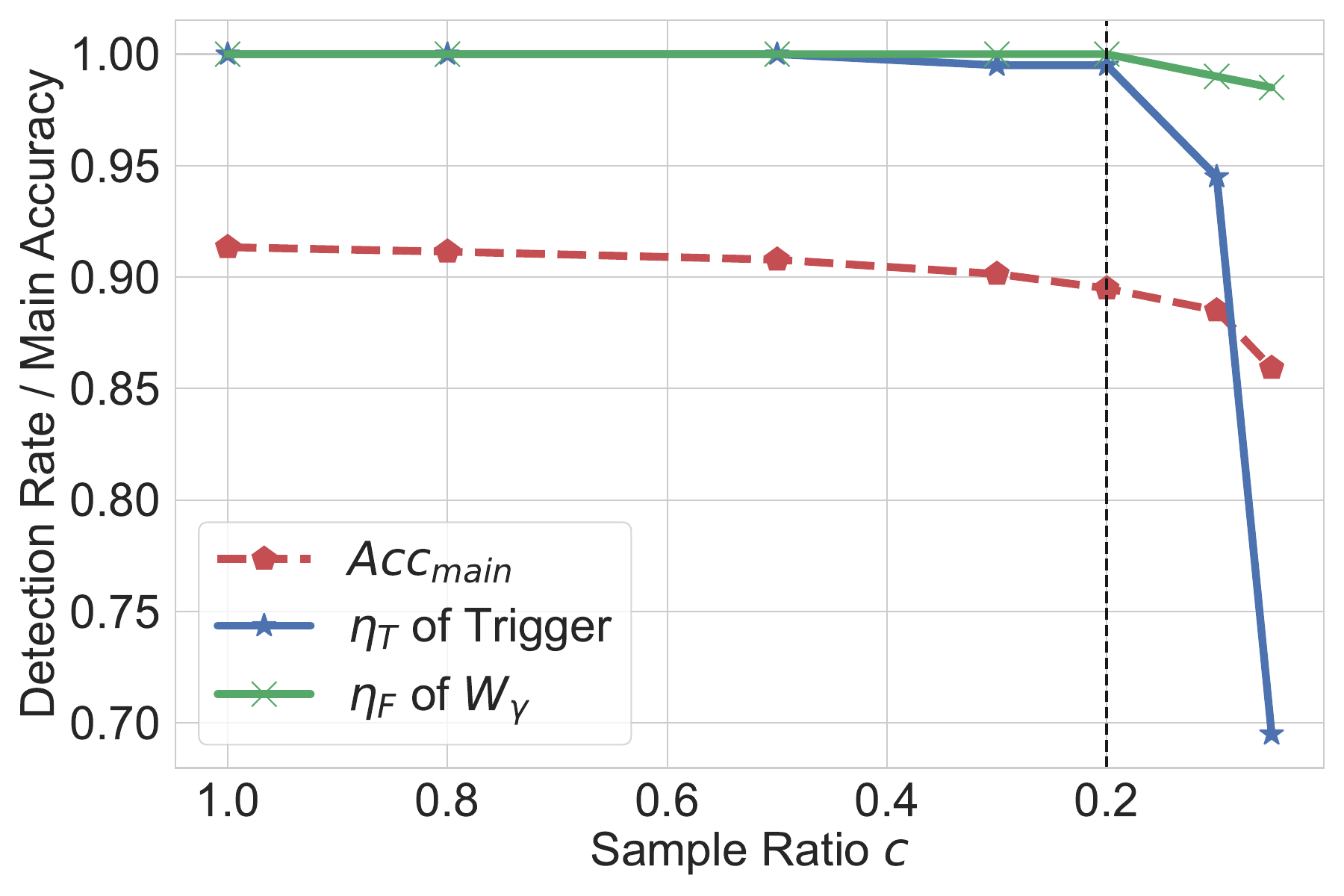}
    \subcaption{\small Sample Ratio with AlexNet}
  \end{subfigure}
  \begin{subfigure}{0.24\textwidth}
    \centering
    \includegraphics[keepaspectratio=true, width=120pt]{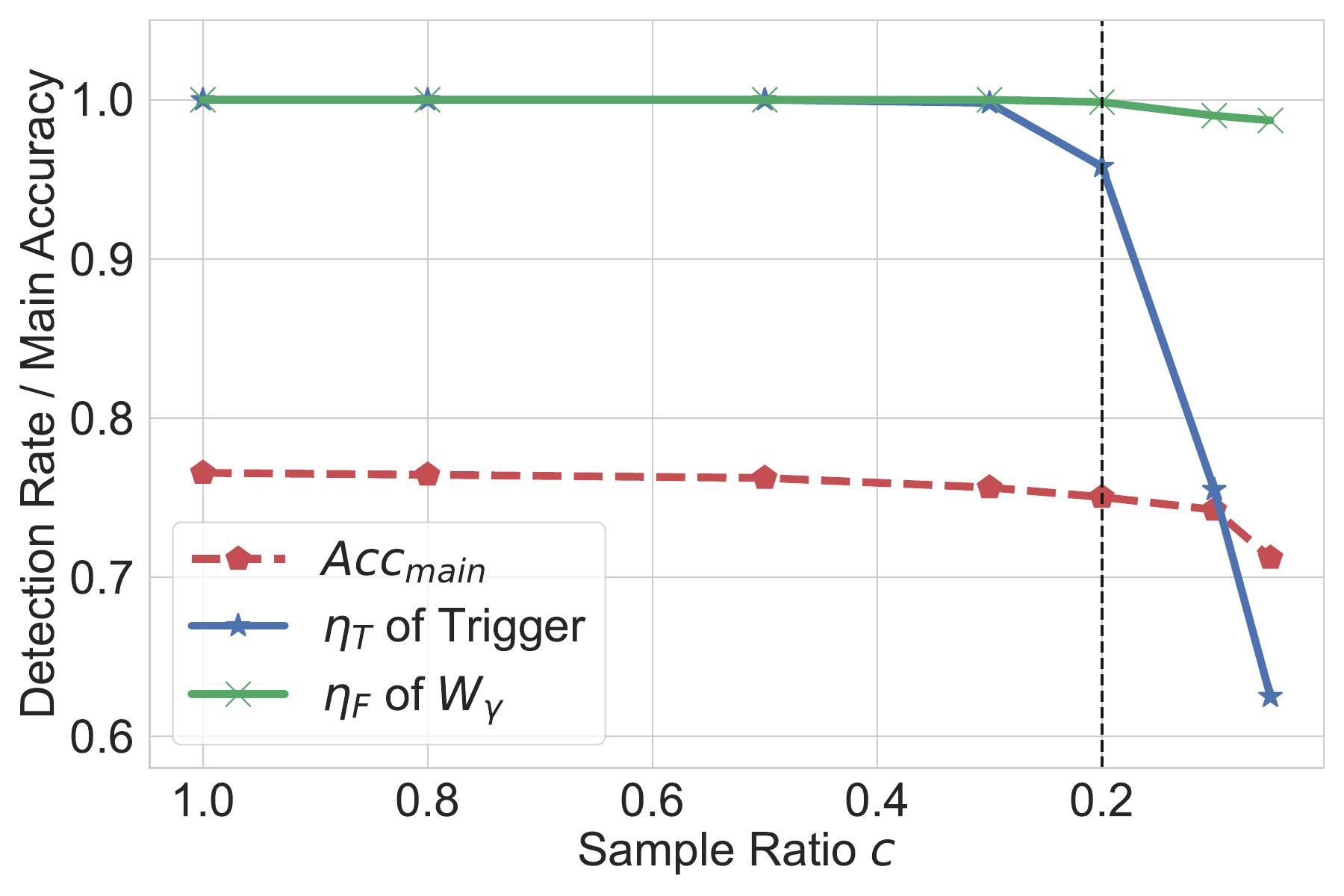}
    \subcaption{\small Sample Ratio with ResNet}
  \end{subfigure}

  \caption{\small Figure describes the robustness of FedIPR under client selection strategy. In a federated learning setting of 10 clients,  respectively, figure (a)-(b) illustrate feature-based detection rate $\eta_F$ and backdoor-based detection rate $\eta_T$ under different sample ratio $c$, whereas the dot lines
illustrate the main task accuracy $Acc_{main}$.} \label{fig:Fraction}

\end{figure}

We  select $cK$ of $K$ clients $(c<1)$ to participate training in each epoch for communication efficiency. 
Fig. \ref{fig:Fraction} shows that the watermarks could not be removed even the sample ratio $c$ is as low as 0.25. 
More specifically, when the sample ratio is larger than 0.2, the 
main classification accuracy $Acc_{main}$ and detection rate $\eta_T$ and $\eta_F$ keep constant. 
This result gives a lower bound of client sampling rate in which watermarks can be effectively embedded 
and verified. 

We give some examples of statistical significance by p-value in Tab. \ref{tab: p-value_Selection}, even in the worst case of detection rate, the p-value of watermarks is guaranteed below 7.02e-20, which provides a  strong evidence to support claim of ownership.

\begin{table}[H]
 \renewcommand{\arraystretch}{1.2}
\resizebox{0.48\textwidth}{!}{
\begin{tabular}{ccccc}
\toprule
Task                      & CIFAR10 & CIFAR10 & CIFAR100 & CIFAR100 \\ \midrule
Watermark Type            & Feature & Backdoor & Feature & Backdoor \\ 
$N/N_{\mathbf{T}}$ Per Client            & 80         & 80         & 80        & 80         \\ 
Detection Rate         & 97.50\%    & 68.75\%    & 96.25\%    & 62.50\%    \\ 
p-value               & 2.68e-21 & 2.74e-36 &  7.02e-20& 6.60e-79 \\ \bottomrule
\end{tabular}}
\caption{\label{tab: p-value_Selection} In the worst case of detection rate, table shows statistical significance of watermarks under client selection strategy.}

\end{table}

\subsubsection{Robustness Against Defensive Aggregation}
Our experiments show that even when defensive methods like Trimmed-Mean, Krum, Bulyan\cite{blanchard2017machine, guerraoui2018hidden, yin2018byzantine} are employed to defense byzantine attacks, 
backdoor-based watermark detection rates of more than 63.25$\%$ can still be maintained, 
which means a near 100 \% probability of detected plagiarism is guaranteed with  p-value less than 1$e^{-30}$. 
The result is shown in Tab. \ref{tab: Bulyan_blackbox}.

\begin{table}[htbp]
 \renewcommand{\arraystretch}{1.2}
\resizebox{0.48\textwidth}{!}{
\begin{tabular}{ccccc}
\toprule
Method                      & Bulyan     & Multi-Krum & Trim-mean  & FedAvg     \\ \midrule
$N_{\mathbf{T}}$ Per Client & 80         & 80         & 80         & 80         \\ 
Detection Rate $\eta_T$         & 68.67\%    & 79.82\%    & 63.25\%    & 98.82\%    \\ 
p-value               & 5.24e-35 & 1.74e-47 &  4.02e-30& 7.20e-78 \\ \bottomrule
\end{tabular}}
\caption{\label{tab: Bulyan_blackbox} Statistical significance of backdoor-based watermarks under defensive aggregation, where 10 clients train AlexNet with CIFAR10 dataset.}

\end{table}

While for feature-based watermarks under defensive methods in SFL, as presented in Tab. \ref{tab: Bulyan_whitebox}, the detection rate remains above 97\%, the p-value of watermarks is guaranteed below 2.68e-21, which provides a strong evidence to support claim of ownership. 

\begin{table}[htbp]
 \renewcommand{\arraystretch}{1.2}
\resizebox{0.48\textwidth}{!}{
\begin{tabular}{ccccc}
\toprule
Method                      & Bulyan     & Multi-Krum & Trim-mean  & FedAvg     \\ \midrule
$N$ Per Client     & 80         & 80         & 80         & 80         \\ 
Detection Rate $\eta_F$         & 98.75\%    & 100\%    & 97.5\%    & 100\%    \\ 
p-value               & 6.61e-23 & 0 & 2.68e-21 & 0 \\ \bottomrule
\end{tabular}}
\caption{\label{tab: Bulyan_whitebox} Statistical significance of feature-based watermarks under defensive aggregation, where 10 clients train AlexNet with CIFAR10 dataset.}

\end{table}


\begin{table*}[htbp]
  \centering
  \setlength{\tabcolsep}{1.0mm}
  \renewcommand\arraystretch{1.2}
  \begin{tabular}{|c|c|c|c|c|c|c|c|}
  \hline
  \multirow{10}{*}{Resnet}  & \multicolumn{7}{c|}{$\mathbf{\beta} = 0.1$}  \\ \cline{2-8} 
  & $N$ Per Client     & 50 & \multicolumn{1}{c|}{100}        & \multicolumn{1}{c|}{200} & \multicolumn{1}{c|}{300} & \multicolumn{1}{c|}{400} & \multicolumn{1}{c|}{500}\\ \cline{2-8} 
  & Accuracy $Acc_{main}$ & 68.61\% $\pm$ 0.14\% & 68.54\% $\pm$ 0.20\%   & 68.33\% $\pm$ 0.05\%    & 67.55\% $\pm$0.07 \%         & 67.38\% $\pm$0.09 \%     &  67.33\% $\pm$ 0.10\%  \\ \cline{2-8} 
  & Backdoor-based  $\eta_T$ & 99.73\% $\pm$ 0.14\% & 99.74\% $\pm$ 0.24\%   & 99.85\% $\pm$ 0.13\%         & 99.75\% $\pm$ 0.23\%         & 99.73\% $\pm$ 0.19\%       &  99.69\% $\pm$ 0.13\% \\ \cline{2-8} 
  & Feature-based  $\eta_F$   & 100.0\% $\pm$ 0\% & 99.73\% $\pm$ 0.25\%   & 99.90\% $\pm$ 0.05\%         & 98.96\% $\pm$ 0.76\%         & 81.83\% $\pm$ 1.38\%       & 78.45\% $\pm$ 0.68\% \\ \cline{2-8} 
  & \multicolumn{7}{c|}{$\mathbf{\beta} = 1.0$} \\ \cline{2-8} 
  & $N$ Per Client  & 50  & \multicolumn{1}{c|}{100}        & \multicolumn{1}{c|}{200} & \multicolumn{1}{c|}{300} & \multicolumn{1}{c|}{400} & \multicolumn{1}{c|}{500} \\ \cline{2-8} 
  & Accuracy $Acc_{main}$ & 74.52\% $\pm$ 0.23\% & 74.48\% $\pm$ 0.33\%         & 74.64\% $\pm$ 0.05 \%        & 73.96\% $\pm$ 0.35\%         & 73.81\% $\pm$ 0.13\%      & 73.31\% $\pm$ 0.06\%  \\ \cline{2-8} 
  & Backdoor-based  $\eta_T$ & 99.85\% $\pm$ 0.14\% & 99.85\% $\pm$ 0.15\%    & 99.74 \% $\pm$ 0.24\%        & 99.75\% $\pm$ 0.23\%          & 0.9974 $\pm$ 0. 23\%   & 0.9973 $\pm$ 0.13\% \\ \cline{2-8} 
  & Feature-based  $\eta_F$ & 100\% $\pm$ 0\% & 99.60\% $\pm$ 0.10\%  & 99.55\% $\pm$ 0.05\%  & 97.66\% $\pm$ 0.73\%  & 80.87\% $\pm$ 1.53\%  & 77.97\% $\pm$ 0.53\% \\ \hline
  \multirow{10}{*}{Alexnet} & \multicolumn{7}{c|}{$\mathbf{\beta} = 0.1$}    \\ \cline{2-8} 
  & $N$ Per Client   & \multicolumn{1}{c|}{50}         & \multicolumn{1}{c|}{100} & \multicolumn{1}{c|}{150} & \multicolumn{1}{c|}{200} & \multicolumn{1}{c|}{250} & 300 \\ \cline{2-8} 
  &  Accuracy $Acc_{main}$  & 82.30\% $\pm$ 0.31\%    & 82.52\% $\pm$ 0.60\%         & 82.65\% $\pm$ 0.19 \%        & 82.42\% $\pm$ 1.02 \%  & 81.49\% $\pm$ 0.85 \% & 81.38\% $\pm$ 0.93 \% \\ \cline{2-8} 
  & Backdoor-based  $\eta_T$ & 99.24\% $\pm$ 0.25\%   & 99.30\% $\pm$ 0.21\%         & 99.53\% $\pm$ 0.15\%  & 99.82\% $\pm$ 0.13\%     & 99.81\% $\pm$ 0.09\%  & 99.82\% $\pm$ 0.13\% \\ \cline{2-8} 
  & Feature-based  $\eta_F$  & \multicolumn{1}{r|}{100\% +0\%} & 99.84\% $\pm$ 0.08\% & 93.49\% $\pm$ 0.40\% & 86.87\% $\pm$ 0.31\%   & 79.87\% $\pm$ 0.51\% & 76.97\% $\pm$ 0.43\% \\ \cline{2-8} 
  & \multicolumn{7}{c|}{$\mathbf{\beta} = 1.0$}                                                                                                                         \\ \cline{2-8} 
  & $N$ Per Client   & \multicolumn{1}{c|}{50}         & \multicolumn{1}{c|}{100} & \multicolumn{1}{c|}{150} & \multicolumn{1}{c|}{200} & \multicolumn{1}{c|}{250} & 300 \\ \cline{2-8} 
  &  Accuracy $Acc_{main}$  & 89.76\% $\pm$ 0.09\%    & 89.50\% $\pm$ 0.07\%         & 89.46\% $\pm$ 0.21\%         & 88.53\% $\pm$ 0.23\%   &  88.60\% $\pm$ 0.13\% & 88.43\% $\pm$ 0.34\% \\ \cline{2-8} 
  & Backdoor-based  $\eta_T$    & 99.25\% $\pm$ 0.25\%                & 99.34\% $\pm$ 0.46\%         & 99.54\% $\pm$ 0. 23\%        & 99.85\% $\pm$ 0.15\%   & 99.75\% $\pm$ 0.31\% & 99.75\% $\pm$ 0.31\% \\ \cline{2-8}
  & Feature-based  $\eta_F$  & 99.80\% $\pm$ 0.20\% & 99.61\% $\pm$ 0.13\% & 94.24\% $\pm$ 0.13\%  & 87.14\% $\pm$ 0.38\%  & 77.14\% $\pm$ 0.48\% & 75.36\% $\pm$ 0.68\% \\ \hline
  \end{tabular}
  \caption{\label{tab:non-IID} Table illustrates the watermark detection rate $\eta_T, \eta_F$ and main accuracy $Acc_{main}$ in a non-iid federated learning setting, of image classification tasks  including AlexNet on CIFAR10 dataset and ResNet on CIFAR100 dataset. $K = 10$ clients embed feature-based and backdoor-based watermarks, with varying feature-based watermark length $N$ per client, the trigger number is set to 80. The results including non-iid settings sampled from dirichlet distribution with $\beta = 0.1$ and 1.}
 
  \end{table*}

\subsection{Robustness Against Removal Attack}\label{robustness: removal}
In this subsection, we showcase that FedIPR are robust against removal attacks conducted by plagiarizers that attempt to remove the watermarks. The feature-based watermarks embedded in normalization layer are shown to be especially persistent  against both fine-tuning attack and pruning attack, while those watermarks in the convolution layers are not.

\subsubsection{Robustness against Fine-tuning Attack}
Fine-tuning attack on  watermarks is conducted to train the network without the presence of the regularization term, $i.e., L_\mathbf{T}$ and $L_{\mathbf{B},\theta}$. 
In Fig. \ref{fig:removal} (a), it is observed that the detection rate $\eta_F$ of watermarks embedded with normalization layer ($\mathbf{W}_\gamma$) remains at 100\% (blue curve). 
In contrast, the detection rate $\eta_F$ of watermarks embedded with convolution layer ($\mathbf{W}_C$) drops significantly (purple curve). 
The superior robustness of feature-based watermarks embedded in normalization layer is in accordance to the observations reported in \cite{fan2019rethinking}. While for backdoor-based watermarks, the detection rate $\eta_T$ remains above 90\% (yellow curve), which is also robust against fine-tuning attack.

\subsubsection{Robustness against Pruning Attack}
\begin{figure}[htbp]

  \centering
  \begin{subfigure}{0.24\textwidth}
    \centering
    \includegraphics[keepaspectratio=true, width=120pt]{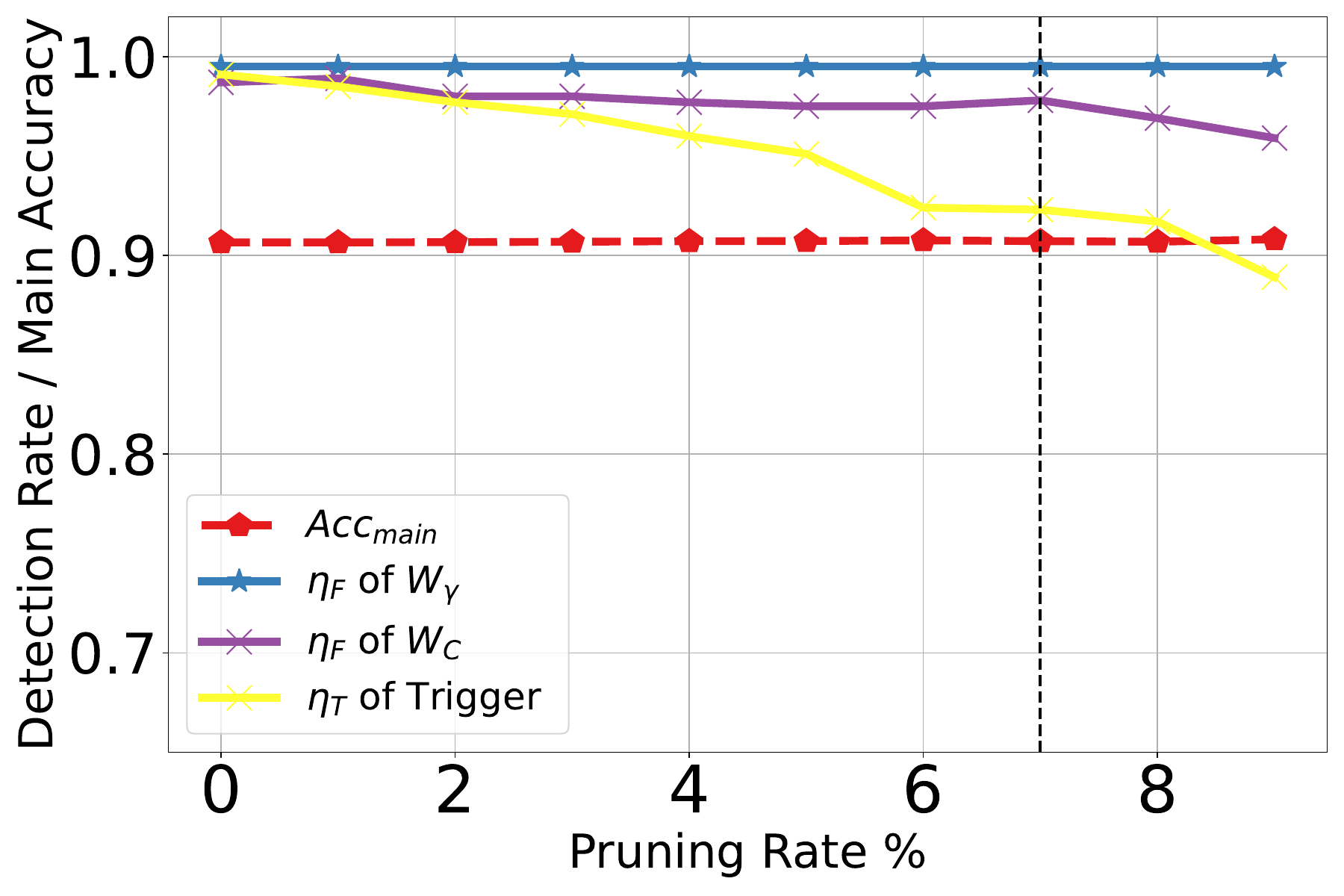}
    \subcaption{\small Fine-tuning Epoch/5}
  \end{subfigure}
  \begin{subfigure}{0.24\textwidth}
  \centering
    \includegraphics[keepaspectratio=true, width=120pt]{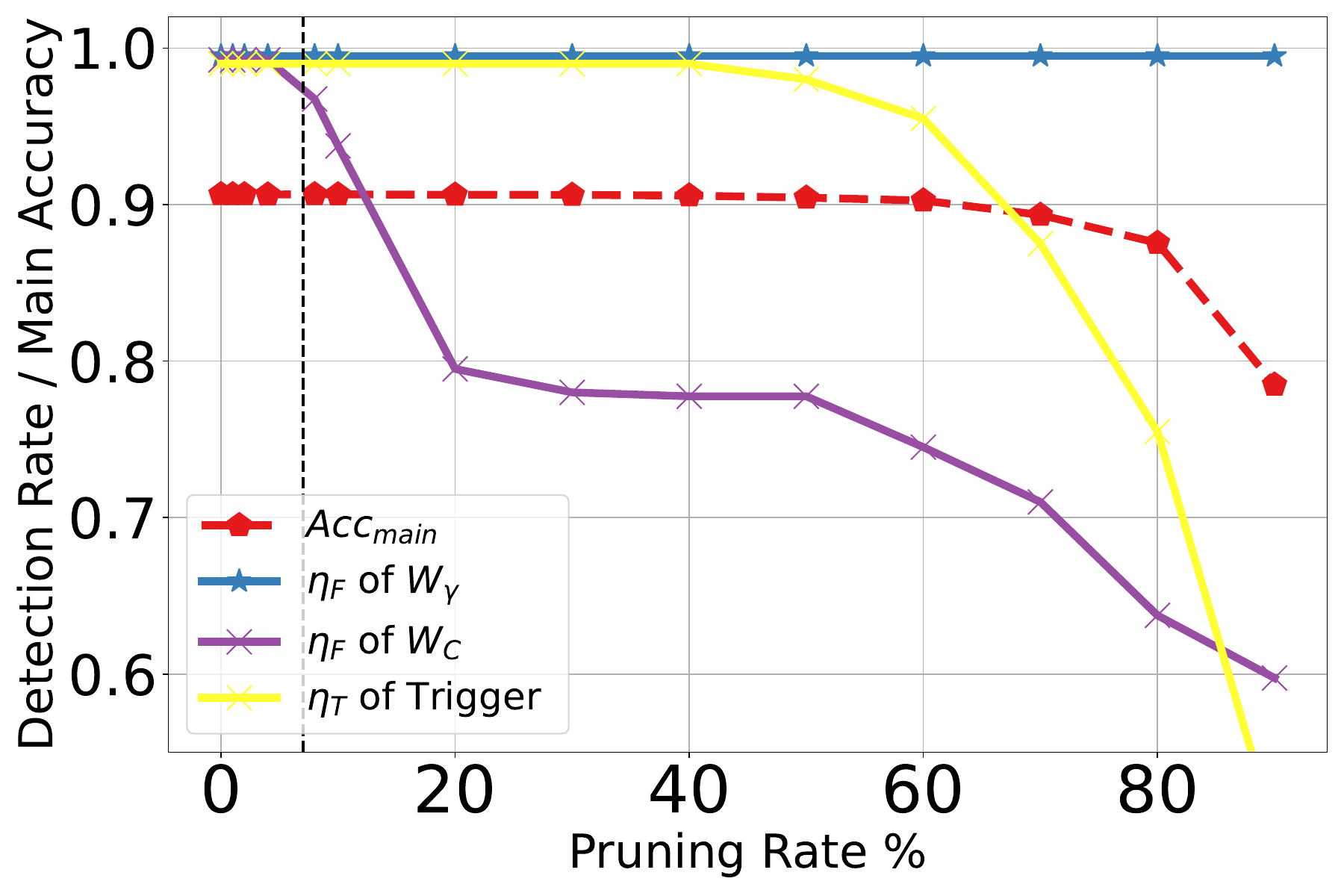}
    \subcaption{\small Pruning Rate}
  \end{subfigure}

  \caption{\small Figure describes the robustness of our FedIPR under removal attacks. In a federated learning setting, $K$ = 10 clients train  AlexNet with CIFAR10 dataset. 
The dot lines in figure (a) and (b) illustrate the main task classification accuracy $Acc_{main}$ under diverse settings. 
Respectively, figure (a) illustrates feature-based and backdoor-based watermark detection rate against model finetuning attack in 50 epochs; 
figure (b) illustrates feature-based and backdoor-based watermark detection rate  against model pruning attack with varying pruning rate. Note that detection rates  $\eta_F$ of feature-based watermarks in convolution layer are severely degraded, but detection rates $\eta_F$ in both normalization layer ($\mathbf{W}_{\gamma}$) are persistent.
} 
  \label{fig:removal}
\end{figure}

The pruning attack removes redundant parameters from the trained model. 
We evaluate the main task performance $Acc_{main}$ and watermark detection rate $\eta_F$ and $\eta_{T}$ under pruning attack with varying pruning rates. 
Fig. \ref{fig:removal} (b) shows watermark detection rate $\eta_F$ and $\eta_{T}$ while varying proportions of network parameters are pruned. It is observed that the detection rate $\eta_F$ of watermarks embedded in the normalization layer is stable all the time, 
while $\eta_F$ with $\mathbf{W}_C$ are severely degraded.  This fact shows that the watermarks on normalization parameters are more robust against pruning attack. While for backdoor-based watermarks, the detection rate $\eta_T$ remains above 90\% (yellow curve) when pruning rate is less than 60\%, which is also robust against pruning attack.  

\subsection{Robustness of Triggers against Adversary}
An adversary might try to obtain counterfeited triggers $\hat{\mathbf{T}}$ by generating adversarial examples with a surrogate network $\mathbb{N}_{sur}$, and attempt to pass the ownership verification of target model $\mathbb{N}$ with those triggers $\hat{\mathbf{T}}$ as follows:
\begin{equation}
            \begin{small}
		\mathcal{V}_B\big( \mathbb{N}, \hat{\mathbf{T}} \big) =  \left\{ \begin{array}{cc}
			\text{TRUE}, & \text{if }\mathop{{}\mathbb{E}}_{\hat{\mathbf{T}}}( \mathbb{I}(\mathbf{Y}_{\hat{\mathbf{T}}}  \neq \mathbb{N}( \mathbf{X}_{\hat{\mathbf{T}}} ) ) ) \leq \epsilon_B,  \\
			\text{FALSE}, & \text{otherwise},  \\
		    \end{array} \right.
            \end{small}
 \end{equation}  
We give results in 2 cases to show that in SFL setting, adversary has little knowledge of the original training data or backdoor triggers, thus it is hard for an adversary to obtain triggers that can pass the verification.

\noindent\textbf{Case1:} If the attacker randomly generates some adversarial samples from surrogate model $\mathbb{N}_{sur}$ as triggers $\hat{\mathbf{T}}$, but does not retrain the model with the triggers $\hat{\mathbf{T}}$. 
When the triggers $\hat{\mathbf{T}}$ are input to the API for verification,  the accuracy that the model outputs the target label is an almost random guessing (e.g., in a CIFAR10 classification task, the detection rate of trigger is about 10\%), which is shown in the following Tab. \ref{case1}. The results indicates that the random generated triggers can not pass the verification without retraining.

\begin{table}[htbp]
\renewcommand{\arraystretch}{1.2}
\centering
\begin{tabular}{cccccc}
\toprule
\multirow{2}{*}{Trigger type} & \multicolumn{5}{c}{Number of Triggers}                                          \\

\cline{2-6} 
                              & 100& 200& 300 & 400 &500\\ 
\midrule
True trigger $\mathbf{T}$                 & 100\% & 99.63\%  & 99.65\%   & 99.52\% & 99.72\% \\
Counterfeited $\hat{\mathbf{T}}$                & 9.34\%   & 11.21\%   & 10.15\%  & 9.36\%  & 11.09\%  \\ 
\bottomrule 
\end{tabular}
\caption{In the CIFAR10 classification task with AlexNet,  an attacker randomly generates some adversarial samples $\hat{\mathbf{T}}$ from the surrogate model $\mathbb{N}_{sur}$ with a set of base images (unrelated with the private training data). Table shows the accuracy that the model $\mathbb{N}$ outputs triggers as targeted label. 
}\label{case1}
\end{table}

\noindent\textbf{Case2:} If the attacker randomly generates some triggers $\hat{\mathbf{T}}$ and retrains the model $\mathbb{N}$ with those triggers, results in the following Fig. \ref{case2} show that the 200 trigger images can be embedded with 80 epochs of training, the triggers are generated with unrelated base images with CIFAR10 dataset, 
which will result in main task accuracy decades larger than  10 percent. Such performance descend defeats the purpose of plagiarism to obtain an existing model at no cost.

\begin{figure}[htbp]
\vspace{-10pt}
  \centering
  \begin{subfigure}{0.24\textwidth}
    \centering
    \includegraphics[keepaspectratio=true, width=110pt]{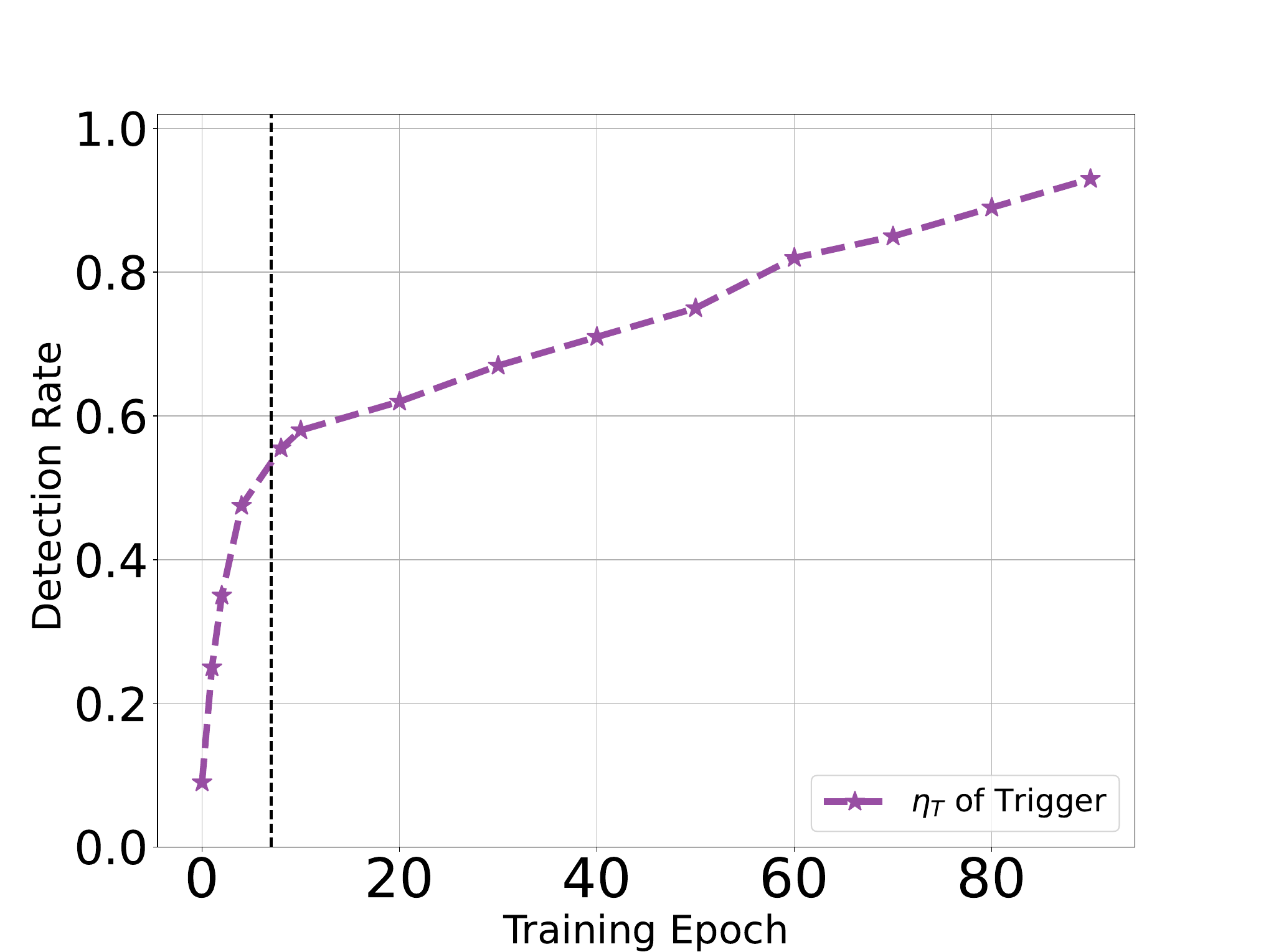}
    \subcaption{\small Trigger Detection Rate}
  \end{subfigure}
  \begin{subfigure}{0.24\textwidth}
  \centering
    \includegraphics[keepaspectratio=true, width=110pt]{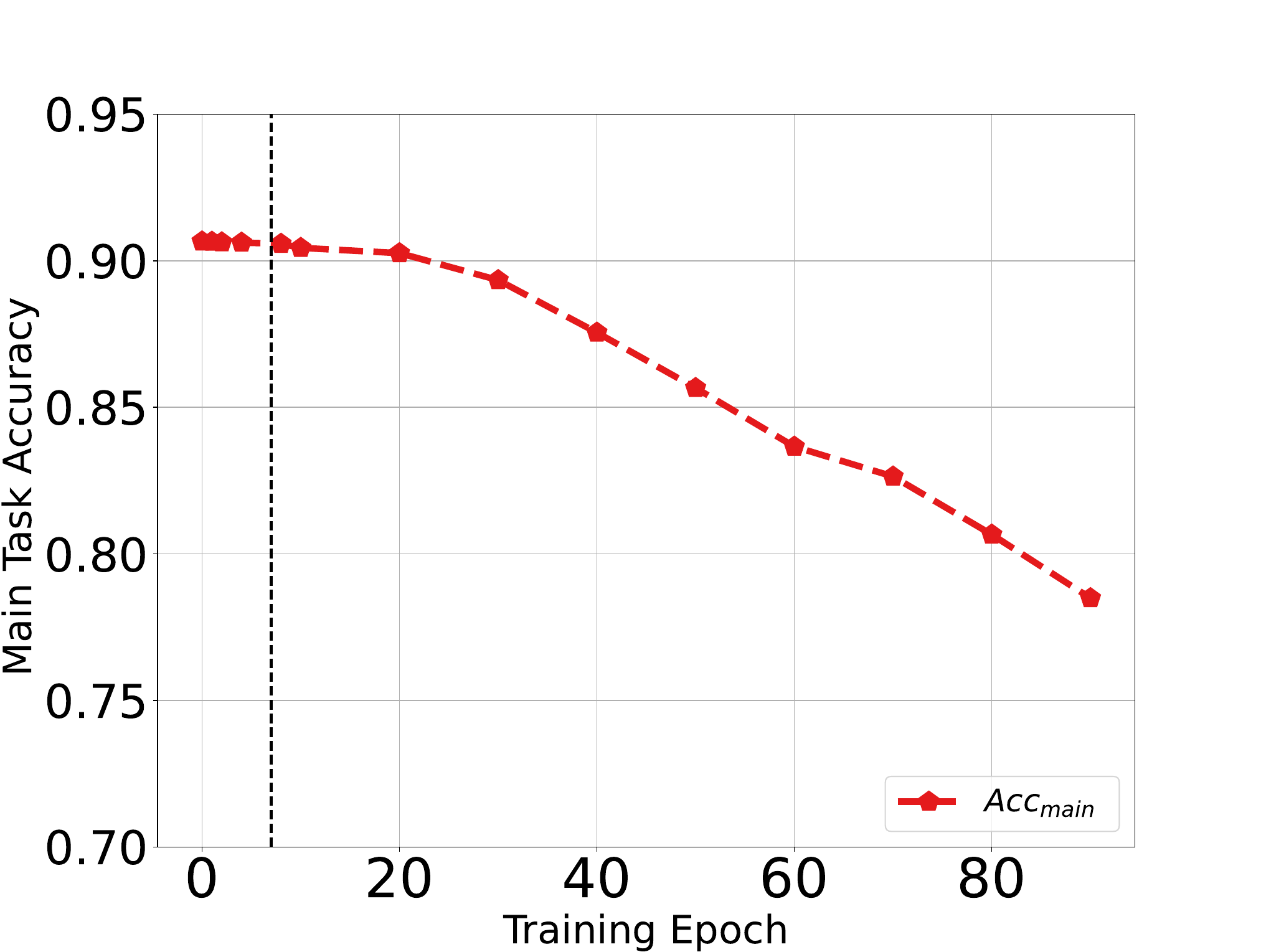}
    \subcaption{\small Main Task Accuracy}
  \end{subfigure} 
	\caption{Figure presents the trigger detection accuracy and the main accuracy while adversary retrains the model $\mathbb{N}$ with triggers $\hat{\mathbf{T}}$, in the CIFAR10 classification task with AlexNet. }
	\label{case2}
	\vspace{-10pt}
\end{figure}

\subsection{Robustness under Non-iid Setting} \label{robustness: noniid}

In federated learning, data distributions across clients are often not identical and independent distributed (iid). We also evaluate FedIPR in \textit{lable-skew} non-iid federated learning setting, where we assume each client's training examples are drawn with class labels following a \textit{dirichlet distribution}\cite{li2021federated}, $\beta > 0$ is the concentration parameter controlling the identicalness among users.

In our experiments, we conduct image classification experiments with AlexNet on CIFAR10 and ResNet on CIFAR100, the dirichlet parameter is $\beta = 0.1$ and 1. The results presented in Tab. \ref{tab:non-IID} indicate that FedIPR works with non-iid setting.

In non-iid setting, while a large set of feature-based watermarks are embedded in FedDNN model, we give some examples of statistical significance by p-value in the worst case of detection rate in Tab. \ref{tab: p_value_noniid}.

\begin{table}[H]
 \renewcommand{\arraystretch}{1.2}
\resizebox{0.48\textwidth}{!}{
\begin{tabular}{ccccc}
\toprule
Task                      & CIFAR10    & CIFAR10  & CIFAR100  & CIFAR100 \\ \midrule
Non-iid $\beta$         & 0.1         & 1         & 0.1      & 1    \\ 
$N$ Per Client          & 300         & 300         & 500     & 500    \\ 
Detection Rate $\eta_F$         & 76\%    & 75\%    & 78\%    & 77\%    \\ 
p-value               & 2.42e-20 & 7.16e-19 & 4.76e-38 & 2.33e-35 \\ \bottomrule
\end{tabular}}
\caption{\label{tab: p_value_noniid} In the worst case of detection rate, table shows the statistical significance of watermarks under non-iid SFL.}

\end{table}

As shown in Tab. \ref{tab: p_value_noniid}, even in the worst case, the p-value of watermarks can be guaranteed below 7.16e-19, which provides a  strong evidence to support claim of ownership.

\subsection{Watermarks Defeat Freerider Attacks}\label{sect: freerider_exp}

As a precaution method for freerider attack, watermarks are embedded into the FedDNN, the benign clients can verify ownership by extracting predefined watermarks from FedDNN, 
while freeriders can not detect watermarks because they do not perform actual training. We conduct experiments testing the local models by three types of clients including plain freerider, 
freerider with Gaussian noise (defined in Sect. \ref{freerider}) and benign clients that contribute data and computation.

\begin{figure}[h] 
	\vspace{-10pt}         
	
	\centering
	 \begin{subfigure}{0.155\textwidth}
		\centering
		\includegraphics[keepaspectratio=true, width=90pt]{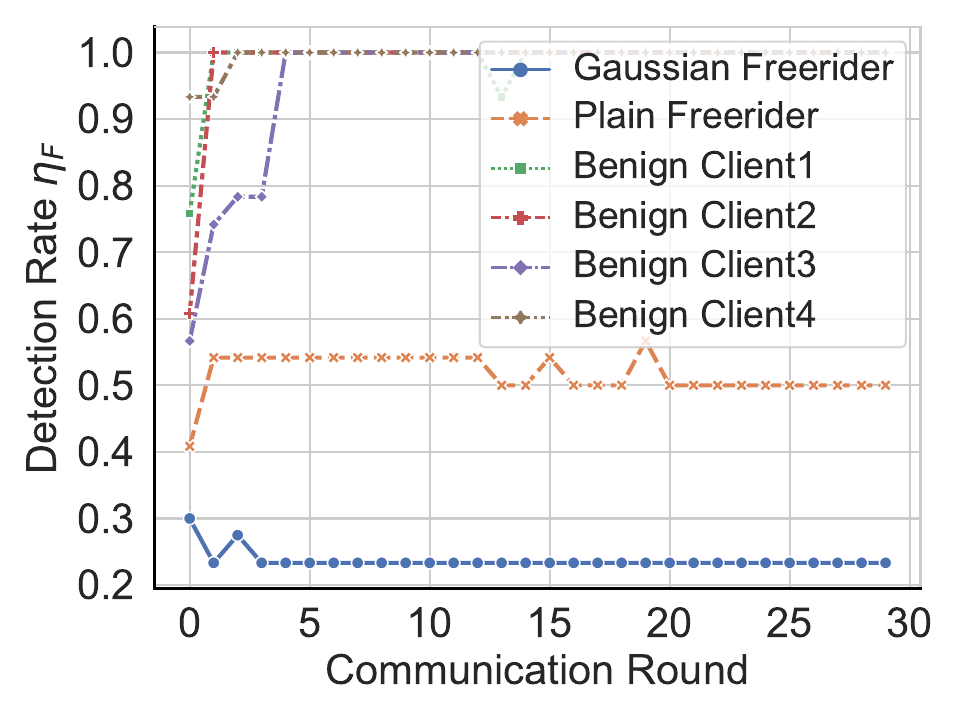}
		\subcaption*{\small 20 bits with AlexNet}
	  \end{subfigure}
	  \begin{subfigure}{0.155\textwidth}
		\centering
		\includegraphics[keepaspectratio=true, width=90pt]{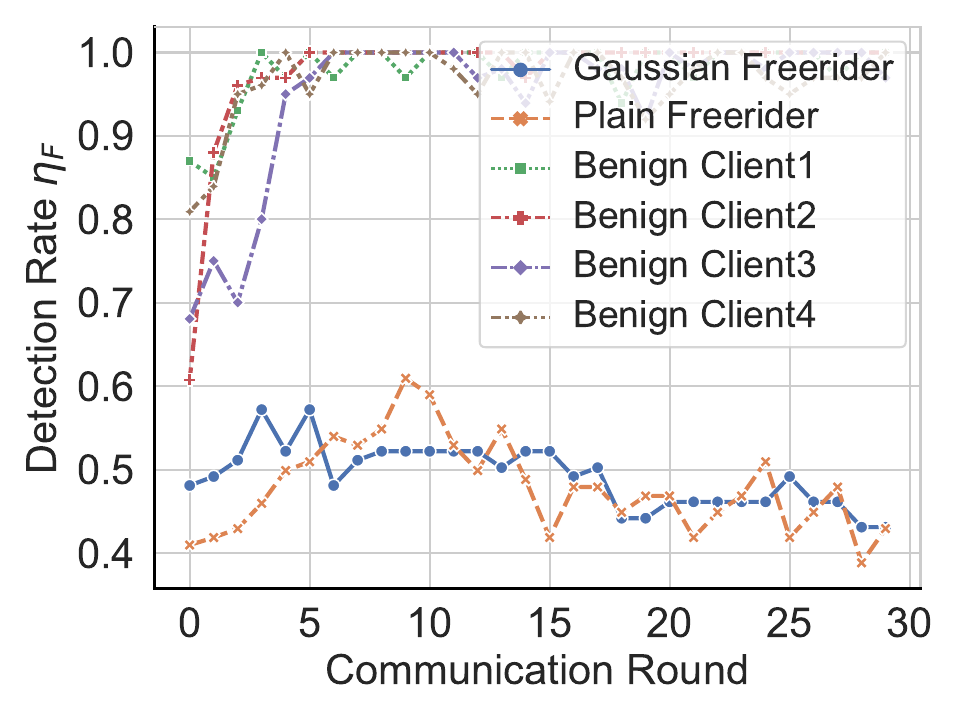}
		\subcaption*{\small 40 bits with AlexNet}
	  \end{subfigure}
	  \begin{subfigure}{0.155\textwidth}
		\centering
		\includegraphics[keepaspectratio=true, width=90pt]{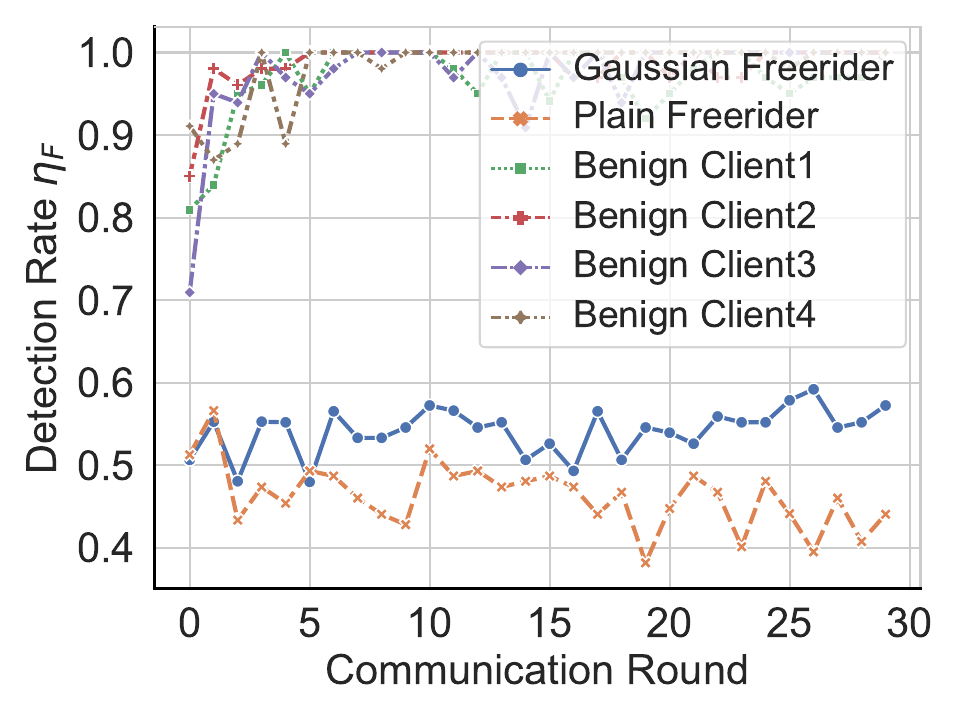}
		\subcaption*{\small 60 bits with AlexNet}
	  \end{subfigure}
	  
	 \begin{subfigure}{0.155\textwidth}
		\centering
		\includegraphics[keepaspectratio=true, width=90pt]{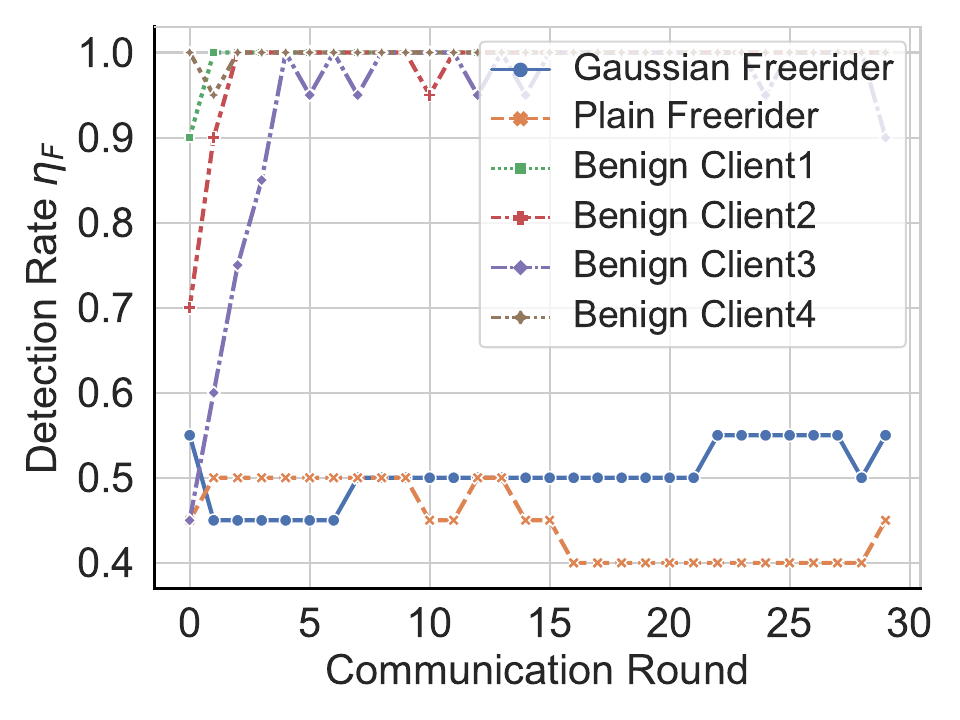}
		\subcaption*{\small 20 bits with ResNet}
	  \end{subfigure}
	 \begin{subfigure}{0.155\textwidth}
		\centering
		\includegraphics[keepaspectratio=true, width=90pt]{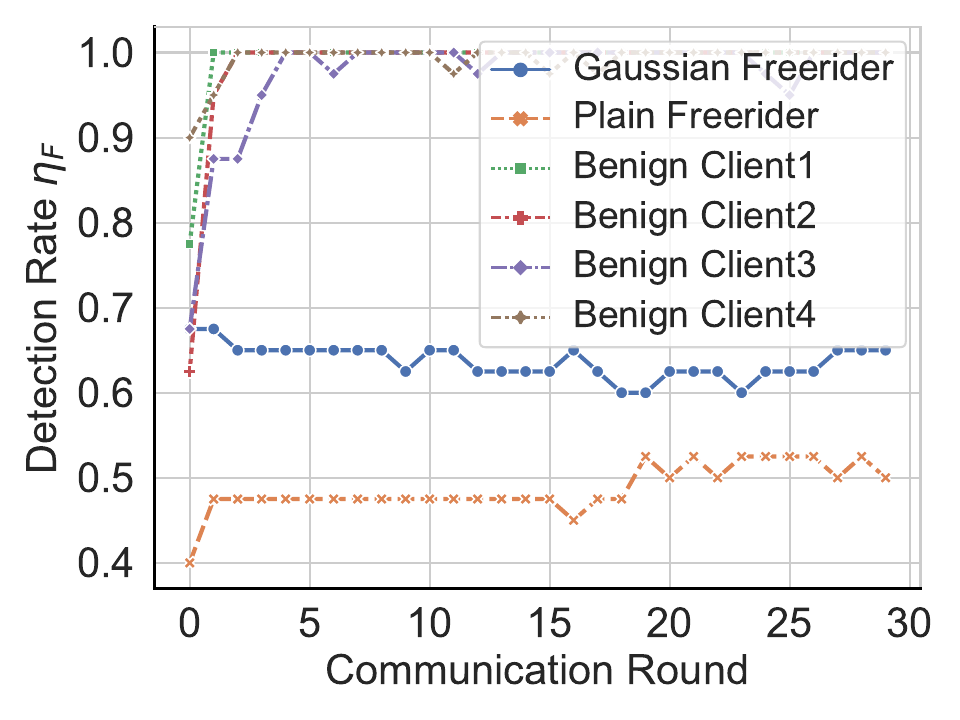}
		\subcaption*{\small 40 bits with ResNet}
	  \end{subfigure}	  
	 \begin{subfigure}{0.155\textwidth}
		\centering
		\includegraphics[keepaspectratio=true, width=90pt]{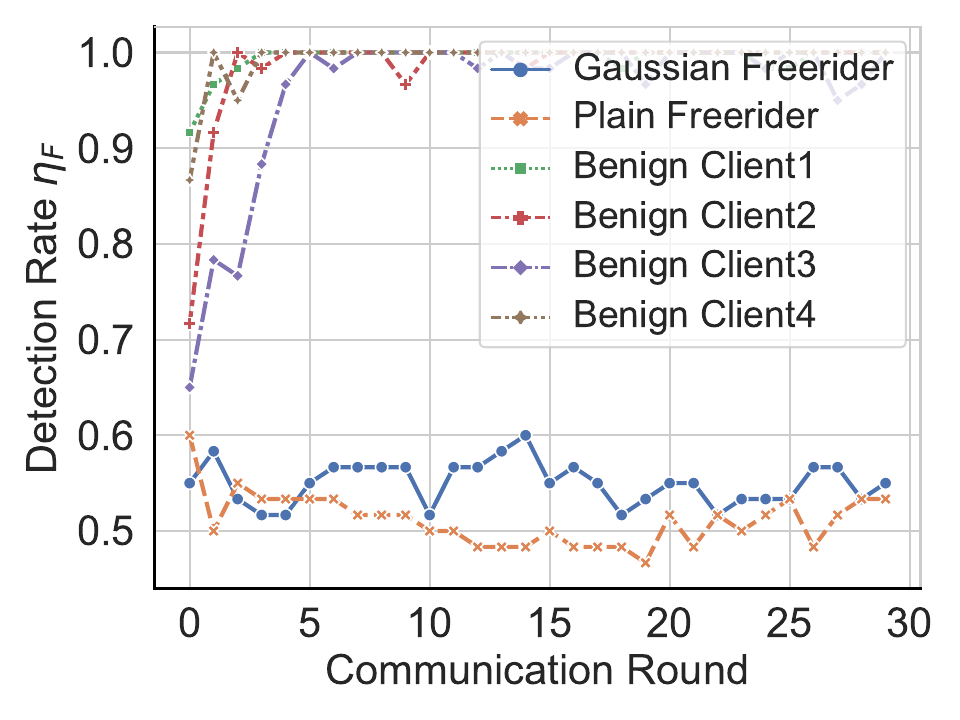}
		\subcaption*{\small 60 bits with ResNet}
	  \end{subfigure}	  

	\centering
	\caption{Comparisons   between three different types of clients including: (1) freerider clients with previous local models. (orange lines) (2) freerider clients disguised with Gaussian noise. (blue lines) 
	(3) four benign clients in a 20 client SFL. The feature-based watermark detection rate $\eta_F$ 
	is measured in each communication round. Note the sharp contrast of $\eta_F$ between four benign clients and two freerider clients (plotted in orange and blue).}
	\vspace{-5pt}
	\label{fig:freerider}
\end{figure}

We consider a setting that the server conducts feature-based verification on each client's local models, in a 22-clients federated learning including one freerider client with Gaussian noise and 
one plain freerider client with previous local models. In each communication round, the results are presented in Fig. \ref{fig:freerider}, which 
show that the server can detect the benign clients' watermarks in the global model at quite early stage (in 30 communication rounds) of FedDNN model training, 
the watermark detection rate $\eta_F$ is nearly 100\%; 
while the freeriders failed to verify their watermarks because they do not contribute actual training, the $\eta_F$ detection rate is an almost random guess (50\%). 

\section{Conclusion}\label{Discussion}

This paper presents a novel ownership verification scheme to protect the Intellectual Property Right (IPR) of Federated DNN models against external plagiarizers who illegally copy, re-distribute the models. 
To our best knowledge, it is the first ownership verification scheme that aims to protect model intellectual property rights under secure federated learning setting. This work addresses a crucial issue remained open in secure federated learning research, since the protection of valuable federated learning models is as important as protecting data privacy. 

On the technical side, this work demonstrates that reliable and persistent watermarks could be embedded into local models without disclosing the presence and extraction parameters of these watermarks. In particular, normalization scale parameters based on watermarks are extremely robust under federated learning strategies and against removal attacks. We wish that the formulation illustrated in this paper will lead to watermark embedding and verification in various federated learning settings. 


\ifCLASSOPTIONcompsoc

\fi


\bibliography{nnpass_related}{}
\bibliographystyle{IEEEtran}

\clearpage
\appendices
\setcounter{equation}{0}
\setcounter{theorem}{0}
\setcounter{defn}{0}
\section{Proof of Theorem \ref{thm: thm1}}

\begin{defn} \label{def:U}For $\{\mathbf{\theta}_k\}_{k=1}^K$ of $K$ clients, the embedding matrix $\{\mathbf{E}_k\}_{k=1}^K \in \{\mathbf{\theta}_k\}_{k=1}^K$ form a combined matrix 
\begin{equation}
	\mathbf{U}^{M\times KN} = \{\mathbf{E}_1^{M\times N}, \mathbf{E}_2^{M\times N},\cdots, \mathbf{E}_K^{M\times N}\}, 
\end{equation} 
whereas $M$ denotes the channel number of targeted watermark parameters $\mathbf{W} \in \mathbb{R}^M$. Let $\Tilde{\mathbf{U}}^{M\times KN}$ be matrix combined with 
\begin{equation}
 \begin{split}
    \{(\mathbf{E}_1diag(\mathbf{B}_1))^{M\times N}, (\mathbf{E}_2diag(\mathbf{B}_2))^{M\times N}\\,\cdots,(\mathbf{E}_Kdiag(\mathbf{B}_K))^{M\times N}\},
\end{split}   
\end{equation}
where $\mathbf{B}_k = (t_{k1}, t_{k2}, \cdots, t_{kN}) \in \{+1, -1\}^N$ is the watermark of $k$-th client.
\end{defn}

\begin{defn}\label{def:detection rate}\textbf{Detection Rate Defined by Matrix }\\
Let $\#\{\mathbf{W}^T\mathbf{A} > 0\}$ be the number of positive elements of $\mathbf{W}^T\mathbf{A}$, for any matrix $\mathbf{W}^T\mathbf{A}$, 
for any matrix $\mathbf{A}\in \mathbb{R}^{M \times KN}$ and vector $\mathbf{W}^{M \times 1}$. 
\begin{equation}
\eta_F=\E\limits_{\Tilde{\mathbf{U}}}(\underset{\mathbf{W} \in \mathbb{R}^M}{sup}\text{\#}\{\mathbf{W}^T \Tilde{\mathbf{U}} >  0\})/KN.
\end{equation}
\end{defn}

\noindent\textbf{Remark:} 
The domain of $\mathbf{W}$ in Def. \ref{def:detection rate} and following analysis we consider is total space $\mathbb{R}^M$ for parameter $\mathbf{W}$, 
which does not consider the influence of optimization process $L_D$ in Eq. \eqref{eq:FedDNN-min-loss}. Optimization process of main task $L_D$ may lead parameters $\mathbf{W}$ to converge in a subspace of total space.

The following Theorem \ref{thm: thm1} first elucidates the condition under which a feasible solution exists for $K$ different watermarks embedding. Moreover, if the condition is not satisfied, the lower bound of detecting rate $\eta_F$ is provided.

\begin{theorem}
Let $\mathbf{U}$, $\tilde{\mathbf{U}}$ and $\eta_F$ defined above. If $\mathbf{U}$ is column non-singular matrix, 
\begin{item}
\item \textbf{Case 1:} If $rank(\mathbf{U}) =KN\leq M$, then there exists $\mathbf{W}$ such that  $\mathbf{W}^T\Tilde{\mathbf{U}} \geq 0$. Moreover, we have
$ \eta_F =1$.
\item \textbf{Case 2:} If $rank(\mathbf{U})=M <KN$, then, we have,
\begin{equation}
\eta_F \geq \frac{KN+M}{2KN}.
\end{equation}
\end{item}
\end{theorem}
\begin{proof}

    




For the \textbf{Case 1}, if $rank(\mathbf{U}) = K$, then the column of $\mathbf{U}$ ($U_1, U_2, \cdots U_{KN}$) is independent, and the column of $\Tilde{\mathbf{U}}$ ($\Tilde{U_1}, \Tilde{U_2}, \cdots \Tilde{U}_{KN}$) is also independent. Thus,
\begin{equation}
\begin{aligned}
&y_1\Tilde{U_1} + y_2\Tilde{U_2} + \cdots + y_{KN}\Tilde{U_{KN}} = 0 \\
&\Longleftrightarrow y_1=y_2=\cdots=y_{KN}=0
\end{aligned}
\end{equation}
Therefore the solution of $\Tilde{\mathbf{U}}\vec{y} = \vec{0}$ is only $\Vec{0}$, moreover, $\Tilde{\mathbf{U}}\vec{y} = \vec{0}$ doesn't have non-negative solutions except $\Vec{0}$. According to Gordan's theorem \cite{alon1986regular}, Either $\Tilde{\mathbf{U}}\vec{y} > 0$ has a solution y, or $\Tilde{\mathbf{U}}\vec{y} = \vec{0}$ has a nonzero solution y with $\vec{y} \geq \vec{0}$. Since the latter statement is wrong, there exists $\mathbf{W} = \vec{y}^T$ such that $\mathbf{W}\Tilde{\mathbf{U}}>0$. Therefore, $\eta_F \geq 1$. And since $\eta_F \leq 1$, we have $\eta_F =1$.


For the \textbf{Case 2} as $rank(\mathbf{U})= M \leq KN$, since $\Tilde{\mathbf{U}}=(U_1diag(\mathbf{B}_1), \dots, U_{KN}diag(\mathbf{B}_{KN}))$,
where $\mathbf{B}_i \in \{-1, 1\}, 1\leq i \leq KN $, 
$rank(\Tilde{\mathbf{B}})=rank(\mathbf{B})=M < KN $.

Without loss of generality, we assume there is a partition of $\tilde{\mathbf{U}}$ as $\Tilde{\mathbf{U}} = (\Tilde{\mathbf{U}}_1', \Tilde{\mathbf{U}}_2')$, where $\Tilde{\mathbf{U}}_1' \in R^{M\times M}, \Tilde{\mathbf{U}}_2' \in R^{M \times KN-M}$ and $rank(\Tilde{\mathbf{U}}_1') = M$. Denote $\Omega = \{ x\in R^M|x^T\Tilde{\mathbf{U}}_1'>0\}$ for any $\Tilde{\mathbf{U}}_1'$, thus 
\begin{equation}\label{eq:47}
	\begin{split}
		&\E\limits_{\Tilde{\mathbf{U}}}\underset{x}{sup}\text{\#}\{x^T \Tilde{\mathbf{U}} >  0\}\\ 
		&= \E\limits_{(\Tilde{\mathbf{U}}_1', \Tilde{\mathbf{U}}_2')}sup_{x \in R^M}\big(\text{\#}(x^T\Tilde{\mathbf{U}}_1'>0) + \text{\#}(x^T\Tilde{\mathbf{U}}_2'>0)\big)\\
		&\geq \E\limits_{(\Tilde{\mathbf{U}}_1', \Tilde{\mathbf{U}}_2')}sup_{x \in \Omega}\big (\text{\#}(x^T\Tilde{\mathbf{U}}_1'>0) + \text{\#}(x^T\Tilde{\mathbf{U}}_2'>0)\big)\\
		&= M + \E_{\Tilde{\mathbf{U}}_2'}sup_{x \in \Omega}\text{\#}(x^T\Tilde{\mathbf{U}}_2'>0).
	\end{split}
\end{equation}
The last inequality is because $x^T \tilde{\mathbf{U}}_1' > 0$ when $x \in \Omega$. So $\text{\#}(x^T\Tilde{\mathbf{U}}_1'>0) =M$.\\
Since $\Tilde{\mathbf{U}}_2'$ is independent of $\Tilde{\mathbf{U}}_1'$,
\begin{equation}
	\begin{split}
		&\E_{\Tilde{\mathbf{U}}_2'}sup_{x \in \Omega}\text{\#}(x^T\Tilde{\mathbf{U}}_2'>0)\\
		&= \frac{1}{2}[\E_{\Tilde{\mathbf{U}}_2'}sup_{x \in \Omega}\text{\#}(x^T\Tilde{\mathbf{U}}_2'>0)+\E_{-\Tilde{\mathbf{U}}_2'}sup_{x \in \Omega}\text{\#}(x^T\Tilde{\mathbf{U}}_2'>0)]\\
		&= \frac{1}{2}[\E_{\Tilde{\mathbf{U}}_2'}sup_{x \in \Omega}\text{\#}(x^T\Tilde{\mathbf{U}}_2'>0)+ \E_{\Tilde{\mathbf{U}}_2'}sup_{x \in \Omega}\text{\#}(x^T\Tilde{\mathbf{U}}_2'<0)]. 
	\end{split}
\end{equation}

\noindent Consequently, 

\begin{equation} \label{eq:49}
	\begin{split}
		&\E_{\Tilde{\mathbf{U}}_2'}sup_{x \in \Omega}\text{\#}(x^T\Tilde{\mathbf{U}}_2'>0)\\
		&= \frac{1}{2}[\E_{\Tilde{\mathbf{U}}_2'}(sup_{x \in \Omega}\text{\#}(x^T\Tilde{\mathbf{U}}_2'>0)+ sup_{x \in \Omega}\text{\#}(x^T\Tilde{\mathbf{U}}_2'<0))] \\
		& \geq \frac{1}{2}[\E_{\Tilde{\mathbf{U}}_2'}(\E_{x \in \Omega}\text{\#}(x^T\Tilde{\mathbf{U}}_2'>0)+ \E_{x \in \Omega}\text{\#}(x^T\Tilde{\mathbf{\mathbf{U}}}_2'<0))
		\\&\quad+ \E_{x \in \Omega}\text{\#}(x^T\Tilde{\mathbf{U}}_2'=0) -\E_{x \in \Omega}\text{\#}(x^T\Tilde{\mathbf{U}}_2'=0)]\\
		&\geq \frac{1}{2} [\E_{\Tilde{\mathbf{U}}_2'}\E_{x\in\Omega}\text{\#}(x^T\Tilde{\mathbf{U}}_2 \in R^{KN-M})- \E_{\Tilde{\mathbf{U}}_2'}\E_{x\in\Omega}\text{\#}(x^T\Tilde{\mathbf{U}}_2'=0)]\\
		&=\frac{1}{2}(KN-M).
	\end{split}
\end{equation}

\noindent The last equation because $\E_{\Tilde{\mathbf{U}}_2' \in R^{M \times KN-M}}\text{\#}(x^T\Tilde{\mathbf{U}}_2'=0)] = 0$ for any nonzero vector $x$. Finally, according to Eq. \eqref{eq:47} and \eqref{eq:49}, we have: 
\begin{equation}
	\begin{split}
 \eta_F = \E_{\Tilde{\mathbf{U}}}sup_{x \in R^M} \text{\#}(x^T\Tilde{\mathbf{U}}>0) & \geq  M+\frac{1}{2}(KN-M) \\
	&=\frac{1}{2}(KN+M). 
	\end{split}
\end{equation}

\end{proof}

\section{Hypothesis Testing Details}


\noindent\textbf{Hypothesis Testing.} 
We treat both feature-based and backdoor-based watermark detection as a hypothesis testing process, whereas $\mathcal{H}_0$ is "the model is not an (illegal) copy of the model owned by its claimed owner" versus  $\mathcal{H}_1$ "model is is an (illegal) copy". Under hypothesis $\mathcal{H}_0$, given $N$ bit-length watermark,  the event that the model output the right watermark for each bit follows the binomial distribution $B(N, 1/C)$, whereas $C$ is the class number of the task. 

\noindent\textbf{Optimal bit-length and maximal bit-length $N$.}  
For \textbf{case 1} of Theorem \ref{thm: thm1}, we take the watermark detection rate $\eta_F$ as a constant which is independent of the bit-length $N$,  
 the upper bound of p-value is: 
\begin{equation}
	p\text{-}value \leq \sum_{i=\eta_0N}^{N}\binom{N}{i} (1/C)^i(1-1/C)^{N-i}, N<M/K, 
\end{equation}

\begin{table*}[htbp] 
  \vspace{-10pt}
  \centering
  \renewcommand{\arraystretch}{1}
  \normalsize

\resizebox{0.9\textwidth}{!}{
  \begin{tabular}{c|ccc}
    \hline
    Hyper-parameter & AlexNet & ResNet-18 & DistilBERT \\ \hline
    Optimization method & SGD & SGD & SGD\\
    Learning rate & 0.01 & 0.01 & 0.01\\
    Batch size & 16 & 16 & 32\\
    
    Data Distribution & IID and non-IID & IID and non-IID & IID\\
    Global Epochs & 200 & 200 & 80\\
    Local Epochs & 2 & 2 & 1\\
    Learning rate decay & 0.99 at each global Epoch &  0.99 at each global Epoch & linear schedule with warmup\\
    Regularization Term & BCE loss, Hinge-like loss& BCE loss, Hinge-like loss & BCE loss, Hinge-like loss \\ 
    $\alpha$ of Regularization Loss & 0.2, 0.5, 1, 5 & 0.2, 0.5, 1, 5 & 100 \\
    Feature-based watermark parameters $\mathbf{W}$ & $\mathbf{W}^k$ and $\mathbf{W}_{\gamma}$ & $\mathbf{W}_k$ and $\mathbf{W}_{\gamma}$ & $\mathbf{W}_{\gamma}$\\
    
    Trigger-based watermark type & Adversarial sample & Adversarial sample & \textbackslash{}\\
    \hline
\end{tabular}}
\caption{Training parameters for Federated AlexNet$_{\mathbf{p}}$ and ResNet$_{\mathbf{p}}$-18, respectively ($\dagger$ the learning rate is scheduled as 0.01, 0.001 and 0.0001 between epochs [1-100], [101-150] and [151-200] respectively). }
\label{tab:train-params}
\vspace{-10pt}
\end{table*}

in which $C=2$. The p-value is a \textit{monotonically decreasing} function of the bit-length $N$. Therefore, the optimal bit-length is $N_{opt} = M/K$ in this case. 

\begin{figure}[H]
\vspace{-10pt}
	\centering
	\includegraphics[width=2.2in]{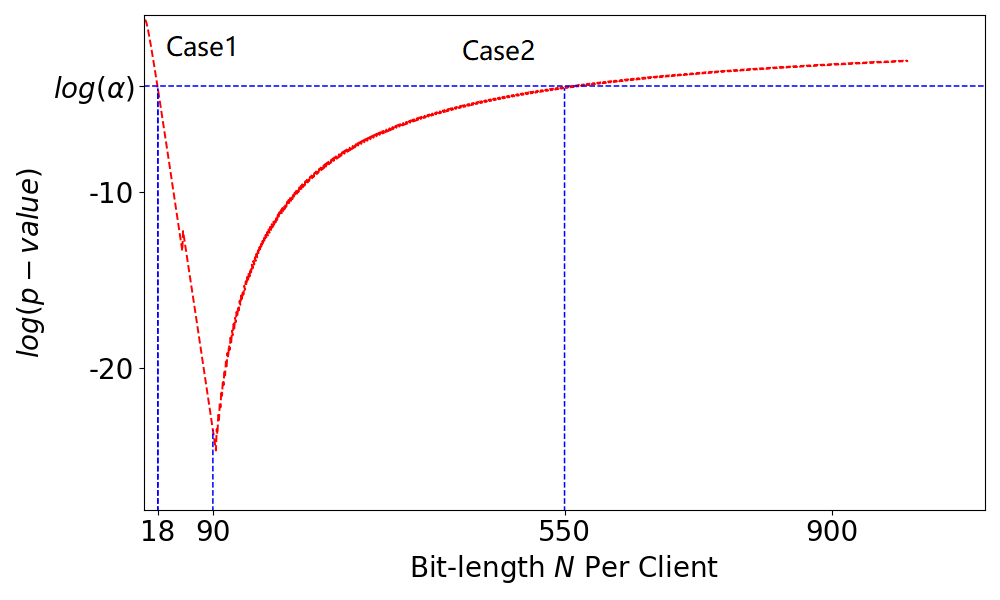}
	\caption{\textit{The optimal bit-length} and \textit{the acceptable range of watermark bit-length} that provide strong confidence of ownership verification. As shown in figure, $K=10, M=896$, we take $\eta_F=0.98$ in case 1, the optimal bit-length for p-value is $N_{opt}=M/K=90$, for an acceptable level $\alpha=0.0001$, the acceptable range of watermark bit-length is from $N=18$ to $550$. }
	\vspace{-10pt}
\end{figure}

For \textbf{case 2} of Theorem \ref{thm: thm1}, following Eq. \eqref{eq: case2} we take $\eta_F$ as a function inversely proportional to the bit-length $N$, 
thus the upper bound of p-value: 
\begin{equation}
	p\text{-}value \leq \sum_{i=\frac{KN+M}{2K}}^{N}\binom{N}{i} (1/C)^i(1-1/C)^{N-i}, N\geq M/K,   
\end{equation}
which \textit{monotonically increases} with the bit-length $N$ of watermarks. To sum up, the optimal bit-length for the smallest p-value is $N_{opt} = M/K.$

The optimal bit-length determined as such gives rise to the "smallest p-value," i.e., the most significant watermark verification. In case a p-value is only required to be less than statistical significance level $\alpha$ e.g. 0.0001, we can determine as follows the range of bit-length $N$ that fulfill the requirement.

\noindent\textbf{p-value of backdoor-based watermark.} 

Under hypothesis $\mathcal{H}_0$, given $N$ trigger-set samples, the event that the model $\mathbb{N}$ output the right label $y^T$ of $x^T$ follows the binomial distribution $B(N, 1/C)$, whereas $C$ is the class number of the task. 
The p-value for type I error is the probability of classifying at least $\eta_TN$ samples correctly 
\begin{equation}
	p\text{-}value = \sum_{i=\eta_TN}^N \binom{N}{i} (1/C)^i(1-1/C)^{N-i}. 
\end{equation}

\section{Experimental Setting Details}

\begin{table}[htbp]
  \centering
    \renewcommand{\arraystretch}{1.05}
    \normalsize
  \resizebox{0.48\textwidth}{!}{
    \begin{tabular}{c|cc}
      \hline
      Hyper-parameter & Value \\ \hline
      Optimization method & Projected Gradient Descent \\
      Norm type & L2  \\
      Norm of noise & 0.3  \\
      Learning rate & 0.01 \\
      PGD Batch size & 128  \\
      Iterations & 80  \\
      Vanilla Classification model & CNN with 3 convolution layers\\
      \hline
      \end{tabular}
  }
  \caption{Training parameters for Projected Gradient Descent Adversarial Training}
  \label{tab:PGD-params}
\end{table}

\begin{table}[htbp]
 \centering
  \renewcommand{\arraystretch}{1.05}
      
      \setlength{\tabcolsep}{1.5mm}
        
      \begin{tabular}[l]{c|c|c|c}
        \toprule
        Layer name & Output size & Weight shape & Padding \\ 
        \hline
        Conv1 & 32 $\times$ 32 & 64 $\times$ 3 $\times$ 5 $\times$ 5 & 2 \\
        MaxPool2d & 16 $\times$ 16 & 2 $\times$ 2 &  \\
        Conv2 & 16 $\times$ 16 & 192 $\times$ 64 $\times$ 5 $\times$ 5 & 2 \\
        Maxpool2d & 8 $\times$ 8 & 2 $\times$ 2 &  \\
        Conv3 & 8 $\times$ 8 & 384 $\times$ 192 $\times$ 3 $\times$ 3 & 1 \\
        Conv4 & 8 $\times$ 8 & 256 $\times$ 384 $\times$ 3 $\times$ 3 & 1 \\
        Conv5 & 8 $\times$ 8 & 256 $\times$ 256 $\times$ 3 $\times$ 3 & 1 \\
      
        MaxPool2d & 4 $\times$ 4 & 2 $\times$ 2 &  \\
        Linear & 10 & 10 $\times$ 4096 &  \\ \bottomrule
        
      \end{tabular} 
      \caption{Modified AlexNet Architecture (Embed feature-based watermarks across Conv3, Conv4 and  Conv5, thus $M$ =896)}\label{tab: AlexNet}
      \vspace{-10pt}
  \end{table}	

\begin{table}[htbp]
   \centering
  \renewcommand{\arraystretch}{1.0}
      \setlength{\tabcolsep}{1.5mm}
     
      \begin{tabular}[c]{c|c|c|c}
      
        \toprule
        Layer name & Output size & Weight shape & Padding \\ 
        \hline
        Conv1 & 32 $\times$ 32 & 64 $\times$ 3 $\times$ 3 $\times$ 3 & 1 \\ 
        \hline
        Res2 & 32 $\times$ 32 & $\begin{bmatrix}
          64 \times 64 \times 3 \times 3 \\
          64 \times 64 \times 3 \times 3 \\
        \end{bmatrix} \times 2$ & 1 \\
        \hline
        Res3 & 16 $\times$ 16 & $\begin{bmatrix}
          128 \times 128 \times 3 \times 3 \\
          128 \times 128 \times 3 \times 3 \\
        \end{bmatrix} \times 2$ & 1 \\
        \hline
        Res4 & 8 $\times$ 8 & $\begin{bmatrix}
          256 \times 256 \times 3 \times 3 \\
          256 \times 256 \times 3 \times 3 \\
        \end{bmatrix} \times 2$ & 1 \\
        \hline
        Res5 & 4 $\times$ 4 & $\begin{bmatrix}
          512 \times 512 \times 3 \times 3 \\
          512 \times 512 \times 3 \times 3 \\
        \end{bmatrix} \times 2$ & 1 \\
        \hline
        Linear & 100 & 100 $\times$ 512 &  \\ \bottomrule
      \end{tabular}
      \caption{Modified ResNet-18 Architecture (Only embed feature-based watermarks across Res5 Block, thus $M$ = 2048)}\label{tab: Resnet18}

  \end{table}

\end{document}